\documentclass[]{sensenova}

\usepackage{hyperref}
\usepackage{cleveref}
\usepackage{verbatim}
\usepackage{wrapfig}
\usepackage{graphicx}
\usepackage{floatrow}
\usepackage{placeins}
\usepackage{subcaption}
\usepackage{listings}
\usepackage{algorithm}
\usepackage{subcaption}
\usepackage{dsfont}

\usepackage{mmstyles}

\usepackage[toc,page,header]{appendix}

\usepackage{xspace}
\usepackage[utf8]{inputenc}  
\usepackage[T1]{fontenc}     
\usepackage{CJKutf8} 
\usepackage[table]{xcolor} 

\usepackage[misc]{ifsym}

\usepackage{CJKutf8} 
\usepackage[colorinlistoftodos,prependcaption,textsize=small]{todonotes}

\usepackage{fontawesome5}

\usepackage{soul} 
\definecolor{HLGreen}{rgb}{0.78,0.95,0.78}
\definecolor{HLRed}{rgb}{0.9725,0.8431,0.8549}   
\definecolor{benchblue}{RGB}{58, 122, 252}

\newcommand{\colornum}[1]{%
  \footnotesize
  \ifdim #1 pt < 0pt
    \textcolor{sensepurple}{#1}%
  \else
    \textcolor{benchblue}{#1}%
  \fi
}

\usepackage{cuted} 

\definecolor{bestcolor}{RGB}{219, 208, 237}
\definecolor{secondcolor}{RGB}{241, 237, 248}
\definecolor{thirdcolor}{RGB}{211, 222, 190}
\definecolor{line-blue}{RGB}{243, 248, 252}
\definecolor{line-green}{RGB}{200,242,200}
\definecolor{line-red}{RGB}{255,215,215}

\usepackage{adjustbox}
\usepackage{makecell}
\usepackage{booktabs}

\newlength{\ModelW}
\newlength{\NonModelW}
\newcolumntype{M}{>{\centering\arraybackslash}p{\NonModelW}}
\usepackage{graphicx}
\usepackage{subcaption}
\graphicspath{{assets/figures/}}

\usepackage{array}      
\usepackage{caption}
\usepackage{subcaption} 

\usepackage{enumitem}

\usepackage{multirow}
\usepackage{makecell}

\usepackage{booktabs,tabularx,array}
\newcolumntype{Y}{>{\centering\arraybackslash}X}
\newcolumntype{L}{>{\raggedright\arraybackslash}X}

\usepackage{framed}
\usepackage{algorithm}
\usepackage{algorithmic}
\usepackage{listings}
\usepackage{soul}          

\definecolor{cotredbg}{RGB}{255,220,220}   
\definecolor{cotgreenbg}{RGB}{220,255,220} 

\title{Has GPT-6-Astra Achieved Embodied Intelligence?}

\author{

\textbf{Joint Research Team}
\vspace{-10mm}

}

\abstract{

GPT-6-Astra demonstrates a level of competence in robotic manipulation and navigation that challenges the conventional paradigm of building specialized models for individual tasks and embodiments. In this report, we ask not only how capable GPT-6-Astra is, but what its capabilities imply for the field.
\textbf{First, we probe the limits of standalone GPT-6-Astra.} We conduct a large-scale evaluation across 12 representative benchmarks spanning tabletop manipulation, humanoids, mobile platforms, and dexterous hands. Under a unified protocol that allows in-context learning (ICL) from training-set demonstrations and substantially increases the number of execution steps, GPT-6-Astra achieves unprecedented performance (\cref{fig:teaser}). Our analysis suggests that future research should increasingly focus on improving
1) the generality of in-context learning
2) cost and execution time, 
3) physical-state verification,
4) recovery from irreversible failures, and
5) long-horizon preservation of progress.
\textbf{Second, we study the agentic harness for embodied control.} Besides exploring the ideal action representation and chucking parameters, we identify that close-loop coordination between a high-level "System 2" reasoner (\ie, GPT-6-Astra) and a low-level "System 1" controller (\eg, $\pi_{0.5}$) is costly but effective.
\textbf{Third, we examine GPT-6-Astra on real-world deployment.} Our experiments on 8 robot configurations reveal substantial gaps: 1) robot demonstrations required for in-context learning are expensive to collect, and 2) it is implausible to trade longer execution for higher success rates with physical constraints.
\textbf{Fourth, we envision the next generation of embodied benchmarks.} We develop a prototype that 1) derives tasks from real-world needs rather than current robot capabilities, 2) provides no in-domain demonstrations, and 3) judges agents on a world clock that keeps running while they deliberate.
Taken together, our results suggest that GPT-6-Astra has largely crossed the threshold of embodied competence, but not yet that of practical embodied intelligence. The key remaining challenge is to translate benchmark success into reliable, efficient, and scalable real-world deployment.

\checkdata[Homepage]{\url{https://www.sensenova.ai/has-gpt-6-astra-achieved-embodied-intelligence}} 
}

\begin{document}
\maketitle

\begin{figure}[htbp]
    \centering
    \vspace{-6mm}
    \includegraphics[width=\linewidth]{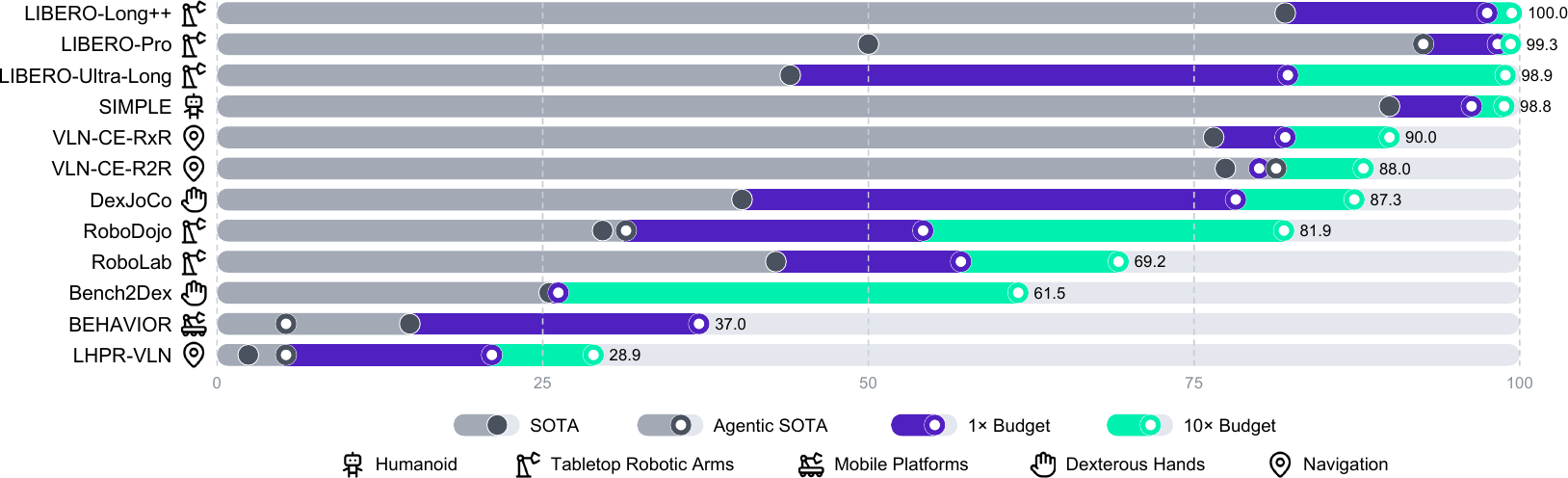}
    \caption{\textbf{Probing the capability upper bound of GPT-6-Astra across 12 benchmarks} under a unified protocol that 1) allows for in-context learning (ICL) on the training set and 2) extends the step limit (budget) up to $10\times$. 
    RoboLab has no training set and thus no ICL; BEHAVIOR is not extended to $10\times$ due to its large official budget. 
    SOTA cut-off date: October 12, 2026. Details in~\cref{tab:teaser_results}. }
    \label{fig:teaser}
\end{figure}

\section{Introduction}
\label{sec:intro}

The advent of GPT-6-Astra marks a major shift in embodied intelligence: a general-purpose model can now tackle robotic manipulation and navigation tasks that have traditionally been dominated by specialized models, such as VLAs and WAMs, while exhibiting capabilities including zero-shot generalization, in-context learning, and failure recovery. 
In this technical report, we use GPT-6-Astra as a lens to examine a broader question: what does this new level of embodied capability mean for the research community, and where should the field go next? We first conduct a large-scale evaluation across diverse embodied tasks to characterize the extent of its capabilities and identify which problems are already approaching saturation. We then study harness design to understand the performance--cost trade-offs involved in translating these capabilities into effective control. Next, we validate our findings on real robots, asking how far benchmark success carries over to physical deployment and what gaps remain.Finally, we revisit benchmark design and envision what the next generation of embodied benchmarks should look like.

\textbf{First, we probe the upper bound of GPT-6-Astra's embodied capabilities under a unified protocol.} We conduct a large-scale study across 12 representative benchmarks, covering tabletop robotic arms, humanoids, mobile platforms, and dexterous hands. Under this protocol, GPT-6-Astra outperforms existing methods by substantial margins, including recent agentic approaches that also build on GPT-6-Astra. Notably, as shown in~\cref{fig:teaser}, it approaches saturation on the LIBERO series and SIMPLE, while achieving 81.9\% on RoboDojo and 87.3\% on DexJoCo. The protocol follows two simple and well-motivated principles:

\noindent
\emph{1) We allow in-context learning (ICL) from the training set} whenever successful demonstrations are available, which we consider reasonable since such demonstrations are conventionally accessible when training smaller task-specific policies. GPT-6-Astra studies the available training data and constructs a frozen knowledge package containing task guidance, motion and visual references, executable helpers, recovery procedures, and metadata. At test time, it retrieves and adapts these resources to the current scene. This substantially improves performance when the demonstrations closely cover the test distribution, but the benefit weakens as tasks move away from the data used to build the package.

\noindent
\emph{2) We extensively increase the execution budget}, allowing up to 10$\times$ more interaction steps than the official setting. We find that standard step limits can significantly underestimate GPT-6-Astra's capabilities. Additional interaction yields large gains on tasks that benefit from repeated correction, such as RoboDojo and Bench2Dex, because GPT-6-Astra can detect failed actions, revise its strategy, and recover through further interaction. This emerging failure-recovery capability is the key mechanism that allows a larger execution budget to translate into higher success rates.

\noindent
Our strong empirical results put us in a position to identify the \emph{remaining limitations of GPT-6-Astra} more clearly. Specifically, we find that:
i) in-context learning transfer poorly beyond the tasks from which the demonstrations are drawn;
ii) even simple tasks incur inference cost and execution time far beyond what would be practical for commercial deployment;
iii) GPT-6-Astra often struggles to verify physical success and identify which task condition remains unmet;
iv) GPT-6-Astra often struggles to identify the cause of task failure;
v) it cannot reliably recover from consequential or irreversible failures; and
vi) long-horizon execution remains difficult, as the model struggles to preserve and track progress across many subtasks.

\textbf{Second, we ablate the harness design} and find that effectively unleashing GPT-6-Astra's embodied performance depends critically on both the collaboration protocol and the action interface. We first study the interaction between ``System 2'' (\ie, GPT-6-Astra) and ``System 1'' (\eg, $\pi_{0.5}$~\cite{pi2025pi05}). Closed-loop collaboration, where GPT-6-Astra reviews and refines VLA-proposed actions before execution, substantially improves performance over standalone or open-loop execution, but frequent System-2 supervision incurs significant token cost. Reviewing longer action sequences improves this trade-off, while combining closed-loop collaboration with ICL yields the strongest performance at the highest cost.
We further study action representation and chunking, and identify a representation gap between multimodal reasoning and low-level robot control. Direct joint-angle control underperforms, while more semantically aligned end-effector commands work better; adaptive action chunking further improves performance while reducing unnecessary model calls. Together, these findings suggest that harness design should expose action spaces aligned with the model's reasoning capabilities while delegating precise execution to lower-level controllers, with adaptive interaction to balance performance and inference cost.

%
\textbf{Third, we validate GPT-6-Astra on real-world robots} spanning single and dual robotic arms, grippers, dexterous hands, a wheeled mobile manipulator, and a humanoid. Our study reveals three key findings. 
1) GPT-6-Astra operates effectively across a wide range of embodiments, but its success rate remains highly task-dependent. It performs particularly well on tasks such as tabletop grasping, while deformable-object manipulation, precise contact, and long-horizon tasks remain challenging. 
2) GPT-6-Astra demonstrates surprising active-vision capabilities, intentionally repositioning the camera to obtain observations needed for subsequent actions. 
3) GPT-6-Astra shows that a slow ``System 2'' model can directly collaborate with a high-frequency ``System 0'' locomotion controller to perform humanoid manipulation that requires high-frequency local control.
At the same time, real-world deployment introduces challenges that are largely absent from simulation. 
GPT-6-Astra can no longer easily trade additional execution steps for higher success rates, due to safety constraints, mechanical limits, and real-time requirements. 
Moreover, successful trajectory demonstrations, which are central to our ICL setup in simulation, are often expensive or unavailable in real-world applications.

\textbf{Fourth, we showcase a prototype of what we envision as the next generation of benchmarks} for embodied intelligence.
We propose three principles. 
1) Tasks should be derived from human needs and expressed as people would naturally state them, with evaluation based on the utility of the outcome rather than a fixed goal state.
2) No in-domain training episodes or scene-specific demonstrations should be provided; instead, agents should acquire experience during deployment.
3) The world should continue to evolve while the agent deliberates, making reasoning time and its physical consequences part of the evaluation.
The prototype makes several remaining limitations more apparent in a deployment-oriented setting:
i) GPT-6-Astra focuses on the current step but can lose track of other changes it has caused in the environment, such as leaving a refrigerator open;
ii) without demonstrations, it struggles to figure out how unfamiliar objects and mechanisms work, or how to position its own body to interact with them; and
iii) because the world continues to change while it reasons, its observations can become outdated before the resulting action is executed.
Building such benchmarks also raises practical challenges in reliable judging, simulation fidelity, real-time evaluation, and evaluation cost.

In summary, by pushing GPT-6-Astra toward its capability limits, dissecting the harness that enables this performance, validating its behavior on real robots, and rethinking how embodied intelligence should be evaluated, we aim to clarify both what has changed and what remains fundamentally unsolved.

\section{Large-Scale Study under A Unified Protocol}
\label{sec:study}
\providecommand{\studyplaceholder}[2]{%
  \fcolorbox{red}{red!4}{\parbox[c][#1][c]{\dimexpr\linewidth-2\fboxsep-2\fboxrule\relax}{%
  \centering\footnotesize\textcolor{red}{#2}}}}

We study GPT-6-Astra across twelve embodied settings, spanning manipulation, humanoid loco-manipulation, household activities, and navigation, under a single protocol in which its model weights remain fixed. Two conditions shape the study: in-context learning (ICL), which gives GPT-6-Astra access to the same training demonstrations from which conventional policies are learned, and an extended interaction budget, which gives it more opportunity to correct mistakes within an episode. 

\subsection{Protocol}
\label{sec:study_protocol}
Our study covers twelve evaluation settings, grouped into five categories: tabletop manipulation (RoboDojo~\cite{chen2026robodojo}, RoboLab~\cite{yang2026robolab}, LIBERO-Long++ and LIBERO-Ultra-Long~\cite{yang2026lilo}, and LIBERO-PRO~\cite{zhou2025liberopro}), dexterous manipulation (Bench2Dex~\cite{yang2026bench2dex} and DexJoCo~\cite{wang2026dexjoco}), humanoid loco-manipulation (SIMPLE~\cite{wei2026simple}), long-horizon household activities (BEHAVIOR~\cite{behavior2026challenge}), and vision-language navigation (VLN-CE R2R and RxR~\cite{krantz2020vlnce}, and LHPR-VLN~\cite{song2025lhvln}).

\begin{figure}[t]
  \centering
  \includegraphics[width=\linewidth]{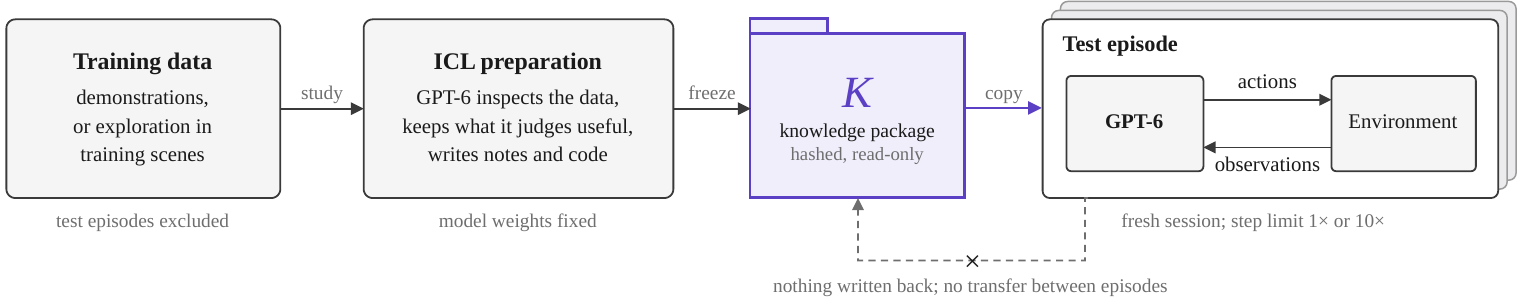}
  \caption{\textbf{In-context learning protocol.} For each task, GPT-6-Astra studies the training data and writes a knowledge package $K$, which is frozen before testing. Each test episode starts a fresh session with  $K$ and runs under the standard ($1\times$) or extended ($10\times$) budget. Model weights remain fixed throughout, and $K$ remains unchanged during testing, and no information passes between episodes.}
  \label{fig:icl_protocol}
\end{figure}

We evaluate GPT-6-Astra under a demonstration-informed protocol with two interaction budgets (1$\times$ and 10$\times$), as illustrated in Fig. \ref{fig:icl_protocol}. The protocol separates preparation using training demonstrations from test-time interaction. We report performance under the standard budget and an extended budget that allows more interaction steps for execution and recovery.

\noindent\textbf{Motivation.}
Learned robotic policies are commonly trained or fine-tuned on demonstrations provided by the benchmark. Evaluating GPT-6-Astra without access to these demonstrations would therefore measure its performance under a different information regime. We use in-context learning (ICL) to investigate how effectively GPT-6-Astra can exploit the available training demonstrations without updating its model weights. Our objective is to characterize its capabilities when the task-relevant information used to train conventional policies is also available to it through contextual preparation.

We additionally investigate the effect of the interaction budget. In our evaluations, GPT-6-Astra often identifies unsuccessful actions, revises its strategy, and recovers through further interaction. A strict step limit can terminate this process before recovery succeeds. We therefore evaluate under two budgets. The standard budget ($1\times$) is the interaction budget prescribed by each benchmark and measures performance under its intended conditions; the extended budget ($10\times$) allows up to ten times as many action steps and measures how task completion improves when the policy has more opportunities to correct mistakes.

\noindent\textbf{In-context learning preparation.}
Before testing, GPT-6-Astra undergoes a separate ICL preparation phase for each task, using the available training demonstrations represented through benchmark-appropriate observations and demonstrated actions. During preparation, GPT-6-Astra may analyze the demonstrations and develop task-specific notes and reusable code. These materials form a knowledge package, which is frozen before evaluation and provided to every test episode of that task. GPT-6-Astra’s model weights remain unchanged. Preparation excludes held-out test episodes and evaluator-private information.

\noindent\textbf{Episode evaluation.}
Each test episode begins in a fresh, isolated session with the task instruction, the frozen knowledge package, and the permitted observation and action interfaces. At each decision point, GPT-6-Astra examines the current observations and specifies the next action or action sequence. The environment executes these actions and returns new observations, which GPT-6-Astra uses to assess the outcome and determine its next action. The interaction history is retained within the episode, allowing GPT-6-Astra to diagnose mistakes and attempt recovery. Neither the knowledge package nor the model weights are updated during testing, and information from one test episode is not transferred to another. Observation access, robot-control action semantics, control frequency, and success criteria remain fixed across budget conditions.

\noindent\textbf{Extended interaction budget.}
The extended budget allows ten times as many action steps as the standard budget. We report outcomes at $1\times$ and at cumulative cutoffs of $2\times,\ldots,10\times$ the standard budget, using the same initialization, policy configuration, and success criterion. An extension continues the same episode, preserving its interaction history and accumulated environmental consequences; additional attempts occur within that episode rather than through resets. Episodes terminate upon verified success, an applicable terminal failure, or exhaustion of the allotted budget. Reporting success across these cutoffs quantifies the contribution of additional interaction while retaining the standard-budget result as a separate comparison point.

\subsection{In-context Learning}
\label{sec_icl_package}

With in-context learning, GPT-6-Astra achieves strong performance across embodied benchmarks without any update to its model weights. Figure~\ref{fig:teaser} compares GPT-6-Astra with the best published result (SOTA) and the best published agentic method (Agentic SOTA) on all 12 settings, both reported under the standard budget. Already at the standard ($1\times$) budget, GPT-6-Astra matches or exceeds the strongest published result on most settings, with the largest margins on LIBERO-Ultra-Long and DexJoCo, where it nearly doubles the best published success rate. To isolate the contribution of ICL, we compare zero-shot and ICL conditions on SIMPLE: task-level ICL raises standard-budget success on the four common tasks from 70.8\% to 90.8\% and shortens successful episodes (Table~\ref{tab:simple_efficiency}). The benefit of ICL, however, depends on how closely the training data covers the test tasks. Where test tasks closely follow the training distribution, as in the LIBERO family, SIMPLE and VLN-CE, the package can retrieve and adapt demonstrated motions or routes almost directly, and success approaches saturation. DexJoCo's contact-rich dexterous tasks are demanding, but phase-indexed motion references and controller helpers cover them well. RoboLab, in contrast, provides no training set from which a task-level package can be built, and its success remains moderate; Sec.~\ref{sec_remaining_failures} discusses this limitation further.

\begin{figure}[t]
  \centering
  \includegraphics[width=\linewidth]{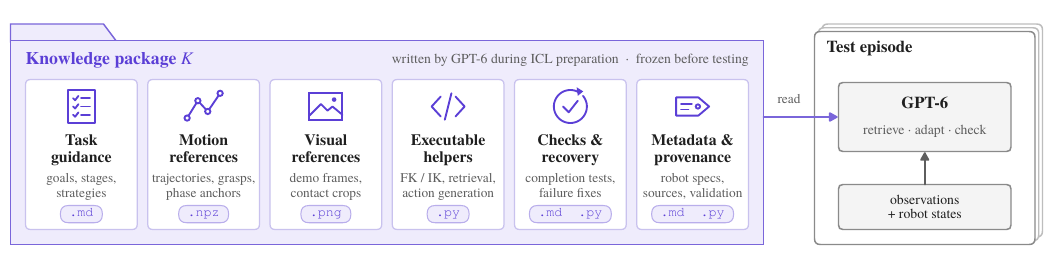}
  \caption{\textbf{Contents of an ICL knowledge package.} GPT-6-Astra decides what to retain, yet packages share six recurring components; file types are typical and content varies by benchmark. At test time, $K$ is read alongside current observations to retrieve, adapt and check actions.}
  \label{fig:icl_package}
\end{figure}

\noindent\textbf{Package contents.}
Each knowledge package is organized around six recurring components (Fig.~\ref{fig:icl_package}): task guidance, motion references, visual references, executable helpers, checks and recovery, and metadata and provenance. This structure is not prescribed. During ICL preparation, GPT-6-Astra independently chooses what to inspect and retain within each benchmark's permitted training data and tool interfaces, and we do not specify which demonstrations, frames, or state/action records it should examine or save; the six components nonetheless recur across benchmarks.

\begin{itemize}
    \item \textbf{Task guidance.}
    Written instructions describe task goals, execution stages, and strategies. They specify subtask ordering, object roles, and constraints that guide the selection and use of stored references.

    \item \textbf{Motion references.}
    Demonstration-derived trajectories, grasp configurations, and phase anchors provide numerical examples of how actions unfold. The agent retrieves relevant segments or poses and adapts them to the current scene and robot state.

    \item \textbf{Visual references.}
    Selected demonstration frames and contact crops illustrate object appearance, grasp regions, and intermediate task states. These examples support interpretation of the stored motions and comparison with current observations.

    \item \textbf{Executable helpers.}
    Reusable Python programs support forward and inverse kinematics, reference retrieval, and action generation. They convert stored references and current geometric estimates into candidate actions.

    \item \textbf{Checks \& recovery.}
    Completion tests describe observable evidence that a grasp, subtask, or overall task has succeeded. Recovery procedures associate failure symptoms with corrective actions, such as adjusting an approach or selecting an alternative grasp.

    \item \textbf{Metadata \& provenance.}
    Robot specifications and interface conventions define how stored states and actions should be interpreted. Source records and validation notes document the origin of the retained material and the scope of its offline checks.
\end{itemize}

\begin{figure}[htbp]
    \centering
    \includegraphics[width=\linewidth]{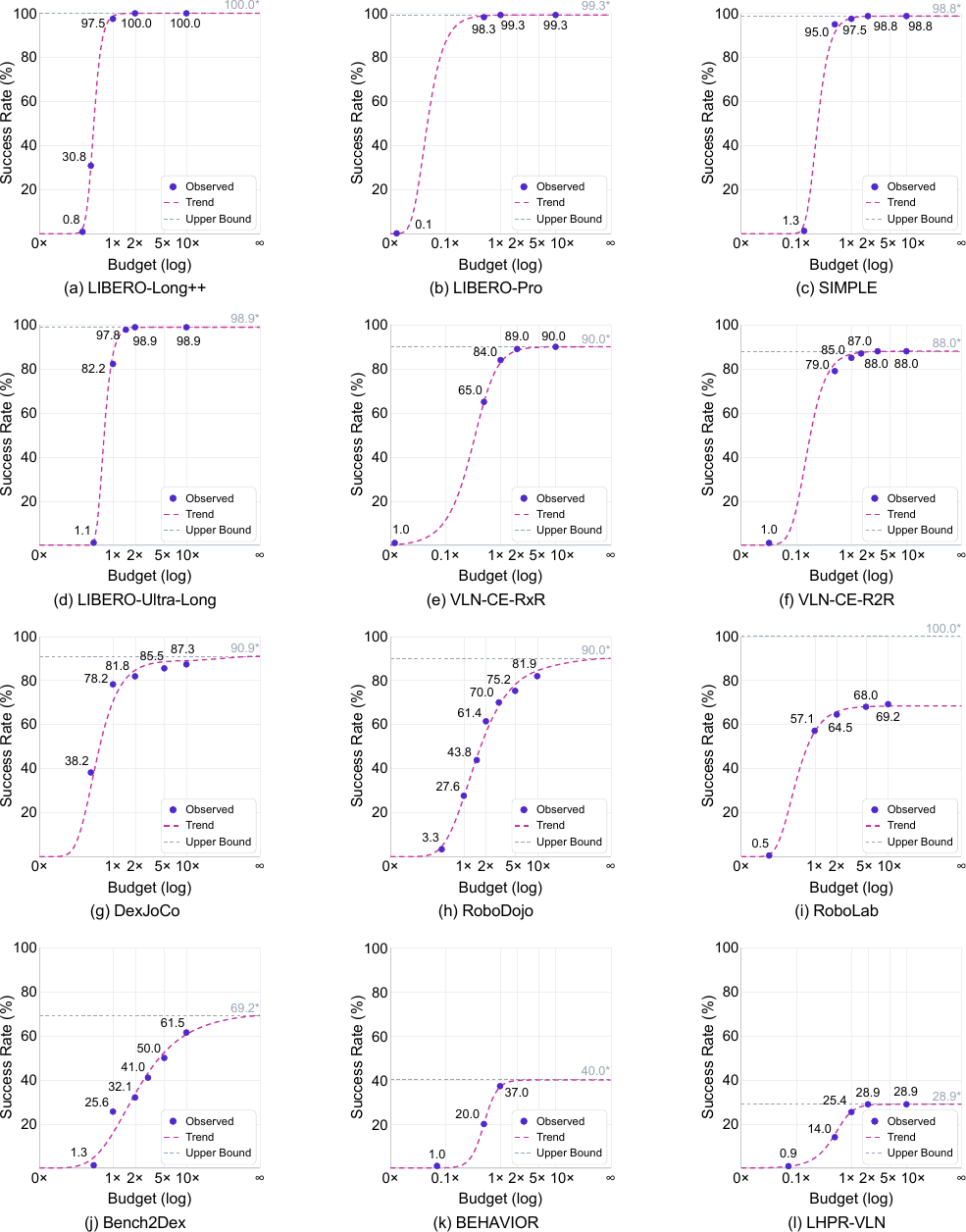}
    \caption{\textbf{Success rate versus step budget.} The budget is expressed as a multiple of each benchmark's official step limit; dots show observed success rates and pink curves show fitted trends. We evaluate budgets up to $10\times$ the official limit ($1\times$ for BEHAVIOR). The gray line marks the upper bound, which assumes that every episode that exhausts the evaluated budget without succeeding has the potential to succeed with an unlimited budget.}
    \label{fig:step_sr}
\end{figure}

Within this shared structure, the retained information reflects each benchmark's demands. Bench2Dex \cite{yang2026bench2dex} and DexJoCo \cite{wang2026dexjoco} emphasize hand configurations, phase-indexed motion examples, and helpers for dexterous control. LIBERO-Long++ and LIBERO-Ultra-Long \cite{yang2026lilo} organize selected demonstrations around composable atomic skills, while LIBERO-PRO \cite{zhou2025liberopro} primarily retains textual strategies and recovery knowledge from exploration. SIMPLE \cite{wei2026simple} preserves complete numerical training trajectories and helpers for extracting and modifying candidate action chunks under current feedback. LHPR-VLN \cite{song2025lhvln} and VLN-CE \cite{krantz2020vlnce} retain navigation strategies, landmark references, stopping guidance, and recovery notes from training-scene practice. BEHAVIOR \cite{behavior2026challenge} emphasizes object-specific grasping affordances, bimanual and tool-contact geometry, and conditional verification and recovery procedures.

\noindent\textbf{Package usage.}
Retrospective usage analyses reveal a common workflow on every benchmark we audited: GPT-6-Astra reads the task guidance, retrieves a candidate motion through a prepared helper, and checks its execution against the visual references. Numerical demonstrations are frequently loaded internally by these helpers, linking stored examples directly to action generation; on DexJoCo, controllers and visual references are used repeatedly during both initial execution and continuation. Rather than replaying demonstrations, GPT-6-Astra adapts them to the current scene. In Bench2Dex Ball Box Loading, it remaps reference joints and compensates for a 4--5\,cm difference in workspace height while preserving the demonstrated bimanual support geometry; in DexJoCo Photograph, it selects an alternative reference pickup after a slip. The packages thus serve as reusable resources for proposing, adapting, and revising actions under feedback, although these analyses do not isolate the causal contribution of individual components.

\subsection{Extending Interaction Budget}
\label{sec_simulation_results}

Beyond the coverage of training data (Sec.~\ref{sec_icl_package}), the demands of the tasks themselves limit performance at $1\times$, leaving four benchmarks far from solved even with ICL: RoboDojo, Bench2Dex, BEHAVIOR, and LHPR-VLN. RoboDojo and Bench2Dex require precise bimanual or dexterous manipulation, often involving repeated corrections. BEHAVIOR and LHPR-VLN require long sequences of household manipulation or navigation, with subgoals tracked across many stages, and have the lowest success rates of all settings.

To assess how much the standard step limit contributes to these shortfalls, we extend the budget to $10\times$ while keeping model weights and ICL packages fixed (Figs.~\ref{fig:teaser} and~\ref{fig:step_sr}). Additional interaction yields large gains on RoboDojo and Bench2Dex. On RoboDojo, success roughly triples, and about three quarters of the episodes unfinished at $1\times$ are completed, with the Precision group gaining most. Success on Bench2Dex more than doubles. These gains indicate that the standard budget can interrupt productive execution and recovery, while additional interaction allows GPT-6-Astra to complete further corrective attempts.

On LHPR-VLN, however, the extended budget yields only a modest gain, and success remains the lowest of all settings. This suggests that the demands of long-horizon navigation, including tracking subgoals across many stages, are not resolved by additional interaction alone (Sec.~\ref{sec_remaining_failures}). We do not extend the budget on BEHAVIOR, whose standard budget is already long.

Benchmarks that already perform well at $1\times$ also benefit from additional interaction. DexJoCo and VLN-CE reach success rates of around 90\%, while the LIBERO family and SIMPLE exceed 95\% and approach saturation (Fig.~\ref{fig:teaser}).

\begin{figure}[t]
    \centering
    \includegraphics[width=\linewidth]{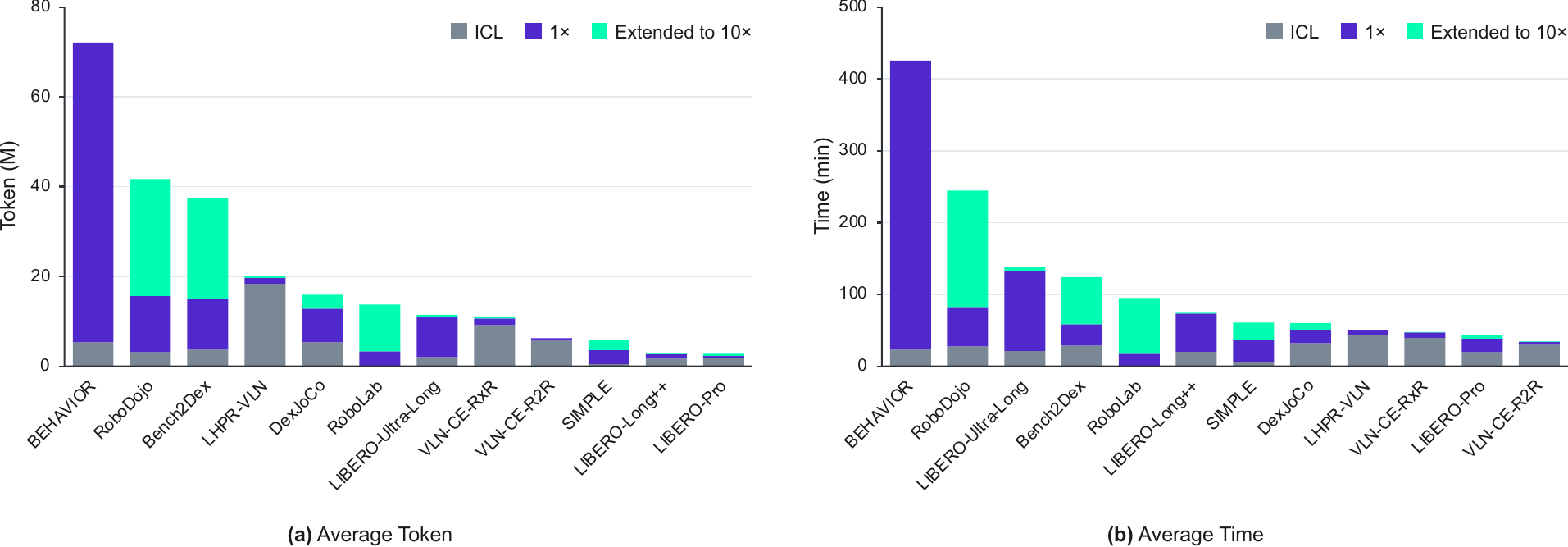}
    \caption{\textbf{(a) Average per-task token usage} and \textbf{(b) average per-task time} for each benchmark, stacked from bottom to top as ICL preparation, the \(1\times\) run, and the additional cost of extending to \(10\times\). Here, \(1\times\) and \(10\times\) denote maximum action-step budgets, not token or time budgets. The relatively small additional token and time costs primarily reflect early termination of extended runs upon success or terminal failure, before the full \(10\times\) budget is exhausted. We observe that the vast majority of tokens consumed are cached input tokens (approximately 98\% in our runs). Accordingly, we use a simple rule of thumb for cost estimation: \textbf{1M tokens $\approx$ 1 USD}.}
    \label{fig:cost}
\end{figure}


\subsection{Limitations and Failure Modes}
\label{sec_remaining_failures}

\begin{figure}[t]
\centering
\includegraphics[width=\linewidth]{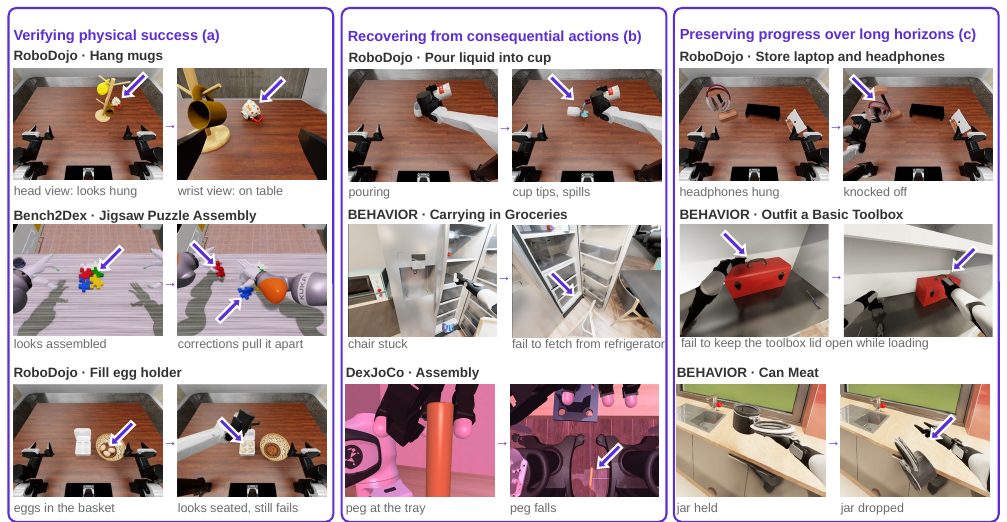}
\caption{\textbf{Residual failures under ICL with the extended budget.} Each pair shows the state before a key event and its outcome; purple arrows mark the relevant objects. \textit{(a) Verifying physical success:} apparent or self-reported completion diverges from the actual task state. \textit{(b) Recovering from consequential actions:} an early mishap, such as a spill or a dropped part, cannot be undone and derails later subtasks. \textit{(c) Preserving progress over long horizons:} progress that later steps depend on, such as an open lid or a placed object, is lost and blocks the task.}
\label{fig:failure_gallery}
\end{figure}

Despite the substantial gains from ICL and the extended budget, several limitations and recurring failure modes remain. We first discuss two limitations of the approach, transfer and cost, and then three behavioral failure modes identified from episode videos and agent logs (Fig.~\ref{fig:failure_gallery}). Infrastructure interruptions remain scored as failures but are considered separately from the behavioral analysis.

\noindent\textbf{ICL packages are largely task-specific, which limits transfer.}
Although ICL packages substantially improve performance across benchmarks, they are built from each benchmark's training data and mostly encode task-level guidance, motion references, and helpers, so their benefit weakens when test tasks move away from the data they were built from. In the RoboDojo Open group, which has no task-specific demonstrations and relies on a package studied from the other four groups, GPT-6-Astra + ICL still benefits from additional interaction but attains the lowest success among all groups (Appendix~\ref{app:robodojo}, Table~\ref{tab:robodojo_results}). Similarly, RoboLab provides no training set from which a package can be built; GPT-6-Astra leads the listed baselines there, yet its absolute success remains moderate (Table~\ref{tab:robolab_results}). Building packages that generalize beyond their source tasks remains an open direction.

\noindent\textbf{Time and cost remain high.}
Strong performance comes at a substantial cost in tokens and wall-clock time, even on benchmarks that GPT-6-Astra nearly solves (Fig.~\ref{fig:cost}). On LIBERO-Long++, where success at the standard budget already exceeds 97\%, each episode still consumes about a million tokens and nearly an hour of inference and tool use; on LIBERO-Ultra-Long, these costs rise to roughly nine million tokens and nearly two hours (Table~\ref{tab:libero_long_results}). Extending the budget increases the cost further, and much of the additional cost is concentrated in difficult episodes that often still fail: on RoboDojo, a single episode under the extended budget takes more than three hours on average. Since the $10\times$ setting only caps the number of action steps and episodes may end earlier upon success or terminal failure, these costs reflect the steps actually executed. Because inference latency is not counted in the action-step budget, this cost also limits deployment in settings that require real-time control.

\noindent\textbf{Verifying physical success.}
GPT-6-Astra can detect that a task is not yet complete without identifying which physical condition remains unmet; consequently, its correction process fails to converge (Fig.~\ref{fig:failure_gallery}(a)). In Bench2Dex Jigsaw Puzzle Assembly, it reports that all four pieces form the required rectangle while noting that the episode has not ended, and continues to adjust the residual offsets, but its corrections pull the pieces apart again instead of closing the remaining gaps. In RoboDojo Hang mugs, GPT-6-Astra takes an overlap between a mug's handle and the peg in its head camera as evidence that the mug is hung, without confirming that the peg passes through the handle; the mug ends up on the table, and in its later regrasps the closing of its fingers is likewise taken as contact that has not occurred. In RoboDojo Fill egg holder, GPT-6-Astra moves all four eggs from the basket into the holder and reports that they remain seated after it withdraws, yet the episode ends as a failure with only partial credit. Visual plausibility and self-reported progress can therefore diverge from the task's actual completion conditions, sustaining repeated correction or premature confidence. Reliable execution requires checks that resolve the task's actual contact, support, and alignment constraints.

\noindent\textbf{Recovering from consequential actions.}
Each attempt changes the state from which recovery must proceed, and some changes cannot be undone (Fig.~\ref{fig:failure_gallery}(b)). In RoboDojo Pour liquid into cup, the cup tips over early in the episode and the liquid spills onto the table; GPT-6-Astra rights the cup and resumes pouring, but the spill cannot be reversed, and the episode ends with most of its budget unused. In DexJoCo Assembly, a peg slips from the grasp and falls to the floor; GPT-6-Astra judges it unreachable and parks the arms with most of the budget remaining. Recovery is also costly when the failed step is a prerequisite for later ones, such as opening a container or securing an object: until it succeeds, downstream subtasks cannot proceed, and repeated retries consume the budget without advancing the task. Such dependencies are especially prevalent in long-horizon benchmarks like BEHAVIOR and LHPR-VLN, where a single unresolved subtask can stall progress toward the overall goal. In LHPR-VLN, for example, an agent may become stuck trying to reach an intermediate destination, lose track of its navigation plan, and resort to aimless exploration without making meaningful progress. Additional budget therefore helps only when a recovery path still exists, and successful recovery usually involves a meaningful change of strategy rather than repetition. In DexJoCo Fold Glasses, for example, GPT-6-Astra switches to a side approach, verifies the grasp with a test lift, and completes the task after exceeding the standard budget.

\noindent\textbf{Preserving progress over long horizons.}
Long-horizon tasks require earlier progress to remain intact while later actions are executed, and small deviations compound across stages (Fig.~\ref{fig:failure_gallery}(c)). In RoboDojo Store laptop and headphones, GPT-6-Astra has already stored the laptop and hung the headphones on their stand, then returns to adjust one earcup; the correction pulls the headband off the stand, and the headphones are not re-hung before the budget runs out. In BEHAVIOR can\_meat, a jar that is already held beside the counter tilts during the next motion and falls to the floor, so progress has to be rebuilt before the task can continue. Beyond such direct disruptions, small errors in grasping, placement, or navigation accumulate over successive stages and alter the conditions for later actions, forcing GPT-6-Astra to re-localize and correct earlier mistakes. Reliable long-horizon execution therefore requires tracking and rechecking conditions established earlier, rather than treating completed subgoals as permanently resolved.

These observations motivate task-relevant verification, recovery strategies that account for irreversible consequences and prerequisite dependencies, and persistent tracking of earlier progress. They suggest that further progress depends on how effectively the agent uses additional interaction to revise its strategy and verify the resulting state.

\section{Harness  Ablation}
\label{sec:ablation}

\begin{figure*}[t]
\centering

\begin{minipage}[t]{0.61\textwidth}
  \vspace{0pt}
  \centering
  \includegraphics[width=\linewidth]
    {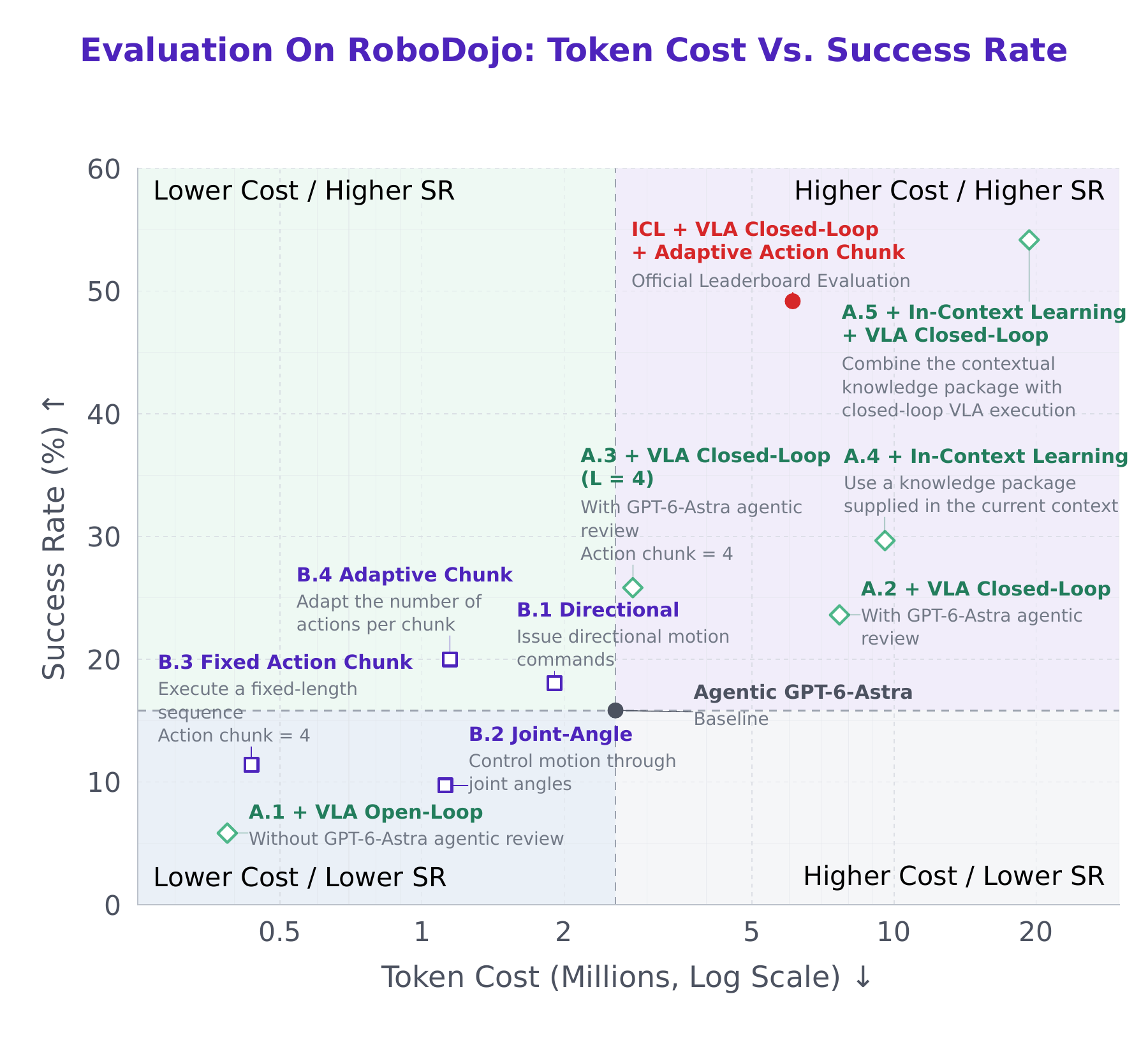}
\caption{Token cost versus performance on RoboDojo for different System~1--System~2 collaboration strategies and action interfaces. Each point represents an experimental configuration, with token cost measured in millions and performance measured by the unweighted mean of category success rates. Among these configurations, the red marker highlights the official RoboDojo leaderboard success rate of our submission, which combines in-context learning, closed-loop VLA execution, and adaptive action chunking. Dashed lines intersect at the Agentic GPT-6-Astra baseline reference point, defining four shaded quadrants.}
  \label{fig:harness_ablation}
\end{minipage}
\hfill
\begin{minipage}[t]{0.38\textwidth}
  \vspace{1.53cm}
  \centering
  \scriptsize
  \setlength{\tabcolsep}{5pt}
  \renewcommand{\arraystretch}{1.4}
  \resizebox{\linewidth}{!}{%
    \begin{tabular}{@{}lrr@{}}
    \toprule
    Setting & SR (\%) $\uparrow$ & Tokens (M) $\downarrow$ \\
    \midrule

    \rowcolor{sensepurple!15}
    \multicolumn{3}{l}{\textbf{A. System 1--System 2 Collaboration}} \\
    Baseline                         & 15.56 & 2.560  \\
    VLA ($\pi_{0.5}$)                & 6.91  & --     \\
    A.1 + VLA Open-loop                 & 5.83  & 0.387  \\
    A.2 + VLA Closed-loop               & 23.61 & 7.664  \\
    A.3 + VLA Closed-loop ($L=4$)        & 25.83 & 2.796  \\
    A.4 + ICL                           & 29.67 & 9.567  \\
    A.5 + ICL \& VLA Closed-loop         & \textbf{54.17} & 19.335 \\
    \midrule

    \rowcolor{sensepurple!15}
    \multicolumn{3}{l}{\textbf{B. Action Representation and Chunking}} \\
    Baseline                         & 15.56 & 2.560  \\
    B.1 Directional motion commands   & 18.06 & 1.908  \\
    B.2 Joint-angle                   & 9.72  & 1.120  \\
    B.3 fixed action chunk ($L=4$)     & 11.39 & 0.435  \\
    B.4 adaptive action chunk         & 20.00 & 1.146  \\

    \bottomrule
    \end{tabular}%
  }
\end{minipage}

\end{figure*}

Section~\ref{sec_icl_package} shows that additional interaction can substantially improve task completion, but its benefits depend on how effectively the agent uses that interaction. Longer execution incurs substantial inference costs and does not reliably resolve failures in physical verification, recovery, or preservation of subtask progress. These findings motivate examining the harness that connects GPT-6-Astra’s reasoning to observations, actions, and feedback.

We therefore conduct controlled ablations to investigate which harness designs improve task performance and at what inference cost. We study two aspects:   System 1–System 2 collaboration and action representations and chunking.

\subsection{System 1--System 2 Collaboration}
\label{sec:system_collaboration}
We use GPT-6-Astra as System 2 and $\pi_{0.5}$~\cite{pi2025pi05} fine-tuned on RoboDojo as System 1 (VLA). We compare three interaction settings: (i) standalone VLA, which executes actions without GPT supervision; (ii) open-loop collaboration, where GPT specifies subgoals and the VLA directly executes joint-action sequences; and (iii) closed-loop collaboration, where the VLA proposes actions that are converted into candidate EEF motions through forward kinematics, allowing GPT to accept, modify, or reject them before execution. For closed-loop collaboration, we vary the review horizon $L\in\{1,4\}$, where $L$ denotes the number of VLA-proposed EEF actions reviewed per GPT call. Finally, we evaluate in-context learning (ICL) following the setup described in Section~\ref{sec_icl_package}, both for the direct-control agent and in combination with closed-loop VLA collaboration ($L=2$).

As shown in Figure~\ref{fig:harness_ablation}, the standalone VLA achieves an average success rate of 6.91\%, indicating limited performance on this benchmark. Open-loop collaboration similarly achieves 5.83\% while using 0.387 million GPT-6-Astra tokens. In contrast, closed-loop collaboration, where System 2 reviews and refines VLA-proposed EEF actions, improves the success rate by up to 20.00 percentage points over open-loop execution. However, per-action review requires 7.664 million tokens. Increasing the review horizon to $L=4$ reduces token consumption by 63.5\% to 2.796 million tokens, while improving the success rate from 23.61\% to 25.83\%. Combining ICL with closed-loop VLA achieves the highest success rate of 54.17\%, exceeding the best non-ICL closed-loop configuration by 28.34 percentage points, but consumes 19.335 million tokens, approximately 6.9 times as many as that configuration.

These findings suggest that delegating execution to System 1 alone is insufficient for reliable task completion, while frequent System 2 supervision incurs substantial token consumption. Reviewing longer action sequences improves both performance and token efficiency relative to per-action review, highlighting the importance of the collaboration protocol. The configuration combining ICL with closed-loop VLA achieves substantially higher performance, although at a much larger token cost. \textbf{A key direction is therefore to make verification and recovery more selective and to incorporate task-relevant experience more efficiently, balancing reliable execution with token consumption.}

We build an agent that combines in-context learning (ICL), closed-loop VLA control, and adaptive action steps, achieving an average score of 57.10 and a success rate of 49.17\% on the RoboDojo leaderboard \footnote{\url{https://robodojo-benchmark.com/leaderboard/rollouts/SenseNova-RoboRSI?bench=sim}} as SenseNova-RoboRSI.

\subsection{Action Representation and Chunking}
\label{sec:action_interface}
We study action representations and execution horizons under the same direct-control framework. Our baseline follows the RoboRSI~\cite{opensensenova2026roborsi} agent framework, where GPT-6-Astra observes RGB images and robot states and generates one Cartesian end-effector (EEF) action per decision. For action representations, we compare absolute Cartesian EEF targets (baseline), directional motion commands that specify relative movement directions and distances, and direct joint-angle control. For execution horizons, we compare single-action execution ($L=1$), fixed action chunks ($L=4$), and adaptive chunking, where GPT-6-Astra selects $L\in\{1,2,3,4\}$. Here, $L$ denotes the number of EEF actions generated per GPT decision, and token consumption includes both input and output tokens.

As shown in Figure~\ref{fig:harness_ablation}, directional motion commands improve the average success rate by 2.50 percentage points, whereas direct joint-angle control leads to a 5.84-percentage-point drop. This suggests a potential representation gap between the embodied intelligence of multimodal models and low-level robot control spaces. Action chunking further reveals a trade-off between success rate and token consumption. Fixed action chunking ($L=4$) uses 0.435 million tokens but decreases the average success rate by 4.17 percentage points relative to the baseline. In contrast, adaptive action chunking improves the average success rate by 4.44 percentage points while using 1.146 million tokens. Thus, adaptive chunking achieves a higher success rate than fixed chunking at a higher token cost. These results indicate that both action representation and execution horizon influence the success rate and token consumption of embodied control.

These findings suggest that effectively translating embodied intelligence into physical actions requires interfaces aligned with the model's reasoning capabilities~\cite{chen2026showharness}. \textbf{A promising direction is to expose semantically meaningful action commands while delegating precise motion execution to robot controllers, with adaptive execution horizons to further balance task performance and token consumption.}
 
\section{Real-World Validation}
\label{sec:real_world}

\begin{table}[t]
\centering
\caption{\textbf{Real-world platforms, ordered from simple to complex embodiments.} Each class probes a different question; settings and per-task results are in App.~\ref{app:real_world}.}
\label{tab:rw_overview}
\footnotesize
\setlength{\tabcolsep}{5pt}
\renewcommand{\arraystretch}{1.0}
\newcommand{\rwthumb}[1]{\includegraphics[width=1.45cm]{figures/images/real_world/thumbs/#1}}
\newcommand{\rwrow}[7]{\raisebox{-0.5\height}{#1} &
  \parbox[c]{3.9cm}{\raggedright\textbf{#2}\\[0pt]{\scriptsize\color{black!60}#3}} &
  \textbf{#5} & \parbox[c]{5.0cm}{\raggedright #6} & \parbox[c]{\linewidth}{\raggedright\scriptsize #7}}
\newcommand{\rwsep}{\\[1pt]\arrayrulecolor{black!12}\specialrule{0.3pt}{1.5pt}{1.5pt}\arrayrulecolor{black}}
\begin{tabularx}{\textwidth}{@{}c l c l >{\raggedright\arraybackslash}X@{}}
\toprule
& \textbf{Embodiment} & \textbf{\#} & \textbf{Main question} & \textbf{Tasks} \\
\midrule
\rwrow{\rwthumb{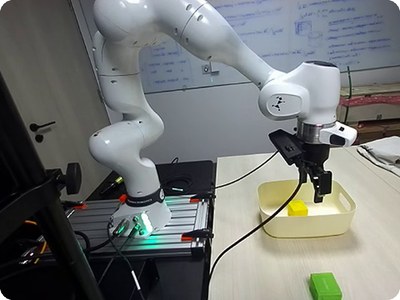}}{Single arm, gripper}{Franka 3, Franka Panda, PiPER + Pika gripper}{1}{3}{Do simulation results transfer, from rigid to deformable and precise tasks?}{Stacking and container nesting; cloth and towel folding; plug insertion} \rwsep
\rwrow{\rwthumb{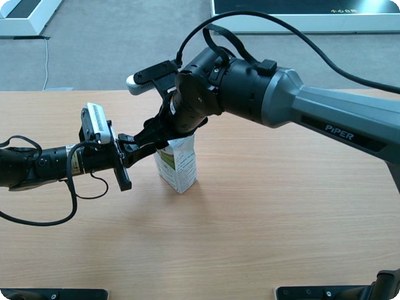}}{Dual arm, grippers}{AgileX Piper}{2}{1}{Can GPT-6-Astra coordinate two arms on harder zero-shot tasks?}{Ordered folding; flower insertion; bimanual holding; sorting} \rwsep
\rwrow{\rwthumb{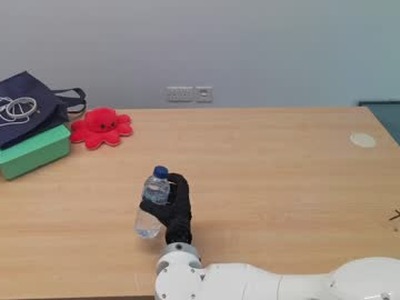}}{Dual arm, dexterous hands}{Tianji M6S}{3}{1}{Can GPT-6-Astra control high-DoF hands?}{Toy and bottle grasping} \rwsep
\rwrow{\rwthumb{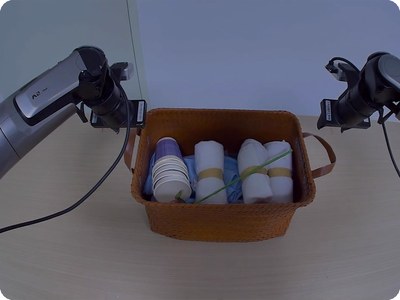}}{Wheeled mobile manipulator}{Galaxea R1 Pro, AgileX Tracer + xArm7}{4}{2}{Can GPT-6-Astra act over room-scale distances and long horizons?}{Find washer and press button; flower collection; tray arrangement; navigation and search} \rwsep
\rwrow{\rwthumb{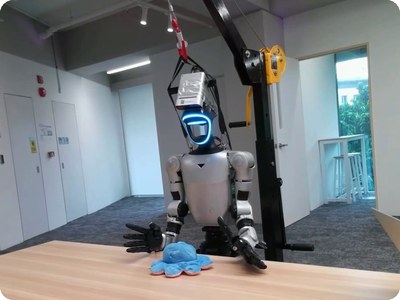}}{Humanoid}{Unitree G1}{5}{1}{Can GPT-6-Astra act as System~2 over low-level locomotion and arm control?}{Tabletop pick-and-drop; turn, approach and stop} \\[1pt]
\bottomrule
\end{tabularx}
\end{table}

%
%
%

Since the simulation protocol of Sec.~\ref{sec:study_protocol} rests on two assumptions that do not hold on physical robots, namely that successful demonstrations of the test task are available and that additional execution steps are free, we further study a \emph{real-world variant} of the protocol that revises both. First, the context changes: since new tasks often come without demonstrations in real-world deployment, GPT-6-Astra operates in a zero-shot setting and receives only structured information provided by a human as context. Second, the budget changes: the step limit is replaced by wall-clock time, because on real hardware the time a task takes determines not only token cost but also safety, hardware health and human labor.

\subsection{Setting: The Real-World Variant}
\label{sec:rw_setting}

We deploy GPT-6-Astra on eight platforms spanning five embodiment classes, ordered from simple to complex in Table~\ref{tab:rw_overview}. The real-world variant departs from the simulation protocol in three respects; per-platform settings, context and results are summarized in Table~\ref{tab:rw_platforms} and detailed in App.~\ref{app:real_world}.

\noindent\textbf{(1) Zero-shot.}
GPT-6-Astra receives no demonstration and no knowledge package distilled from data, and performs as a generalist, without task- or robot-specific training.

\noindent\textbf{(2) Structured human priors as context.}
In place of the knowledge package, GPT-6-Astra receives context written by a human. This context may encode information that is routinely available when a robot is deployed, such as tool schemas, joint limits, camera placement, safety constraints and open/close presets for dexterous hands, and, on some platforms, task-level guidance such as admissible grasp points. No platform exposes high-level semantic skills such as pick, place or insert.

\noindent\textbf{(3) Wall-clock budget and success criteria.}
Trials are budgeted in wall-clock time rather than in steps, normally 60 minutes; per-platform exceptions are listed in App.~\ref{app:real_world}. An operator holds an emergency stop throughout, and a person judges success from the physical outcome rather than from the model's own completion statement. For deformable tasks such as cloth folding, a single success criterion is hard to define across platforms, so their success rates indicate that the task was carried through rather than a common level of quality.

\subsection{Findings}
\label{sec:rw_findings}

\begin{figure}[t]
\centering
\includegraphics[width=\linewidth]{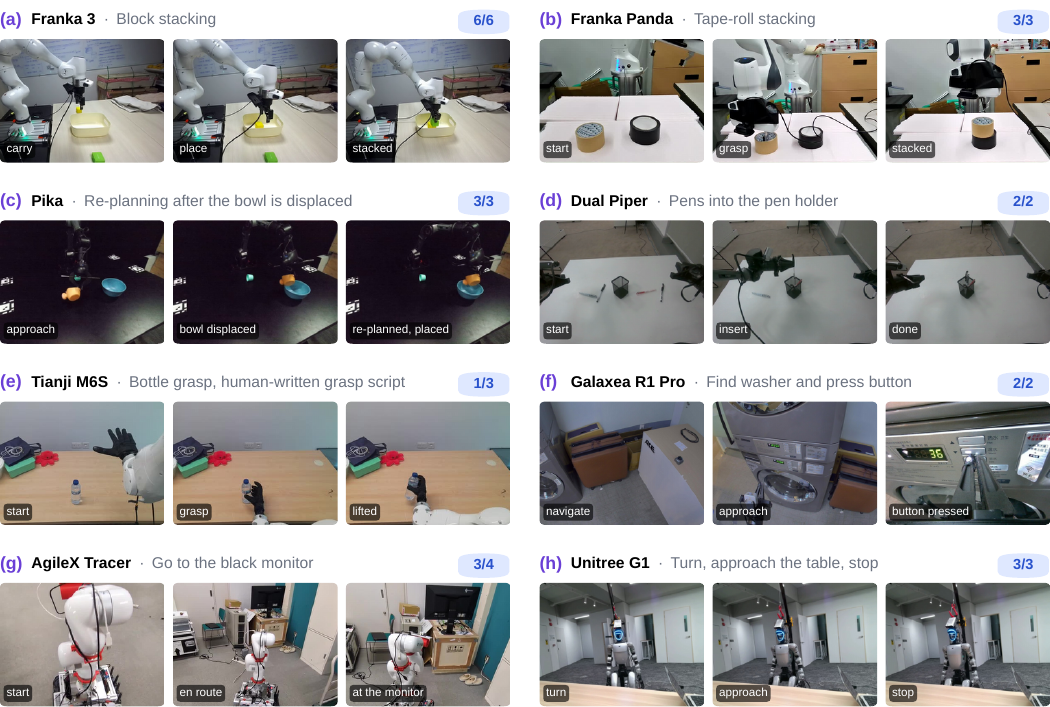}
\caption{\textbf{Zero-shot real-world control, from simple to complex embodiments.} Frames from released videos in temporal order. Chips give successes over all counted trials. (c) GPT-6-Astra re-plans after contact displaces the receiving bowl. (e) Dexterous grasping with the Wuji hands of the Tianji M6S, given human-written grasp scripts (the successful bottle run; the toy grasp succeeds in 3/3). The full gallery is in Fig.~\ref{fig:rw_gallery_full}, and all tasks per platform in App.~\ref{app:real_world}.}
\label{fig:rw_gallery}
\end{figure}

\noindent\textbf{Finding 1: one model drives every embodiment; success depends on the task.}
Without platform-specific training, the same model controls all eight platforms, from single arms to a humanoid (Fig.~\ref{fig:rw_gallery}). What limits success is the nature of the task: rigid pick-and-place is reliable (17/18 trials
) and matches simulation, whereas deformable objects (e.g., cloth folding: 2/4 on the Franka~3), precise contact (e.g., plug insertion: 0/11 across three platforms) and long-horizon mobile tasks (e.g., navigation and search: 9/23 on the AgileX Tracer) remain markedly harder. A second arm extends the range of zero-shot tasks GPT-6-Astra completes (Fig.~\ref{fig:rw_gallery_full}), and GPT-6-Astra re-plans from execution feedback when the scene changes (Fig.~\ref{fig:rw_gallery}c). Per-platform results are in App.~\ref{app:real_world}.

\begin{figure}[t]
\centering
\includegraphics[width=\linewidth]{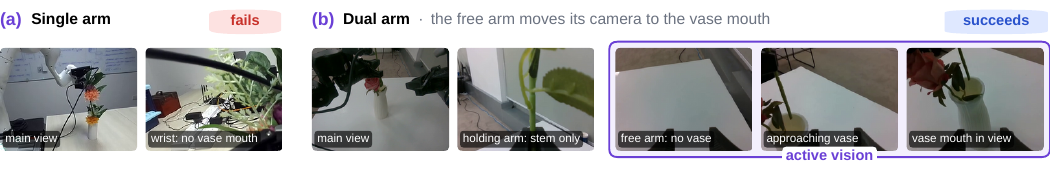}
\caption{\textbf{What GPT-6-Astra can do depends on what it can see.} Flower insertion, main view and wrist views. (a) Single arm (Franka~3): the gripper holds the stem vertically, so its wrist camera cannot see the vase mouth, and the insertion fails. (b) Dual-arm Piper: the wrist view of the holding arm shows only the stem; GPT-6-Astra moves the free arm until its wrist camera sees the vase mouth (boxed frames: no vase, approaching the vase, vase mouth in view), and the insertion succeeds.}
\label{fig:rw_active_vision}
\end{figure}

\noindent\textbf{Finding 2: what GPT-6-Astra can do depends on what it can see.}
Beyond the prompt and the available primitives, real-world success depends critically on the view configuration, that is, on whether the cameras available to the model can observe the region in which the task is decided. The flower insertion task illustrates this dependence. The gripper holds the stem vertically from above, so the wrist camera, which moves with the gripper, cannot see the mouth of the vase below the flower. On a single arm (Franka~3), the wrist camera is the only close-up view: GPT-6-Astra never sees the vase mouth during the insertion, and the attempt fails, with the flower dropped beside the vase (Fig.~\ref{fig:rw_active_vision}a). On the dual-arm Piper, the wrist view of the arm holding the flower is limited in the same way, but GPT-6-Astra moves the free arm until its wrist camera looks at the vase mouth, and then completes the insertion (Fig.~\ref{fig:rw_active_vision}b). The difference is \emph{active vision}: the viewpoint becomes part of the action space. This behavior is largely absent from teleoperated demonstrations: a human operator typically watches the scene from their own vantage point rather than through the robot's cameras, so demonstrations rarely move a camera to resolve an occlusion, and policies learned from them have little to imitate. A zero-shot agent that perceives only through the robot's cameras has to discover it.

\noindent\textbf{Finding 3: GPT-6-Astra as System~2 over existing System~0 motor control.}
On a humanoid, GPT-6-Astra deliberates far more slowly than local motor control runs, so we test whether it can act directly through existing System~0 controllers, without a task-trained System~1 policy. In SIMPLE, GPT-6-Astra issues joint, body and locomotion targets that decoupled WBC or AMO~\cite{li2025amo} controllers execute. On a real Unitree G1 with Wuji hands, it commands arm-joint targets and binary hand open/close states for manipulation (2/3), and turns and walks through the vendor velocity interface (3/3); the two tasks are evaluated separately (Fig.~\ref{fig:rw_gallery}h and Fig.~\ref{fig:rw_gallery_full}, B3). This extends the System~2/System~1 coordination of Sec.~\ref{sec:ablation} down to existing motor-control interfaces, but it establishes feasibility under slow, feedback-driven supervision rather than reliable real-time visuomotor control.

\subsection{What Changes in the Real World}
\label{sec:rw_challenges}
The two ingredients behind the simulation results in Sec.~\ref{sec:study} are each constrained on real-world setting.

\noindent\textbf{Physical constraints limit extended execution.}
Extending the step limit by up to 10$\times$ works in simulation because additional steps and failed attempts carry no consequence; the physical world imposes three constraints that simulation does not model.
(i)~\emph{Safety.} Contacts beyond the controller's limits trigger protective stops that usually end the trial, for instance when the gripper touches the table while folding thin cloth on the Franka~3, or when a plug-insertion attempt on the PiPER arm ends in a joint collision fault (2 of its 3 trials), and GPT-6-Astra cannot recover from such hardware-level faults. Some outcomes, such as a dropped object, are irreversible.
(ii)~\emph{Mechanical limits.} GPT-6-Astra has no direct feedback on the load its commands impose: on the Tianji M6S, finger joints of the Wuji hand stalled under load during repeated grasp attempts, and its only guard was a temperature threshold it had set for itself, which the hand reached (62\,$^{\circ}$C), interrupting the attempt. Actuators overheating can terminate prolonged trials: a torso over-temperature fault ended both flower-collection runs on the Galaxea R1 Pro.
(iii)~\emph{Real time.} In simulation the environment pauses while the model deliberates. On a physical robot, elapsed time is the quantity through which every other cost accumulates: tokens, exposure to safety risks, hardware stress and human supervision. Episodes are long: on the G1, GPT-6-Astra spends about 90\% of each run deliberating, and a single manipulation trial takes over an hour.
The trade-off between longer execution and higher success that GPT-6-Astra exploits in simulation is therefore not freely available on real hardware (App.~\ref{app:real_world}).

\noindent\textbf{Missing demonstrations reshape in-context learning.}
The ICL protocol of Sec.~\ref{sec:study_protocol} presumes that successful demonstrations of the test task, on the test robot, already exist. In real-world deployment new tasks often come without them. Human-written context suffices for a meaningful range of tasks, but it is still prompt engineering by an expert and its quality directly affects the outcome: on the Tianji hands, without predefined grasp primitives GPT-6-Astra did not complete a single grasp (0/3), whereas with human-written grasp primitives it succeeded in 3/3. This context also need not come from a person: on the Galaxea R1 Pro, GPT-6-Astra wrote its own robot profile in a motion-free familiarization session, from vendor documentation and live probes of sensors and controllers, and every task session starts from it. Human-written context represents only one point in a broader design space. Between a complete set of in-domain demonstrations and none at all, a range of alternative context sources may be available at deployment time, which we have not yet evaluated: cross-embodiment demonstrations, self-generated demonstrations in simulation, human demonstrations, the robot's own deployment history, and knowledge packages shared across robots. The principled construction of context for embodied tasks from such heterogeneous sources remains an open problem, and constitutes a substantial research opportunity for in-context learning in real-world deployment.

\section{Towards a Next-Generation Embodied Benchmark}
\label{sec:benchmark}

The preceding sections show that GPT-6-Astra's results on established embodied benchmarks depend heavily on conditions that the benchmarks grant. Its success rises with in-context access to the benchmark's demonstrations and with an extended step budget in which failures cost nothing (Sec.~\ref{sec:study}), and depends on how a general agent is adapted to an action interface built for a learned policy (Sec.~\ref{sec:ablation}). These conditions follow from how most embodied benchmarks were designed, for policies trained by imitation: demonstrations of a task are collected, a policy is fitted to them, and it is evaluated on held-out episodes of the same task~\cite{liu2023libero,nasiriany2026robocasa365,chen2026robodojo,behavior2026challenge}. On real robots, by contrast, neither demonstrations of a new task nor extra time comes for free (Sec.~\ref{sec:real_world}). For a general-purpose agent, such benchmarks therefore measure in part how well it exploits what they provide, and raising their difficulty, the usual response once benchmarks saturate~\cite{kiela2021dynabench,phan2025hle}, would not change this. We instead propose three principles that change how embodied benchmarks are designed and judged, and illustrate them with a household proof of concept.

\subsection{Principles for Next-Generation Benchmarks}
\label{sec:principles}
Following the protocol now common for language agents, in which an agent acts in an environment by any means and is scored on the outcome~\cite{jimenez2024swebench,xie2024osworld,merrill2026terminalbench,shen2026embodiedswe}, we target a setting in which the agent perceives the world only through its robot's sensors, acts through the commands its hardware exposes, and is evaluated on a single deployment in a world that keeps evolving; everything it builds above that interface is part of the system under test. Current embodied benchmarks judge success as a planning domain language would: BDDL~\cite{srivastava2021behavior}, inspired by PDDL~\cite{mcdermott1998pddl}, states a task as a goal condition over the scene's objects that must hold in a single state (\cref{fig:breakfast_spec}, left). Such goals inherit the assumptions of classical planning~\cite{ghallab2004automated,russell2021aima}: a fully specified goal with binary success, a world that changes only through the agent's actions, time counted in steps, and a given world model. Extensions of PDDL relax some of these~\cite{fox2006pddlplus,gerevini2009pddl3,hoeller2020hddl}, but embodied benchmarks have rarely adopted them, and each principle below breaks at least one.

\newsavebox{\bddlbox}\newsavebox{\judgebox}\newlength{\specht}
\begin{figure}[t]
\centering
\begin{lrbox}{\bddlbox}\begin{minipage}{0.465\linewidth}
\begin{lstlisting}[basicstyle=\ttfamily\scriptsize,columns=fullflexible,frame=none,keepspaces=true,aboveskip=0pt,belowskip=0pt]
(define (problem cook_bacon-0)
  (:domain behavior-1k)
  (:objects
    bacon.n.01_1 ... bacon.n.01_6 - bacon.n.01
    tray.n.01_1 - tray.n.01
    electric_refrigerator.n.01_1 - ...
    stove.n.01_1 - stove.n.01
    frying_pan.n.01_1 - frying_pan.n.01 ...)
  (:init
    (ontop bacon.n.01_1 tray.n.01_1) ...
    (not (cooked bacon.n.01_1)) ...
    (inside tray.n.01_1
            electric_refrigerator.n.01_1)
    (ontop frying_pan.n.01_1 stove.n.01_1) ...)
  (:goal
    (and
      (forall
        (?bacon.n.01 - bacon.n.01)
        (cooked ?bacon.n.01))
      (not
        (open ?electric_refrigerator.n.01_1)))))
\end{lstlisting}
\end{minipage}\end{lrbox}%
\begin{lrbox}{\judgebox}\begin{minipage}{0.465\linewidth}
\begin{lstlisting}[basicstyle=\ttfamily\scriptsize,columns=fullflexible,frame=none,keepspaces=true,aboveskip=0pt,belowskip=0pt]
(define (request make_breakfast)
  (:say "Could you make me breakfast?")
  (:context (resident-leaves-at 15 min))
  (:clock world)  ; deliberation advances it
  (:observe until (+ done 5 min))
  (:acceptable              ; any one, graded
    (toast)                         0.5
    (egg toast)                     0.8
    (egg toast coffee)              1.0)
  (:value  1 if served by 15 min,
           0 after 30 min, linear between)
  (:hidden                 ; never shown to agent
    (cooked x) = (core-temp x >= 70 C)
    (burnt x)  = (surface-temp x >= 200 C))
  (:never  (on_fire) (touches robot person))
  (:whenever (open fridge)
     -> (closed fridge) within 1 min)
  (:whenever (on burner) (empty burner)
     -> (off burner) within 2 min)
  (:whenever (water floor)
     -> (dry floor) within 5 min))
\end{lstlisting}
\end{minipage}\end{lrbox}%
\setlength{\specht}{\dimexpr\ht\bddlbox+\dp\bddlbox\relax}%
\ifdim\dimexpr\ht\judgebox+\dp\judgebox\relax>\specht\setlength{\specht}{\dimexpr\ht\judgebox+\dp\judgebox\relax}\fi
\fbox{\begin{minipage}[c][\specht][t]{0.465\linewidth}\usebox{\bddlbox}\end{minipage}}\hfill
\fbox{\begin{minipage}[c][\specht][t]{0.465\linewidth}\usebox{\judgebox}\end{minipage}}
\caption{\textbf{Two specifications of a breakfast task.} Left: the BEHAVIOR-1K activity \texttt{cook\_bacon}~\cite{behavior2026challenge}, with its complete goal. The goal is tested against the current state at every step, and the episode succeeds at the first step at which it holds. Nothing requires the stove to be turned off, and since \texttt{cooked} holds once an object's maximum temperature has passed a threshold, it remains true for bacon that has burned. Right: what a judge for the request ``make breakfast'' needs to express. Any acceptable refinement earns credit, graded by what is served and by when, with a value that decays once the resident has left~\cite{haddawy1998utility}; items count as served only if they never burned. Predicates are defined on physical quantities known only to the judge. The \texttt{:never} and \texttt{:whenever} clauses are the absence~\cite{dwyer1999patterns} and bounded-response~\cite{konrad2005realtime} specification patterns, expressible in metric~\cite{koymans1990mtl} and signal~\cite{maler2004stl} temporal logic and checked over the whole trajectory on the world clock, which keeps running after the agent declares that it is done. All constants are illustrative.}
\label{fig:breakfast_spec}
\end{figure}

\paragraph{Tasks derived from human needs.}
Benchmark tasks are usually ones on which a robot can be shown to succeed, because demonstrations must be collected for them; even generalist policies that operate zero-shot in unseen homes~\cite{pi2025pi05} are evaluated on tasks chosen from the robot's perspective. A deployed robot must instead do what the people it serves need done, whether or not any robot has done it before. Following BEHAVIOR-1K~\cite{li2024behavior1k}, we propose that tasks be derived from such needs and stated as the person would state them. Such tasks also test whether the agent knows the limits of its embodiment: a humanoid may fetch a step stool to reach a ceiling bulb, whereas a wheeled base should ask for help. Real needs also carry obligations that no instruction mentions, such as an unattended stove, a running tap or a cabinet the agent itself left open. These have been studied in narrower forms, as tidying without instructions~\cite{kant2022housekeep,sarch2022tidee}, mitigating risks the agent itself creates~\cite{lu2026isbench} and rescuing objects from a spreading hazard~\cite{zhou2024hazard}; in deployment they also arise while the robot is busy with something else, and the agent must notice them unprompted and decide whether to interrupt its work. We therefore place these obligations (Duty) in the same deployment as the explicit request (Intent), with deadlines on the world clock.

A request stated as a need cannot be judged as a goal. ``Make breakfast'' resembles a high-level action in hierarchical planning~\cite{erol1994htn,russell2021aima}: it admits many refinements, from a slice of toast to a cooked meal, of which a goal formula keeps only the one its author chose (\cref{fig:breakfast_spec}). Hierarchical languages accept any decomposition of the task into an executable plan~\cite{hoeller2020hddl}, yet the refinements serve the person to different degrees, which calls for a utility over outcomes~\cite{keeney1993decisions} rather than the fraction of satisfied goal predicates reported as partial credit~\cite{srivastava2021behavior,behavior2026challenge}. A fixed check can even contradict its own instruction: Bench2Dex's condiment-box-loading task~\cite{yang2026bench2dex} asks the agent to place three condiments into a box, but its success check also requires the soy-sauce and vinegar bottles to stand upright, which the instruction never states. Because a capable agent may also satisfy a request in ways a fixed check did not anticipate, or exploit a check weaker than the request~\cite{krakovna2020specification}, the judge should itself be validated against how people rank alternative outcomes~\cite{christiano2017deep,zheng2023judging}.

\paragraph{Experience acquired at test time.}
Demonstrations are often unavailable for new tasks in deployment, and they also bound what an agent learns: recorded from the operator's viewpoint, they rarely move a camera to resolve an occlusion, whereas an agent that understands its embodiment can make its viewpoint part of the action (Sec.~\ref{sec:real_world}). We therefore propose that the benchmark provide no in-domain training episodes or scene-specific demonstrations, so that the agent must ground its general priors in an unseen home and embodiment through its own exploration, mapping the layout, testing affordances and figuring out idiosyncratic appliances, and may find strategies that no demonstrator used. Planning languages presume the opposite, a given world model~\cite{russell2021aima}, and benchmark goals retain a trace of it: predicates such as \texttt{cooked} or \texttt{toggled\_on} tell the agent which variables matter, whereas a real air fryer has only a winding timer and a toaster's dial yields a continuum of results. The judge should therefore define its predicates on physical quantities kept private from the agent (\cref{fig:breakfast_spec}, right), and charge exploration the same world time and side effects as any other action. Since experience carries over between requests within a deployment and is reset between deployments, the unit of assessment is the deployment, and how quickly the agent becomes competent in it is a quantity of interest~\cite{chollet2019measure,bauer2023ada}.

\paragraph{A world that keeps running.}
An extended step budget raises GPT-6-Astra's success at no cost in simulation, because the simulator pauses while it deliberates: on RoboDojo, the standard budgets correspond to 8--76\,s of commanded motion against an average of 55 minutes of computation (Sec.~\ref{sec_simulation_results}). On a physical robot that time passes in the world, and on the G1 GPT-6-Astra spends about 90\% of each run deliberating (Sec.~\ref{sec:rw_challenges}). Following recent evaluations in which the environment keeps evolving while the agent reasons~\cite{zheng2025rrara,wen2026realtime}, we propose that the world not wait for the agent: the time between an observation and the next command, whether spent on inference, tool use or communication, is simulated under the previous command, as on a physical robot~\cite{black2025rtc}, while the host's simulation and rendering time is not charged. The judge must then observe the whole trajectory on the world clock, including some time after the agent declares that it is done, since a final-state goal cannot see an egg that was ready and then burned. Obligations become constraints on the trajectory, conditions to be avoided throughout or restored within a deadline once triggered~\cite{dwyer1999patterns,konrad2005realtime}; PDDL3 states such constraints for plans~\cite{gerevini2009pddl3}, and PARTNR and IS-Bench check ordering and triggered safety conditions in embodied tasks~\cite{chang2025partnr,lu2026isbench}, but without deadlines. Time also sets the value of an outcome: a resident leaving in fifteen minutes is better served by eggs and toast in ten than by a full breakfast in an hour. The judge should therefore assign a utility that depends jointly on what was achieved and when~\cite{haddawy1998utility} (\cref{fig:breakfast_spec}, right), which makes the agent's thinking time part of its performance.

\subsection{A Household Proof of Concept}
\label{sec:proof_of_concept}
\begin{figure}[t]
  \centering
  \includegraphics[width=\textwidth]{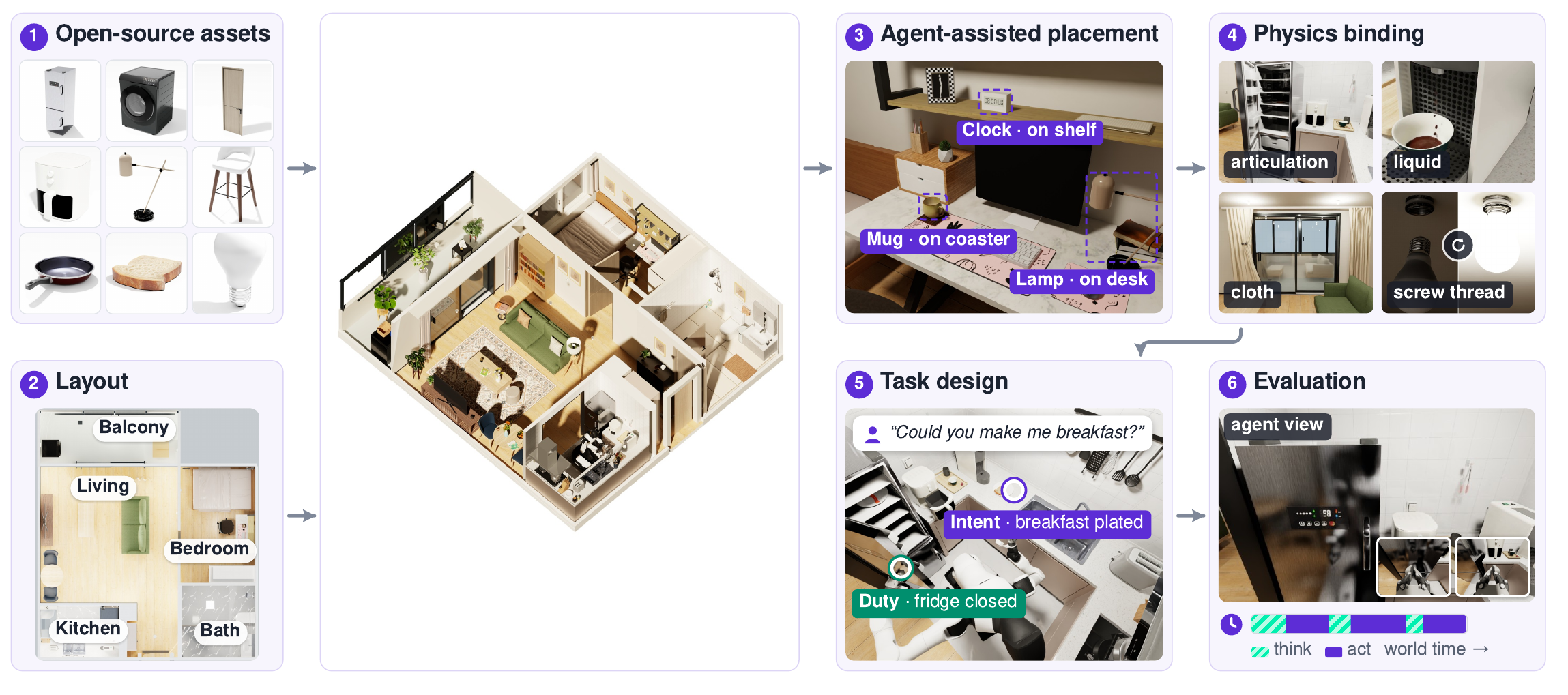}
  \caption{\textbf{Building and using the household environment.} Publicly available assets (1) and a floor plan (2) are composed into an interactive home. Coding agents place everyday items at real-world scale on named supports under human review (3), and every part is bound to physics: articulation, liquids, cloth and contact (4). A task states only a need (5): asked to make breakfast, the agent must work out the fridge, air fryer and toaster by itself, and is to be judged both on the request (Intent) and on obligations no one stated, such as closing the fridge (Duty). The agent sees only through the robot's onboard cameras, and its deliberation advances world time (6).}
  \label{fig:ele_pipeline}
\end{figure}

We ground these principles in the household. Requests there are open-ended, chores chain many steps across rooms and appliances, much of the work is contact-rich, and the home keeps changing while the robot works, which are the conditions under which GPT-6-Astra still falls short (Secs.~\ref{sec:study} and~\ref{sec:real_world}). We built a fully interactive one-bedroom apartment in NVIDIA Isaac Sim~\cite{nvidia2025isaacsim} through a reproducible pipeline (\cref{fig:ele_pipeline}): publicly available assets~\cite{yin2026geniesim,jin2025artvip,li2024behavior1k,shen2026embodiedswe} and a floor plan form the home, coding agents furnish it at real-world scale under human review, and every part is bound to physics, including articulation, contact, position-based fluids and cloth~\cite{muller2007pbd,macklin2013pbf}, and the heating and cooking processes of everyday appliances. Because judged outcomes are only as meaningful as the physics behind them, the environment is built for causal fidelity: whatever an agent's decisions depend on, from an object's mass and friction to heat flowing from burner to pan to food, should be simulated with the right cause and effect and grounded in real-world references.

\begin{figure}[!t]
\centering
\includegraphics[width=\linewidth]{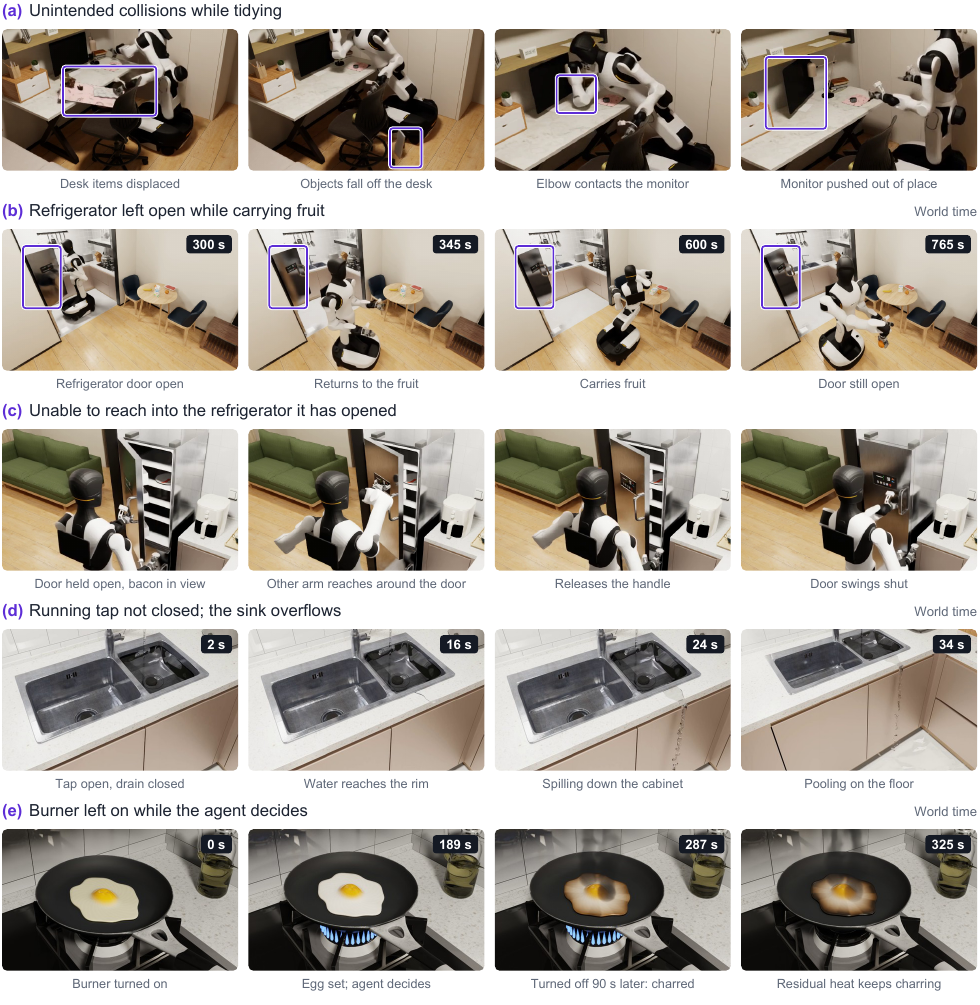}
\caption{\textbf{GPT-6-Astra in the household proof of concept.} All runs are zero-shot, with contact-based grasping and no demonstrations. (a)~While moving a book during tidying, the robot displaces desk items, knocks objects onto the floor and pushes the monitor out of place with its elbow. (b)~Having opened the refrigerator to store the milk, the robot turns to carrying fruit to the table, and the door stays open from 300\,s to 765\,s of world time. (c)~Making breakfast, the robot holds the refrigerator door open with one arm, but from where it stands the door blocks its other arm from the bacon inside; when it releases the handle to reach in, the door swings shut. (d)~The running tap is not closed, and with the drain closed the sink fills until water spills down the cabinet onto the floor. (e)~GPT-6-Astra watches an egg through the robot's head camera and decides to turn the burner off once it has set; deciding takes 8\,s of world time with the burner still on, and by the time the burner is turned off, about 90\,s after the egg has set, the egg has charred. Purple outlines highlight relevant objects; timestamps denote elapsed world time. External camera views are used for visualization; the agent sees only through the robot's onboard cameras.}
\label{fig:household_failures}
\vspace{5mm}
\end{figure}

Tasks in the apartment state only a need, such as making breakfast before the resident leaves (\cref{fig:ele_pipeline}, 5--6). We ran GPT-6-Astra on such tasks in the zero-shot setting of Sec.~\ref{sec:real_world}: it controls an AgiBot G2~\cite{yin2026geniesim} through the robot's own interface, observes only the onboard cameras and proprioception, and receives no demonstrations, only a description of the embodiment. The question is whether a model that comes close to saturating the benchmarks of Sec.~\ref{sec:study} meets the demands of Sec.~\ref{sec:principles} once the benchmark supplies nothing. \cref{fig:household_failures} shows five representative failures.

\noindent\textbf{Finding 1: GPT-6-Astra attends to the object in hand and loses track of the rest of the home.}
Tasks derived from human needs ask the agent to serve the whole request, including obligations that no one states; GPT-6-Astra serves only the step in front of it. Tidying a desk, it moves the book in its gripper while its arm knocks other items to the floor and pushes the monitor out of place (\cref{fig:household_failures}a). Putting groceries away, it opens the refrigerator to store the milk, then turns to carrying fruit to the table and leaves the door open from 300\,s to 765\,s of world time (\cref{fig:household_failures}b). In both episodes, the rest of the home, including the robot's own arm and the states it has changed, seems to enter the model's decisions only when it blocks the next step, much as on the real robots of Sec.~\ref{sec:rw_challenges}. A check on the final state would miss both: the refrigerator is closed by the end of the episode, and where the book ends up says nothing about the rest of the desk. The judge in \cref{fig:breakfast_spec} therefore tracks obligations over the whole trajectory rather than only at the end.

\noindent\textbf{Finding 2: Without demonstrations, GPT-6-Astra does not work out how unfamiliar fixtures move or where its own body has to be.}
With no demonstration to follow, the agent has to discover both through its own interaction. Making breakfast, GPT-6-Astra has to take bacon from the refrigerator (\cref{fig:household_failures}c). It pulls the door open and holds it with one arm, but from where its base stands in the narrow aisle, the open door lies between its free arm and the shelf, and the other arm can only reach around it. Getting to the bacon would require placing the base so that the swinging door and the robot's own body both fit between the refrigerator and the counter behind it. GPT-6-Astra plans for the gripper and the handle but not for the rest of its body: it lets go of the handle to reach in, the door swings shut, and it is back where it started. Faced with an unfamiliar tap, it likewise keeps pushing the handle instead of turning it while the sink fills and overflows (\cref{fig:household_failures}d). In both cases it repeats and adjusts the same motion. On the benchmarks of Sec.~\ref{sec:study}, the demonstrations it distills in context show both, so this weakness does not appear there.

\noindent\textbf{Finding 3: GPT-6-Astra bases decisions on observations that become outdated while it thinks.}
In a world that keeps running, an action must suit the state of the world when it takes effect. Watching an egg through the head camera, GPT-6-Astra correctly decides to turn the burner off once the egg has set, but each decision takes seconds of world time with the burner still on, and by the time the burner is off the egg has charred (\cref{fig:household_failures}e). The model reasons as though the world waits for it, which in the simulators of Sec.~\ref{sec:study} it does.

These episodes are few and qualitative, but they point the same way. On the benchmarks of Sec.~\ref{sec:study}, GPT-6-Astra's ability to decide what to do is enough. In a home that states only needs, offers no demonstrations and does not pause, it must also act on that decision, and this is where it currently falls short.

\subsection{Open Challenges}
\label{sec:open_challenges}
Realizing these principles in a full benchmark raises several open problems, which building our prototype made concrete. The first challenge is evidence for the judge. To score what happened rather than check a list of predicates, the judge needs physical facts, such as whether an egg was flipped or burned, which simulators still model with heuristic state changes~\cite{li2024behavior1k}. Learned judges do not yet fill this gap: vision-language success detectors reach at most 60\% balanced accuracy on contact-rich assembly~\cite{navasardyan2026failbench}, and vision-language judges of agent trajectories can be misled by the agent's own reasoning~\cite{khalifa2026gaming}. The second is simulation fidelity. Because the judge reads the simulator's state, a modeling error looks the same as agent behavior: a wrongly modeled joint can make an agent appear to succeed, and a contact artifact can make it appear to fail, and simulation has so far been shown to predict real-robot results for short tabletop tasks, including a few with deformable objects~\cite{li2024simpler,jain2025polaris,zhang2025softbody}, but not for long-horizon household activities. The third is the cost of a world that does not pause. If the world keeps running while the agent thinks, every physical process must be simulated during inference, and scores then depend on latency, hardware and service load; existing benchmarks avoid this by pausing the environment~\cite{zhang2025videogamebench}, which removes inference speed from the score, or by advancing time per generated token~\cite{wen2026realtime}, which is reproducible but does not reflect deployed latency. Finally, a deployment lasting an hour or more is expensive to repeat, while even short robot trials are typically run too few times to separate policies reliably~\cite{snyder2025step}.

\section{Conclusion}
\label{sec:conclusion}
In conclusion, GPT-6-Astra represents a clear leap in embodied intelligence, but not its endpoint. Our study shows that strong harness design with in-context learning and sufficient execution budget can push many isolated benchmarks close to saturation, while the remaining challenges shift toward transfer across tasks, long-horizon reasoning, failure recovery, and real-world constraints on time, cost, and safety. These findings also suggest that benchmark success should be interpreted with increasing care: the next frontier is not simply to solve more isolated tasks, but to build embodied systems that can operate reliably and efficiently under realistic deployment conditions.

\clearpage

\bibliographystyle{plainnat}
\bibliography{main}

\clearpage


\beginappendix
\section{Contributors}
\label{sec:contributors}

\textbf{Project Leads:} 
Zhongang Cai*, 
Zhonghua Wu*, 
Qingyi Tao*.

\textbf{Simulation Benchmarks:}
Quanzhou Li*, 
Han Chong Chua*,
Ruisi Wang*,
Xiaoming Wu*,
Tianqi Liu*,
Bo Liu*,
Pengfei Liu*,
Jing Zuo*,
Hongcan Xiao*,
Qinyue Li.

\textbf{Harness Framework:}
Weiye Xu*,
Yuyao Shao*,
Silei Wu*,
Jiapeng Luo*,
Hanming Deng*,
Tailai Chen*,
Linjun Dai*,
Ziyi Shang*,
Xiaowan Li,
Donghao Wu,
Tianbo Zhong,
Chenjing Ding.

\textbf{Real-World Validation:}
Tianqi Liu*,
Ziang Cao*,
Hanfei Zhao*,
Jiaxu Wu*,
Yuexiao Ma*,
Xiaoming Wu*,
Ziyi Wang*,
Weichen Fan*,
Zexin He*,
Zhi Hou*,
Beichen Wen,
Zilong Lin,
Long Zhuo,
Zhichao Liao,
Dawei Su.

\textbf{Benchmark Prototype:}
Zerui Gong*,
Zhe Yang*,
Oscar Qian*,
Baiqiao Yin*.

\textbf{Analysis \& Presentation:}
Wanqi Yin*,
Chenyang Gu.

\textbf{Collaborators:}
Hokin Deng,
Haiwen Diao,
Xiao Fu,
Bo Li,
Chunxiao Liu,
Liang Pan,
Quan Wang, 
Felix Juefei-Xu.

\textbf{Advisory Board:}
Lei Yang*,
Chen Change Loy*,
Lewei Lu*,
Ziwei Liu*,
Dahua Lin*,
Atsushi Yamashita,
Xingang Pan,
Dacheng Tao,
Xiaogang Wang.

* Core Contributors.

\section{Related Work}
\label{sec:related}

\subsection{Foundation Models as Embodied Agents}
\label{sec:related_agents}

Large language and vision-language models have long been used to bring semantic knowledge into robot control, but mostly at the level of high-level reasoning. SayCan~\cite{ahn2022saycan} grounds an LLM's proposals in the affordances of a pretrained skill library, and Inner Monologue~\cite{huang2022inner} closes the loop by feeding success detection and scene descriptions back to the planner. Code as Policies~\cite{liang2023code} prompts the model to write programs that call perception and control APIs, while VoxPoser~\cite{huang2023voxposer} and ReKep~\cite{huang2024rekep} have it compose 3D value maps or relational keypoint constraints that a motion planner then optimizes. Gemini Robotics-ER~\cite{team2025geminirobotics} extends this line with a model trained specifically for embodied reasoning. In all of these systems the foundation model specifies \textit{what} to do, while a predefined skill set, planner or separately trained controller decides \textit{how}. We instead ask whether a frontier model can act as the controller itself, emitting end-effector actions in closed loop from its own observations.

A second line of work, which has grown rapidly in 2026, borrows the harness paradigm of coding agents~\cite{jimenez2024swebench,merrill2026terminalbench} and holds that embodied performance depends as much on the infrastructure around the model as on the model itself. Some harnesses redesign the action interface: RoboDawn~\cite{guo2026robodawn} and Show-Harness~\cite{chen2026showharness} expose discrete, semantically meaningful motion commands that a frozen VLM can issue directly, and Robo-Harness K1~\cite{li2026roboharnessk1} equips the model with calibrated perception tools such as depth, object tracking and grasp hypotheses. Others focus on context and verification: Thea~\cite{wang2026thea} maintains a scene graph as persistent context and treats evaluation outcomes as exit codes, and Zetta~\cite{ding2026zetta} evolves code-based runtime critics and recovery skills around a frozen base policy. Guava~\cite{liu2026guava} distills a frontier-model harness into a compact open model, while RHO~\cite{elmaaroufi2026rho}, HARBOR~\cite{li2026harbor} and EmbodiedSWE~\cite{shen2026embodiedswe} use coding agents offline to build policy repositories, automate reinforcement-learning pipelines, or write programs against the full simulator state. Related studies examine how agents adapt from in-context experience~\cite{bauer2023ada,wagenmaker2025behavioral} and how reasoning latency interacts with a changing environment~\cite{wen2026realtime,zheng2025rrara}. Our ablation in Sec.~\ref{sec:ablation} complements these system-level proposals by isolating the contribution of individual harness components, namely the action interface, subagent roles and memory, under a common protocol.

\subsection{Robot Policy Learning}
\label{sec:related_policy}

The dominant paradigm for robot control learns a policy from demonstrations. Early large-scale efforts such as BC-Z~\cite{jang2022bcz} and RT-1~\cite{brohan2023rt1} trained language-conditioned policies on large collections of teleoperated episodes, while Implicit BC~\cite{florence2022implicit}, BeT~\cite{shafiullah2022bet}, ACT~\cite{zhao2023act} and Diffusion Policy~\cite{chi2023diffusion} introduced action representations better suited to multimodal, high-precision behavior, and 3D Diffusion Policy~\cite{ze2024dp3} further conditions on point clouds. These policies are typically trained and evaluated per task, and most established manipulation benchmarks, including several used in this report, were designed around this setting (Sec.~\ref{sec:benchmark}).

Vision-language-action models (VLAs) inherit semantic knowledge from web-scale pretraining by fine-tuning a vision-language model to output actions. RT-2~\cite{zitkovich2023rt} first demonstrated this transfer, OpenVLA~\cite{kim2024openvla} provided an open-source counterpart, and Octo~\cite{octo2024} and RDT~\cite{liu2025rdt} trained generalist policies on cross-embodiment data such as Open X-Embodiment~\cite{oxe2024} and DROID~\cite{khazatsky2024droid}. The $\pi$ series~\cite{black2024pi0,pertsch2025fast,pi2025pi05} pairs a VLM backbone with flow-matching or tokenized action experts and reaches open-world generalization in unseen homes, while GR00T N1~\cite{bjorck2025gr00t} and Gemini Robotics~\cite{team2025geminirobotics} extend the recipe to humanoids and to frontier-scale backbones. Recent systems such as Xiaomi-Robotics-0~\cite{cai2026xiaomi} and MolmoAct2~\cite{fang2026molmoact2} target real-time execution and deployable action reasoning. Grounding language in physical action nonetheless remains data-hungry, which has motivated the alternatives below.

World action models (WAMs) start instead from video or world-model backbones and predict how the scene will evolve together with the actions that produce it~\cite{wang2026wamsurvey,shen2026wamsurvey}. UniPi~\cite{du2023unipi} generates videos and recovers actions with an inverse-dynamics model, GR-1 and GR-2~\cite{wu2024gr1,cheang2024gr2} pretrain on large-scale video, and Seer~\cite{tian2025seer}, VPP~\cite{hu2025vpp}, UVA~\cite{li2025uva} and UWM~\cite{zhu2025uwm} couple future prediction and action generation more tightly through predictive representations or joint diffusion. In 2026 this family has scaled rapidly. Cosmos Policy~\cite{kim2026cosmospolicy} and LingBot-VA~\cite{li2026lingbotva} turn large video foundation models into policies, Fast-WAM~\cite{yuan2026fastwam} finds that most of the benefit comes from video modeling during training rather than from imagining the future at test time, and DreamZero~\cite{ye2026dreamzero} reports that WAMs generalize zero-shot to new tasks and environments substantially better than VLAs. A robustness study on LIBERO-Plus and RoboTwin~2.0-Plus likewise finds WAMs more robust to visual and language perturbations~\cite{zhang2026wamrobust}.

The most recent systems blur the boundary between the two families. Motus~\cite{bi2025motus} and BagelVLA~\cite{hu2026bagelvla} train unified models in which understanding, video-generation and action experts share attention, VLA-JEPA~\cite{sun2026vlajepa} augments a VLA with a latent world model, and $\pi_{0.7}$~\cite{pi2026pi07} conditions its action expert on subgoal images produced by a world model. Despite their differences, all of these approaches learn a policy from robot data. We ask the complementary question of how far a general-purpose frontier model can go as a policy without benchmark-specific policy training, and our harness ablation (Sec.~\ref{sec:ablation}) also invokes a VLA ($\pi_{0.5}$) as a callable tool.

\subsection{Embodied Benchmarks}
\label{sec:related_bench}

LIBERO~\cite{liu2023libero} remains the most widely used simulation suite for language-conditioned manipulation. LIBERO-Plus~\cite{fei2025liberoplus} and LIBERO-PRO~\cite{zhou2025liberopro} perturb its scenes and instructions to expose brittleness and memorization, and LIBERO-Long++ and LIBERO-Ultra-Long~\cite{yang2026lilo} stress visual clutter and chains of 9 to 16 skills. SIMPLER~\cite{li2024simpler} evaluates real-world policies in matched simulation, RoboCasa365~\cite{nasiriany2026robocasa365} scales simulated household tasks for training and benchmarking generalist robots, RoboDojo~\cite{chen2026robodojo} unifies simulated and real-robot evaluation of generalist policies, and RoboLab~\cite{yang2026robolab} adds high-fidelity scenes with controlled perturbations. For dexterous control, Bench2Dex~\cite{yang2026bench2dex} benchmarks visuo-tactile bimanual manipulation across hand designs, and DexJoCo~\cite{wang2026dexjoco} provides task-oriented dexterous tasks covering tool use, bimanual coordination and long-horizon execution in MuJoCo.

For navigation, Room-to-Room~\cite{anderson2018r2r} established instruction following on a discrete navigation graph, VLN-CE~\cite{krantz2020vlnce} moved the task to continuous environments with low-level actions, and LHPR-VLN~\cite{song2025lhvln} extends it to long-horizon, multi-stage instructions.

Household benchmarks broaden the scope further. BEHAVIOR and BEHAVIOR-1K~\cite{srivastava2021behavior,li2024behavior1k,behavior2026challenge} derive everyday activities from human needs and evaluate full-length household tasks. Housekeep~\cite{kant2022housekeep} and TIDEE~\cite{sarch2022tidee} test commonsense tidying without explicit instructions, HAZARD~\cite{zhou2024hazard} places agents in dynamically changing environments, PARTNR~\cite{chang2025partnr} evaluates multi-agent planning, and EmbodiedBench~\cite{yang2025embodiedbench} and IS-Bench~\cite{lu2026isbench} evaluate multimodal LLM agents on vision-driven tasks and on interactive safety, respectively. Simulation platforms such as Genie Sim~\cite{yin2026geniesim} continue to raise the fidelity of these evaluations.

Benchmarks lose discriminative power as leading systems saturate them~\cite{kiela2021dynabench,ott2022mapping}. Evaluation of language agents has responded with harder tasks~\cite{phan2025hle,sun2026ale,arcprize2026arcagi3} and with outcome-based protocols in which an agent acts in an environment by any means and is scored on the result~\cite{jimenez2024swebench,xie2024osworld,merrill2026terminalbench}, a protocol that EmbodiedSWE~\cite{shen2026embodiedswe} brings to robotics. As discussed in Sec.~\ref{sec:benchmark}, most of the embodied benchmarks above were designed for policies trained by imitation, and we argue that general-purpose agents call for a different design.

\subsection{Empirical Studies of Frontier Models in Embodied Settings}
\label{sec:related_studies}

Each generation of frontier models has prompted empirical studies of its embodied abilities. Shortly after the release of ChatGPT and GPT-4V, \citet{vemprala2024chatgpt} and \citet{wake2024gpt4v} examined how these models could be prompted to plan and program robot tasks, and NavGPT~\cite{zhou2024navgpt}, MapGPT~\cite{chen2024mapgpt} and Open-Nav~\cite{qiao2025opennav} studied zero-shot vision-and-language navigation with LLMs reasoning over textual or map-based descriptions of the scene. GPT-4 also generated dense end-effector trajectories and replanned with external visual feedback~\cite{kwon2024trajectory}. CaP-X~\cite{fu2026capx} compared action abstractions, interaction modes, and perceptual grounding, exposing the dependence of model performance on supplied tools and interfaces. Our previous study~\cite{cai2025has} evaluated GPT-5 on eight spatial-intelligence benchmarks under a unified taxonomy and found unprecedented strength that nonetheless fell short of human performance across a broad range of tasks. The present report extends that effort from spatial to embodied intelligence, and from perception and reasoning to closed-loop control.

The release of GPT-6-Astra~\cite{openai2026gpt6summary} set off a wave of such studies within weeks. In navigation, two evaluations on R2R-CE~\cite{dai2026howfar,dai2026lightsup} find that GPT-6-Astra rivals or surpasses prior zero-shot and supervised methods from monocular RGB input, yet still struggles to turn correct local judgments into reliable route execution and stopping decisions. In manipulation, \citet{zhang2026unexpected} report that GPT-6-Astra ranks first among public entries on the RoboDojo leaderboard while other frontier LLMs succeed in fewer than 2\% of trials, but also that one-shot demonstrations lower rather than raise its success rate and that real-robot trials had to be halted because of unsafe actions. Astranomy~\cite{yang2026astranomy} compares direct, agentic, and policy-assisted control, while Ditto-Bench~\cite{long2026astrarobotcontrol} examines action interfaces, memory, and reasoning effort in contact-rich manipulation. Other studies examine how demonstration formats and accumulated interaction memory affect in-context adaptation~\cite{liu2026roboicl,cheng2026gptpolicy,yao2026ebench,yu2026retrace}, the reuse of body knowledge and prior experience together with sim-to-real transfer~\cite{he2026robotmanip}, direct manipulation on LIBERO~\cite{liberoagent2026}, token-efficient agents that compose classical primitives with VLA calls~\cite{si2026fewer}, and GPT-6-Astra's capabilities as an embodied policy more broadly~\cite{astraembodied2026,yang2026robotworld,humanclaw2026astra}. RoboMME studies examine full-history direct control~\cite{jedlicki2026memory} and hierarchies combining Astra planning, learned execution, and lightweight monitoring~\cite{chen2026astrarobomme}. Real-robot studies demonstrate coding-agent control and correction~\cite{jia2026agentpolicy}, while broader evaluations expose reliability gaps and substantial execution time and inference costs~\cite{artificialanalysis2026robotics,pantograph2026frontier,robocurve2026stationery,robocurve2026astra}. Vals Robo-bench~\cite{vals2026robobench} incorporates inference delays through continuously running physics and wall-clock budgets. The Embodied Agent Arena~\cite{huang2026arena} compares seven VLMs on 1{,}000 cases drawn from 32 benchmarks and finds that GPT-6-Astra excels at precise estimation and contact localization but not at coordinating goal-directed sequences, \citet{rasheed2026hardvision} observe a similar split between semantic and metric-geometric tasks across computer vision, and RoboQuest~\cite{liu2026roboquest} examines evidence acquisition and exploration stopping decisions. RoboHarm~\cite{robocurve2026roboharm} reports that GPT-6-Astra completed 60 of 100 deliberately harmful robot-arm trials and refused only two.

These studies typically examine one benchmark family, one capability or one harness at a time, and their conclusions do not always agree: demonstrations lower GPT-6-Astra's success on RoboDojo in \citet{zhang2026unexpected}, yet RoboICL~\cite{liu2026roboicl} reports gains when demonstrations are combined with interaction memory, and RoboDawn~\cite{guo2026robodawn} reports similar gains from one-shot demonstrations with a frozen VLM. Our study differs in evaluating manipulation and navigation benchmarks under a single protocol, ablating harness design, validating on real robots, and drawing on these results to propose principles for the next generation of embodied benchmarks.

\section{Full Simulation Results}
\begingroup
\renewcommand{\topfraction}{0.9}
\renewcommand{\bottomfraction}{0.85}
\renewcommand{\textfraction}{0.08}
\renewcommand{\floatpagefraction}{0.85}
\providecommand{\reviewnote}[1]{}
\begin{table}[!htbp]
\fontsize{7}{8.4}\selectfont
\setlength{\tabcolsep}{2.5pt}\renewcommand{\arraystretch}{1.15}

\centering
\caption{Benchmark-level success rates (\%) underlying Fig.~\ref{fig:teaser}. For each benchmark, we list the best published result (SOTA) with its method, the best published agentic method (Agentic SOTA), and GPT-6-Astra + ICL under the standard ($1\times$) and cumulative extended (up to $10\times$) budgets. Benchmarks are grouped by embodiment. Published baselines are reported under the standard budget. The $1\times$ execution in this table is performed separately from $10\times$, with the respective budget included in the context during execution. Dashes mean no published agentic result or no extended-budget run. $*$ denotes the result with VLA model, while ${\ddagger}$ denotes the zero-shot results without access to the training data. }
\label{tab:teaser_results}

\begin{tabular}{@{}l lr lr !{\vrule width .4pt} rrr@{}}
\toprule
\multirow{3}{*}{Embodiment / benchmark}
& \multicolumn{4}{c}{Published baselines}
& \multicolumn{3}{c}{GPT-6-Astra (ours)} \\
\cmidrule(lr){2-5}\cmidrule(lr){6-8}
& \multicolumn{2}{c}{SOTA} & \multicolumn{2}{c}{Agentic SOTA}
& \multirow{2}{*}{Up to $1\times$}
& \multicolumn{2}{c}{Up to $10\times$} \\
\cmidrule(lr){2-3}\cmidrule(lr){4-5}\cmidrule(lr){7-8}
& Method & SR & Method & SR
& & Cutoff at $10\times$ & Cutoff at $1\times$ \\
\midrule
\rowcolor{sensepurple!15}
\multicolumn{8}{@{}l}{\textbf{Tabletop robotic arms}} \\
LIBERO-Long++~\cite{yang2026lilo} & LiLo-VLA~\cite{yang2026lilo} & 82.0 & -- & -- & 97.5 & 100.0 & 97.5 \\
LIBERO-PRO~\cite{zhou2025liberopro} & $\pi_{\mathrm{RLinf}}$~\cite{zhang2026harnessvla} & 50.0 & RPent / GPT-6 Astra~\cite{zhang2026harnessvla} & 92.6 & 98.3 & 99.3 & 98.3\\
LIBERO-Ultra-Long~\cite{yang2026lilo} & LiLo-VLA~\cite{yang2026lilo} & 44.0 & -- & -- & 82.2 & 98.9 &  82.2\\
RoboDojo~\cite{chen2026robodojo} & Awomo-0.5~\cite{awomo2026awomo05} & 29.6 & PhysicalRSI~\cite{physicalrsi2026selfharness} & 31.4 & 54.2\rlap{$^{*}$} & 81.9 & 27.6\\
RoboLab$^{\ddagger}$~\cite{yang2026robolab} & FLUX 3 Action~\cite{bfl2026flux3action} & 42.9 & -- & -- & 57.1 & 69.2 & 57.1\\
\midrule
\rowcolor{sensepurple!15}
\multicolumn{8}{@{}l}{\textbf{Humanoid}} \\
SIMPLE~\cite{wei2026simple} & Psi-0~\cite{wei2026psi0} & 92.5 & -- & -- & 96.3 & 98.8 & 97.5\\
\midrule
\rowcolor{sensepurple!15}
\multicolumn{8}{@{}l}{\textbf{Mobile platforms}} \\
BEHAVIOR~\cite{behavior2026challenge} & Rondo~\cite{behavior2026challenge} & 14.8 & Agent~\cite{behavior2026challenge} & 5.3 & 37.0 & -- & --\\
\midrule
\rowcolor{sensepurple!15}
\multicolumn{8}{@{}l}{\textbf{Dexterous hands}} \\
DexJoCo~\cite{wang2026dexjoco} & $\pi_{0.5}$~\cite{pi2025pi05} & 40.3 & -- & -- & 78.2 & 87.3 & 78.2 \\
Bench2Dex~\cite{yang2026bench2dex} & $\pi_{0.5}$~\cite{pi2025pi05} & 25.5 & -- & -- & 25.6 & 61.5 & 25.6 \\
\midrule
\rowcolor{sensepurple!15}
\multicolumn{8}{@{}l}{\textbf{Navigation}} \\
VLN-CE RxR~\cite{krantz2020vlnce} & Qwen-RobotNav~\cite{zhang2026qwenrobotnav} & 76.5 & -- & -- & 82.0 & 90.0 & 84.0 \\
VLN-CE R2R~\cite{krantz2020vlnce} & Robostral Navigate~\cite{bounhar2026robostral} & 77.4 & GPT-6-Astra (ultra)~\cite{dai2026lightsup}
& 81.3 & 80.0 & 88.0 & 85.0 \\
LHPR-VLN~\cite{song2025lhvln} & FantasyVLN~\cite{zuo2026fantasyvln} & 2.4 & Claude Code Opus-5 & 5.3 & 21.1 & 28.9 & 25.4\\
\bottomrule
\end{tabular}

\vspace{3pt}
\begin{minipage}{\linewidth}
\end{minipage}
\end{table}

\subsection{LIBERO-PRO}
\noindent\textbf{Tasks and demonstrations.}
LIBERO-PRO~\cite{zhou2025liberopro} applies task-preserving perturbations to the four LIBERO suites and separates each suite into Task and Swap variants. We evaluate the complete official benchmark: eight suites, ten tasks per suite, and ten fixed initial states per task, for 800 episodes in total. Before evaluation, we freeze a package selected by the exact base task, combining matching Task/Swap task-family notes with global strategy, perception, verification, and recovery notes distilled from a separate public exploration phase. The package contains no reported evaluation trajectory or outcome, current initial state, or reusable pixels, coordinates, object suffixes, or action sequences. We refer to our configuration as \textbf{GPT-6-Astra + ICL}. It uses maximum reasoning effort and direct robot control, with both the VLA and SAM3 disabled. All 800 episodes are valid and uncensored.

\begin{table}[!htbp]
\fontsize{7}{8.4}\selectfont
\setlength{\tabcolsep}{2pt}\renewcommand{\arraystretch}{1.15}

\centering
\caption{Task-level performance and computational cost on LIBERO-PRO~\cite{zhou2025liberopro}. Each task row pairs the official Task and Swap perturbations of one base task (ten fixed initial states per perturbation). Four representative published baselines are integrated into the family aggregate rows because their task-level results are not reported. Our GPT-6-Astra + ICL results use the standard ($1\times$) and cumulative extended (up to $10\times$) budgets.}
\label{tab:libero_pro_results}

\maxsizebox{\textwidth}{!}{%
\begin{tabular}{@{}l rrrr !{\vrule width .4pt} r rr rrr rrr@{}}
\toprule
\multirow{4}{*}{Task family / base task}
& \multicolumn{4}{c}{Published baselines}
& \multicolumn{9}{c}{GPT-6-Astra + ICL (ours)} \\
\cmidrule(lr){2-5}\cmidrule(lr){6-14}
& \multicolumn{4}{c}{Task / Swap SR (\%)}
& \multirow{3}{*}{$N$}
& \multicolumn{2}{c}{Task / Swap SR (\%)}
& \multicolumn{3}{c}{Tokens (millions)}
& \multicolumn{3}{c}{Time (minutes)} \\
\cmidrule(lr){2-5}\cmidrule(lr){7-8}\cmidrule(lr){9-11}\cmidrule(lr){12-14}
& \multirow{2}{*}{GPT-6-Astra~\cite{zhang2026harnessvla}} & \multirow{2}{*}{ASPIRE~\cite{lu2026aspireagenticskillsdiscovery}} & \multirow{2}{*}{$\pi_{\mathrm{RLinf}}$~\cite{zhang2026harnessvla}} & \multirow{2}{*}{$\pi_{0.5}$~\cite{pi2025pi05}}
& & $1\times$ & up to $10\times$ & \multirow{2}{*}{ICL} & $1\times$ & up to $10\times$
& \multirow{2}{*}{ICL} & $1\times$ & up to $10\times$ \\
& & & & & & step limit & step limit & & step limit & step limit & & step limit & step limit \\
\midrule
\rowcolor{sensepurple!15}
\textbf{Spatial} & 100 / 98 & 60 / 51 & 42 / 59 & 1 / 20 & 200 & 100 / 100 & 100 / 100 & 1.61 & 0.36 & 0.36 & 18.8 & 13.4 & 13.4 \\
Black bowl: between plate/ramekin $\rightarrow$ plate & -- & -- & -- & -- & 20 & 100 / 100 & 100 / 100 & 1.67 & 0.32 & 0.32 & 19.0 & 10.1 & 10.1 \\
Black bowl: next to ramekin $\rightarrow$ plate & -- & -- & -- & -- & 20 & 100 / 100 & 100 / 100 & 1.76 & 0.34 & 0.34 & 23.0 & 11.9 & 11.9 \\
Black bowl: table center $\rightarrow$ plate & -- & -- & -- & -- & 20 & 100 / 100 & 100 / 100 & 1.71 & 0.30 & 0.30 & 18.7 & 10.3 & 10.3 \\
Black bowl: on cookie box $\rightarrow$ plate & -- & -- & -- & -- & 20 & 100 / 100 & 100 / 100 & 1.42 & 0.38 & 0.38 & 17.3 & 12.5 & 12.5 \\
Black bowl: top cabinet drawer $\rightarrow$ plate & -- & -- & -- & -- & 20 & 100 / 100 & 100 / 100 & 1.72 & 0.45 & 0.45 & 18.3 & 15.1 & 15.1 \\
Black bowl: on ramekin $\rightarrow$ plate & -- & -- & -- & -- & 20 & 100 / 100 & 100 / 100 & 1.40 & 0.27 & 0.27 & 18.1 & 10.5 & 10.5 \\
Black bowl: next to cookie box $\rightarrow$ plate & -- & -- & -- & -- & 20 & 100 / 100 & 100 / 100 & 1.44 & 0.47 & 0.47 & 18.7 & 20.8 & 20.8 \\
Black bowl: on stove $\rightarrow$ plate & -- & -- & -- & -- & 20 & 100 / 100 & 100 / 100 & 1.88 & 0.31 & 0.31 & 18.9 & 14.1 & 14.1 \\
Black bowl: next to plate $\rightarrow$ plate & -- & -- & -- & -- & 20 & 100 / 100 & 100 / 100 & 1.83 & 0.35 & 0.35 & 18.7 & 13.9 & 13.9 \\
Black bowl: on cabinet $\rightarrow$ plate & -- & -- & -- & -- & 20 & 100 / 100 & 100 / 100 & 1.30 & 0.37 & 0.37 & 17.0 & 14.6 & 14.6 \\
\midrule
\rowcolor{sensepurple!15}
\textbf{Object} & 100 / 99 & 95 / 98 & 71 / 78 & 1 / 17 & 200 & 100 / 100 & 100 / 100 & 1.76 & 0.28 & 0.28 & 19.6 & 18.1 & 18.1 \\
Alphabet soup $\rightarrow$ basket & -- & -- & -- & -- & 20 & 100 / 100 & 100 / 100 & 1.56 & 0.26 & 0.26 & 21.6 & 12.6 & 12.6 \\
Cream cheese $\rightarrow$ basket & -- & -- & -- & -- & 20 & 100 / 100 & 100 / 100 & 1.60 & 0.31 & 0.31 & 19.1 & 15.7 & 15.7 \\
Salad dressing $\rightarrow$ basket & -- & -- & -- & -- & 20 & 100 / 100 & 100 / 100 & 2.42 & 0.26 & 0.26 & 21.4 & 17.8 & 17.8 \\
BBQ sauce $\rightarrow$ basket & -- & -- & -- & -- & 20 & 100 / 100 & 100 / 100 & 1.52 & 0.30 & 0.30 & 20.8 & 21.1 & 21.1 \\
Ketchup $\rightarrow$ basket & -- & -- & -- & -- & 20 & 100 / 100 & 100 / 100 & 1.83 & 0.31 & 0.31 & 18.9 & 25.6 & 25.6 \\
Tomato sauce $\rightarrow$ basket & -- & -- & -- & -- & 20 & 100 / 100 & 100 / 100 & 2.11 & 0.28 & 0.28 & 21.0 & 24.7 & 24.7 \\
Butter $\rightarrow$ basket & -- & -- & -- & -- & 20 & 100 / 100 & 100 / 100 & 1.45 & 0.29 & 0.29 & 16.6 & 22.1 & 22.1 \\
Milk $\rightarrow$ basket & -- & -- & -- & -- & 20 & 100 / 100 & 100 / 100 & 1.45 & 0.29 & 0.29 & 19.3 & 16.7 & 16.7 \\
Chocolate pudding $\rightarrow$ basket & -- & -- & -- & -- & 20 & 100 / 100 & 100 / 100 & 2.15 & 0.26 & 0.26 & 19.0 & 13.3 & 13.3 \\
Orange juice $\rightarrow$ basket & -- & -- & -- & -- & 20 & 100 / 100 & 100 / 100 & 1.55 & 0.23 & 0.23 & 18.7 & 11.7 & 11.7 \\
\midrule
\rowcolor{sensepurple!15}
\textbf{Goal} & 88 / 99 & 45 / 81 & 45 / 42 & 2 / 38 & 200 & 99 / 100 & 100 / 100 & 2.04 & 0.43 & 0.43 & 20.1 & 17.1 & 17.2 \\
Open middle cabinet drawer & -- & -- & -- & -- & 20 & 100 / 100 & 100 / 100 & 1.54 & 0.94 & 0.94 & 16.8 & 29.7 & 29.7 \\
Bowl $\rightarrow$ stove & -- & -- & -- & -- & 20 & 90 / 100 & 100 / 100 & 1.46 & 0.48 & 0.49 & 18.0 & 16.3 & 16.8 \\
Wine bottle $\rightarrow$ cabinet top & -- & -- & -- & -- & 20 & 100 / 100 & 100 / 100 & 1.49 & 0.27 & 0.27 & 20.3 & 11.2 & 11.2 \\
Open top drawer; bowl inside & -- & -- & -- & -- & 20 & 100 / 100 & 100 / 100 & 2.78 & 0.66 & 0.66 & 24.1 & 21.7 & 21.7 \\
Bowl $\rightarrow$ cabinet top & -- & -- & -- & -- & 20 & 100 / 100 & 100 / 100 & 2.32 & 0.35 & 0.35 & 20.8 & 13.3 & 13.3 \\
Push plate to stove front & -- & -- & -- & -- & 20 & 100 / 100 & 100 / 100 & 2.55 & 0.54 & 0.54 & 20.3 & 24.9 & 24.9 \\
Cream cheese $\rightarrow$ bowl & -- & -- & -- & -- & 20 & 100 / 100 & 100 / 100 & 2.97 & 0.28 & 0.28 & 20.7 & 14.4 & 14.4 \\
Turn on stove & -- & -- & -- & -- & 20 & 100 / 100 & 100 / 100 & 1.66 & 0.15 & 0.15 & 23.8 & 9.9 & 9.9 \\
Bowl $\rightarrow$ plate & -- & -- & -- & -- & 20 & 100 / 100 & 100 / 100 & 2.11 & 0.26 & 0.26 & 18.3 & 13.5 & 13.5 \\
Wine bottle $\rightarrow$ rack & -- & -- & -- & -- & 20 & 100 / 100 & 100 / 100 & 1.50 & 0.40 & 0.40 & 17.8 & 16.1 & 16.1 \\
\midrule
\rowcolor{sensepurple!15}
\textbf{Long-horizon} & 85 / 72 & 38.3 / 22.6 & 49 / 14 & 1 / 8 & 200 & 97 / 90 & 99 / 95 & 1.83 & 1.02 & 2.92 & 19.8 & 27.5 & 46.6 \\
Soup and tomato sauce $\rightarrow$ basket & -- & -- & -- & -- & 20 & 100 / 100 & 100 / 100 & 2.16 & 0.61 & 0.61 & 22.1 & 20.0 & 20.0 \\
Cream cheese and butter $\rightarrow$ basket & -- & -- & -- & -- & 20 & 100 / 90 & 100 / 90 & 1.74 & 0.57 & 4.41 & 17.8 & 16.7 & 52.8 \\
Turn on stove; moka pot on stove & -- & -- & -- & -- & 20 & 100 / 100 & 100 / 100 & 2.47 & 0.71 & 0.71 & 24.2 & 22.8 & 22.8 \\
Black bowl in bottom drawer; close drawer & -- & -- & -- & -- & 20 & 80 / 100 & 100 / 100 & 1.60 & 1.65 & 1.91 & 19.9 & 35.0 & 39.7 \\
White mug left; yellow-white mug right & -- & -- & -- & -- & 20 & 100 / 100 & 100 / 100 & 1.70 & 0.87 & 0.87 & 19.9 & 32.3 & 32.3 \\
Book $\rightarrow$ back of caddy & -- & -- & -- & -- & 20 & 90 / 90 & 90 / 90 & 1.55 & 0.50 & 3.78 & 19.1 & 21.7 & 56.1 \\
White mug on plate; pudding to right & -- & -- & -- & -- & 20 & 100 / 100 & 100 / 100 & 1.97 & 0.64 & 0.64 & 21.6 & 20.9 & 20.9 \\
Soup and cream cheese $\rightarrow$ basket & -- & -- & -- & -- & 20 & 100 / 100 & 100 / 100 & 1.46 & 0.64 & 0.64 & 17.4 & 23.4 & 23.4 \\
Both moka pots $\rightarrow$ stove & -- & -- & -- & -- & 20 & 100 / 90 & 100 / 100 & 1.57 & 1.04 & 1.09 & 17.3 & 32.7 & 33.9 \\
Yellow-white mug in microwave; close & -- & -- & -- & -- & 20 & 100 / 30 & 100 / 70 & 2.11 & 3.00 & 14.51 & 19.2 & 49.1 & 164.0 \\
\midrule
\rowcolor{sensepurple!15}
\textbf{Overall / mean} & \textbf{92.63} & \textbf{61.36} & \textbf{50.00} & \textbf{11.00} & \textbf{800} & \textbf{98.25} & \textbf{99.25} & 1.81 & 0.52 & 1.00 & 19.6 & 19.0 & 23.8 \\
\bottomrule
\end{tabular}%
}

\vspace{3pt}
\begin{minipage}{\textwidth}
\fontsize{8}{9.6}\selectfont
\emph{Note.} Group aggregate rows precede their ten base-task rows. Within a family, each performance entry is Task / Swap SR; the overall row is a single SR over all 800 episodes. The four baseline columns retain GPT-6-Astra and the three strongest representative policy baselines from the suite-level comparison. The GPT-6-Astra column denotes the published RPent configuration with low reasoning, which also uses the official RPent memory; our method uses maximum reasoning and no VLA or SAM3 calls. Dashes mean unavailable task-level baseline results, not zero. $N$ is the number of our evaluation episodes. Official RPent memory was already publicly available; for the final evaluation, we reorganized this memory by task into ICL packages. Because the official exploration cost was not reported, the ICL column provides, for reference, the actual mean cost of our reproduction of 40 independent task-specific explorations. The packages produced by these reproduced explorations were not used in the final evaluation. Test costs exclude ICL; up-to-$10\times$ includes the original $1\times$ execution and continuation in the same simulator state and conversation, and is not directly comparable with standard-budget baselines. Tokens are observed provider input-plus-output usage and are lower bounds because 117 episodes have partial usage accounting; wall time is fully observed.
\end{minipage}
\end{table}

\noindent\textbf{Observation and control settings.}
GPT-6-Astra observes calibrated RGB images from a fixed third-person \texttt{agentview} camera and the Panda wrist camera (native $256\!\times\!256$ observations, with $1024\!\times\!1024$ inspection renders), together with the measured end-effector pose and gripper state. The agent interface also provides calibration-based pixel back-projection into the current scene; LIBERO-PRO does not expose object poses or evaluator diagnostics to the planner. The underlying action is the benchmark's seven-dimensional Panda OSC-pose command at 20\,Hz: three translational components, three axis--angle rotational components, and one gripper command. The interface compiles model-selected Cartesian motion, wrist-rotation, and gripper primitives into these actions. A unit translational or rotational component corresponds to 5\,cm or 0.5\,rad, respectively.

\noindent\textbf{Budgets and continuation.}
The standard $1\times$ boundary is the first of 10{,}000 environment steps or 100 planner turns. An unsuccessful episode then continues from the same simulator state and planner conversation to a cumulative $10\times$ environment-step budget; the continuation is not terminated by the 100-turn $1\times$ cap. The frozen $1\times$ outcome is never replaced by the continuation. Table~\ref{tab:libero_pro_results} reports success rate (SR) for every paired Task/Swap base task, the four family aggregates, and all 800 episodes, while integrating prior agentic and policy-based methods at their reported suite-level granularity.

\noindent\textbf{Results and comparison.}
At $1\times$, \textbf{GPT-6-Astra + ICL} succeeds in 786 of 800 episodes (98.25\%), exceeding the published RPent/GPT-6-Astra baseline with low reasoning~\cite{zhang2026harnessvla}, the strongest prior result, by 5.62 percentage points. It obtains 99--100\% SR on six suites; the remaining errors concentrate in Long Task (97\%) and Long Swap (90\%). Continuing to the cumulative $10\times$ boundary raises the result to 794/800 (99.25\%), including 99\% on Long Task and 95\% on Long Swap. Table~\ref{tab:libero_pro_results} also reports the mean test-time token and wall-clock cost at both boundaries.

\FloatBarrier
\subsection{LIBERO-Long++ and LIBERO-Ultra-Long}
\noindent\textbf{Tasks and demonstrations.}
LIBERO-Long++ and LIBERO-Ultra-Long~\cite{yang2026lilo} evaluate compositional manipulation under visual clutter and reordered skill chains. Long++ contains six scenarios in their original and reordered forms, yielding 12 configurations; Ultra-Long contains three scenarios under the original ordering and two reorderings, yielding nine configurations. We follow the released protocol and run ten independent bare-random resets per configuration, for 120 Long++ and 90 Ultra-Long episodes. Before testing, we construct one frozen package per composite task solely from the released atomic-skill training split, which contains 4{,}050 trajectories (150 per atomic skill; 351{,}999 transitions). The package summarizes canonical, geometric-variation, and recovery examples as phase, contact, verification, and failure cues with labeled agent-view and wrist-view contact sheets. It contains no evaluation rollout, initial state, reward, success label, or test-specific coordinates. We refer to our configuration as \textbf{GPT-6-Astra + ICL}. It uses maximum reasoning effort and direct robot control, with both the VLA and SAM3 disabled.

\begin{table}[!htbp]
\fontsize{7}{8.4}\selectfont
\setlength{\tabcolsep}{2.5pt}\renewcommand{\arraystretch}{1.15}

\centering
\caption{Task-level performance and computational cost on LIBERO-Long++ and LIBERO-Ultra-Long~\cite{yang2026lilo}. Published baselines are integrated into the aggregate rows because their task-level results are not reported. Our GPT-6-Astra + ICL experiment evaluates ten episodes per task at the standard ($1\times$) and cumulative extended (up to $10\times$) budgets. Performance entries are success rate (SR) / average progress (AP), both in percent.}
\label{tab:libero_long_results}

\maxsizebox{\textwidth}{!}{%
\begin{tabular}{@{}l rrr !{\vrule width .4pt} r rr rrr rrr@{}}
\toprule
\multirow{4}{*}{Benchmark / task}
& \multicolumn{3}{c}{Published baselines}
& \multicolumn{9}{c}{GPT-6-Astra + ICL (ours)} \\
\cmidrule(lr){2-4}\cmidrule(lr){5-13}
& \multicolumn{3}{c}{SR / AP (\%)}
& \multirow{3}{*}{$N$}
& \multicolumn{2}{c}{SR / AP (\%)}
& \multicolumn{3}{c}{Tokens (millions)}
& \multicolumn{3}{c}{Time (minutes)} \\
\cmidrule(lr){2-4}\cmidrule(lr){6-7}\cmidrule(lr){8-10}\cmidrule(lr){11-13}
& \multirow{2}{*}{LiLo-VLA~\cite{yang2026lilo}} & \multirow{2}{*}{$\pi_{0.5}$~\cite{pi2025pi05}} & \multirow{2}{*}{OpenVLA-OFT~\cite{kim2025oft}}
& & $1\times$ & up to $10\times$ & \multirow{2}{*}{ICL} & $1\times$ & up to $10\times$
& \multirow{2}{*}{ICL} & $1\times$ & up to $10\times$ \\
& & & & & step limit & step limit & & step limit & step limit & & step limit & step limit \\
\midrule
\textbf{Long++: Original} & 78.0 / 88.0 & 83.0 / 93.0 & 7.0 / 11.0 & 60 & 98.3 / 99.6 & 100.0 / 100.0 & 1.68 & 1.04 & 1.08 & 20.3 & 56.8 & 57.9 \\
Stove, then moka pot & -- & -- & -- & 10 & 100.0 / 100.0 & 100.0 / 100.0 & 1.95 & 0.80 & 0.80 & 21.1 & 53.0 & 53.0 \\
Soup, then cream cheese & -- & -- & -- & 10 & 100.0 / 100.0 & 100.0 / 100.0 & 1.58 & 1.26 & 1.26 & 19.9 & 58.5 & 58.5 \\
Soup, then tomato sauce & -- & -- & -- & 10 & 100.0 / 100.0 & 100.0 / 100.0 & 1.56 & 1.16 & 1.16 & 17.7 & 44.5 & 44.5 \\
Cream cheese, then butter & -- & -- & -- & 10 & 90.0 / 97.5 & 100.0 / 100.0 & 1.62 & 1.22 & 1.49 & 20.2 & 66.4 & 72.8 \\
White mug, then yellow-white mug & -- & -- & -- & 10 & 100.0 / 100.0 & 100.0 / 100.0 & 1.97 & 1.00 & 1.00 & 22.4 & 64.0 & 64.0 \\
White mug, then pudding & -- & -- & -- & 10 & 100.0 / 100.0 & 100.0 / 100.0 & 1.43 & 0.80 & 0.80 & 20.4 & 54.3 & 54.3 \\
\midrule
\textbf{Long++: Reordered} & 85.0 / 89.0 & 0.0 / 0.0 & 0.0 / 0.0 & 60 & 96.7 / 98.9 & 100.0 / 100.0 & 1.76 & 1.04 & 1.06 & 20.3 & 49.4 & 50.0 \\
Moka pot, then stove & -- & -- & -- & 10 & 80.0 / 93.3 & 100.0 / 100.0 & 3.28 & 1.12 & 1.22 & 24.4 & 57.9 & 61.5 \\
Cream cheese, then soup & -- & -- & -- & 10 & 100.0 / 100.0 & 100.0 / 100.0 & 1.21 & 1.03 & 1.03 & 22.1 & 62.2 & 62.2 \\
Tomato sauce, then soup & -- & -- & -- & 10 & 100.0 / 100.0 & 100.0 / 100.0 & 1.48 & 1.23 & 1.23 & 17.6 & 75.7 & 75.7 \\
Butter, then cream cheese & -- & -- & -- & 10 & 100.0 / 100.0 & 100.0 / 100.0 & 1.81 & 1.06 & 1.06 & 21.2 & 48.1 & 48.1 \\
Yellow-white mug, then white mug & -- & -- & -- & 10 & 100.0 / 100.0 & 100.0 / 100.0 & 1.21 & 0.94 & 0.94 & 17.1 & 31.3 & 31.3 \\
Pudding, then white mug & -- & -- & -- & 10 & 100.0 / 100.0 & 100.0 / 100.0 & 1.59 & 0.86 & 0.86 & 19.2 & 21.3 & 21.3 \\
\midrule
\rowcolor{sensepurple!15}
\textbf{LIBERO-Long++ average} & \textbf{82.0 / 89.0} & \textbf{42.0 / 46.0} & \textbf{3.0 / 5.0} & \textbf{120} & \textbf{97.5 / 99.2} & \textbf{100.0 / 100.0} & 1.72 & 1.04 & 1.07 & 20.3 & 53.1 & 53.9 \\
\midrule
\textbf{Ultra-Long: Original} & 53.0 / 79.0 & 0.0 / 1.0 & 0.0 / 0.0 & 30 & 63.3 / 90.1 & 100.0 / 100.0 & 2.08 & 9.19 & 10.18 & 21.2 & 185.9 & 199.4 \\
Complete Kitchen Organization & -- & -- & -- & 10 & 50.0 / 83.3 & 100.0 / 100.0 & 2.09 & 4.35 & 5.80 & 21.1 & 118.7 & 142.5 \\
Organize Table & -- & -- & -- & 10 & 70.0 / 95.0 & 100.0 / 100.0 & 2.14 & 16.86 & 17.75 & 20.3 & 268.0 & 280.3 \\
Cooking Preparation Setup & -- & -- & -- & 10 & 70.0 / 92.0 & 100.0 / 100.0 & 2.00 & 6.35 & 7.00 & 22.2 & 171.1 & 175.3 \\
\midrule
\textbf{Ultra-Long: Variant 1} & 37.0 / 84.0 & 0.0 / 0.0 & 0.0 / 0.0 & 30 & 93.3 / 98.1 & 96.7 / 98.3 & 2.03 & 8.37 & 8.38 & 21.3 & 84.0 & 84.1 \\
Complete Kitchen Organization & -- & -- & -- & 10 & 100.0 / 100.0 & 100.0 / 100.0 & 2.31 & 4.14 & 4.14 & 23.7 & 73.6 & 73.6 \\
Organize Table & -- & -- & -- & 10 & 80.0 / 94.4 & 90.0 / 95.0 & 2.32 & 15.27 & 15.30 & 20.7 & 109.0 & 109.3 \\
Cooking Preparation Setup & -- & -- & -- & 10 & 100.0 / 100.0 & 100.0 / 100.0 & 1.48 & 5.69 & 5.69 & 19.4 & 69.3 & 69.3 \\
\midrule
\textbf{Ultra-Long: Variant 2} & 43.0 / 74.0 & 0.0 / 0.0 & 0.0 / 0.0 & 30 & 90.0 / 98.8 & 100.0 / 100.0 & 1.94 & 9.11 & 9.59 & 21.8 & 62.8 & 67.0 \\
Complete Kitchen Organization & -- & -- & -- & 10 & 100.0 / 100.0 & 100.0 / 100.0 & 2.12 & 4.64 & 4.64 & 23.6 & 50.3 & 50.3 \\
Organize Table & -- & -- & -- & 10 & 80.0 / 97.5 & 100.0 / 100.0 & 2.14 & 17.77 & 19.20 & 19.7 & 91.7 & 104.2 \\
Cooking Preparation Setup & -- & -- & -- & 10 & 90.0 / 99.0 & 100.0 / 100.0 & 1.56 & 4.92 & 4.94 & 22.2 & 46.4 & 46.6 \\
\midrule
\rowcolor{sensepurple!15}
\textbf{LIBERO-Ultra-Long average} & \textbf{44.0 / 79.0} & \textbf{0.0 / 0.3} & \textbf{0.0 / 0.0} & \textbf{90} & \textbf{82.2 / 95.7} & \textbf{98.9 / 99.4} & 2.02 & 8.89 & 9.38 & 21.4 & 110.9 & 116.8 \\
\midrule
\rowcolor{sensepurple!15}
\textbf{Overall / mean} & \textbf{69.0 / 86.0} & \textbf{28.0 / 31.0} & \textbf{2.0 / 4.0} & \textbf{210} & \textbf{92.4 / 98.1} & \textbf{99.6 / 99.8} & 1.85 & 4.40 & 4.63 & 20.8 & 77.9 & 80.9 \\
\bottomrule
\end{tabular}%
}

\vspace{3pt}
\begin{minipage}{\textwidth}
\fontsize{8}{9.6}\selectfont
\emph{Note.} Ordering-level aggregate rows precede their task rows; highlighted rows report the two benchmark averages and the cross-benchmark overall mean. Dashes indicate that the published baselines report only ordering-level aggregates, not task-level results. $N$ is the number of our evaluation episodes; published baselines use their released evaluation protocols. The overall performance row macro-averages the nine base scenarios (six Long++ and three Ultra-Long), whereas its resource columns average the 21 task preparations and 210 scored episodes. ICL is a one-time task-specific preparation cost; evaluation costs exclude ICL. The up-to-$10\times$ columns include the original $1\times$ execution and continuation in the same simulator state and conversation, and are not directly comparable with standard-budget baselines. Tokens are observed provider input-plus-output usage and are lower bounds because usage accounting is partial for 8/120 Long++ and 27/90 Ultra-Long episodes; wall time is fully observed.
\end{minipage}
\end{table}

\noindent\textbf{Observation and control settings.}
GPT-6-Astra observes calibrated RGB images from a fixed third-person \texttt{agentview} camera and the Panda wrist camera (native $256\!\times\!256$ observations, with $1024\!\times\!1024$ inspection renders), the measured end-effector pose and gripper state, and the released ordered-skill and instance-pose feedback used by our LiLo adapter. Calibration-based pixel back-projection supplies current-scene 3D geometry. The underlying action is the benchmark's seven-dimensional Panda OSC-pose command at 20\,Hz: three translational components, three axis--angle rotational components, and one gripper command. The agent interface compiles model-selected Cartesian motion, wrist-rotation, and gripper primitives into these actions. A unit translational or rotational component corresponds to 5\,cm or 0.5\,rad, respectively.

\noindent\textbf{Budgets and continuation.}
The standard $1\times$ step budgets are 600 or 800 steps on Long++ and 1{,}800, 2{,}000, or 3{,}200 steps on Ultra-Long. Success rate (SR) requires completion of the entire ordered chain within this boundary. Average progress (AP) is the fraction of the uninterrupted correct skill prefix completed at the same boundary. An unsuccessful episode continues from the same simulator state and planner conversation to a cumulative $10\times$ step budget; the $1\times$ outcome and prefix snapshot remain frozen. There is no separate planner-turn cap; a 12-hour planner wall-clock guard handles infrastructure censoring but does not change the environment-step budget. Prior work does not report an agentic method under this protocol. Table~\ref{tab:libero_long_results} reports every task and compares \textbf{GPT-6-Astra + ICL} with LiLo-VLA and two learned-policy baselines in the corresponding ordering-level aggregate rows.

\noindent\textbf{Results and comparison.}
At $1\times$, \textbf{GPT-6-Astra + ICL} achieves 97.5\% SR and 99.2\% AP on LIBERO-Long++, and 82.2\% SR and 95.7\% AP on LIBERO-Ultra-Long, exceeding LiLo-VLA by 15.5 and 38.2 percentage points in SR, respectively. The cumulative $10\times$ results reach 100.0\% SR/AP on Long++ and 98.9\% SR with 99.4\% AP on Ultra-Long. Macro-averaged over the nine scenarios, the corresponding results are 92.4\%/98.1\% at $1\times$ and 99.6\%/99.8\% at $10\times$. Table~\ref{tab:libero_long_results} also reports one-time preparation cost per composite task and mean test-time cost per episode.

\FloatBarrier
\subsection{RoboDojo}
\label{app:robodojo}
\noindent\textbf{Tasks and demonstrations.}
We refer to our experiment as \textbf{GPT-6-Astra + ICL}.
We evaluate all 42 RoboDojo tasks across 5 groups~\cite{chen2026robodojo}. Generalization, Precision, Long-Horizon and Memory contain 34 tasks in total; Open contains 8 tasks. Each task uses \texttt{layout\_id=0--4}, giving 210 evaluation episodes. We fix \texttt{seed=0}; each layout is initialized with \texttt{env.reset(seed=[layout\_id])}.
For each task in Generalization, Precision, Long-Horizon and Memory, GPT-6-Astra studies 10 demonstrations of that task. The 8 Open tasks have no task-specific demonstrations and share one study of 340 demonstrations from the other 4 groups, 10 per task.

\begin{table}[!htbp]
\fontsize{7}{8.4}\selectfont
\setlength{\tabcolsep}{2pt}\renewcommand{\arraystretch}{1.15}

\centering
\caption{Performance and computational cost by task group and task on RoboDojo~\cite{chen2026robodojo}. The published GPT-6-Astra baseline is reported by the RoboDojo team~\cite{zhang2026unexpected}. Our ICL experiment using GPT-6-Astra evaluates 5 episodes per task (210 in total). Group aggregate rows precede their task results. Results are reported at the standard control-step limit ($1\times$) and with an extended budget of up to $10\times$ that limit. Token usage and time refer to our ICL experiment using GPT-6-Astra.}
\label{tab:robodojo_results}

\maxsizebox{\textwidth}{!}{%
\begin{tabular}{@{}l r rr !{\vrule width .4pt} r rr rrr rrr@{}}
\toprule
\multirow{4}{*}{Task group / task} & \multicolumn{3}{c}{Published baselines} & \multicolumn{9}{c}{GPT-6-Astra + ICL (ours)}\\
\cmidrule(lr){2-4}\cmidrule(lr){5-13}
& \multirow{3}{*}{$N$} & \multicolumn{2}{c}{Mean score / success rate} & \multirow{3}{*}{$N$} & \multicolumn{2}{c}{Mean score / success rate} & \multicolumn{3}{c}{Token usage (millions)} & \multicolumn{3}{c}{Time (minutes)}\\
\cmidrule(lr){3-4}\cmidrule(lr){6-7}\cmidrule(lr){8-10}\cmidrule(lr){11-13}
& & \multirow{2}{*}{GPT-6-Astra~\cite{zhang2026unexpected}} & \multirow{2}{*}{Awomo-0.5~\cite{awomo2026awomo05}} & & $1\times$ & up to $10\times$ & \multirow{2}{*}{ICL} & $1\times$ & up to $10\times$ & \multirow{2}{*}{ICL} & $1\times$ & up to $10\times$\\
& & & & & step limit & step limit & & step limit & step limit & & step limit & step limit\\
\midrule
\rowcolor{sensepurple!15}
\textbf{Generalization} & 600 & 35.32 / 32.67 & 37.70 / 30.89 & 60 & 34.92 / 28.33 & 81.42 / 75.00 & 3.03 & 15.58 & 52.47 & 28.9 & 82.2 & 329.2\\
Stack bowls & 50 & 68.80 / 64.00 & 74.07 / 70.67 & 5 & 66.00 / 60.00 & 100.00 / 100.00 & 2.55 & 14.99 & 18.98 & 28.2 & 99.2 & 131.0\\
Push T & 50 & 68.00 / 68.00 & 5.33 / 5.33 & 5 & 20.00 / 20.00 & 100.00 / 100.00 & 3.31 & 12.62 & 21.54 & 30.8 & 79.0 & 144.4\\
Pack objects into box & 50 & 11.00 / 0.00 & 31.33 / 10.67 & 5 & 24.00 / 0.00 & 80.00 / 60.00 & 3.42 & 21.32 & 98.96 & 26.9 & 100.7 & 569.5\\
Fold clothes & 50 & 76.80 / 72.00 & 74.93 / 70.67 & 5 & 60.00 / 60.00 & 84.00 / 80.00 & 3.37 & 9.49 & 26.50 & 30.1 & 40.6 & 146.0\\
Hang mugs & 50 & 0.60 / 0.00 & 36.33 / 18.67 & 5 & 19.00 / 0.00 & 88.00 / 80.00 & 3.25 & 17.91 & 59.06 & 31.3 & 91.9 & 338.8\\
Sweep blocks & 50 & 8.00 / 8.00 & 1.33 / 1.33 & 5 & 0.00 / 0.00 & 80.00 / 80.00 & 4.07 & 21.08 & 69.73 & 31.4 & 123.0 & 353.8\\
Pour liquid into cup & 50 & 8.00 / 8.00 & 64.00 / 64.00 & 5 & 60.00 / 60.00 & 60.00 / 60.00 & 2.40 & 8.26 & 18.24 & 25.7 & 50.8 & 117.3\\
Make toast & 50 & 0.00 / 0.00 & 19.33 / 12.00 & 5 & 5.00 / 0.00 & 45.00 / 40.00 & 2.46 & 28.44 & 128.47 & 27.4 & 167.0 & 786.0\\
Arrange largest number & 50 & 76.00 / 68.00 & 26.07 / 14.67 & 5 & 54.00 / 40.00 & 100.00 / 100.00 & 2.66 & 11.65 & 18.96 & 28.7 & 53.1 & 103.0\\
Sort nesting dolls by size & 50 & 16.00 / 16.00 & 22.67 / 22.67 & 5 & 40.00 / 40.00 & 100.00 / 100.00 & 2.69 & 17.64 & 21.24 & 29.5 & 65.0 & 91.6\\
Store laptop and headphones & 50 & 0.80 / 0.00 & 18.93 / 5.33 & 5 & 8.00 / 0.00 & 40.00 / 0.00 & 3.27 & 15.48 & 137.86 & 27.6 & 82.1 & 1125.3\\
Stack blocks & 50 & 89.80 / 88.00 & 78.07 / 74.67 & 5 & 63.00 / 60.00 & 100.00 / 100.00 & 2.91 & 8.11 & 10.08 & 28.6 & 34.0 & 43.1\\
\midrule
\rowcolor{sensepurple!15}
\textbf{Precision} & 400 & 12.65 / 4.00 & 39.66 / 33.50 & 40 & 28.62 / 15.00 & 91.00 / 87.50 & 3.40 & 9.17 & 37.86 & 32.5 & 35.4 & 198.0\\
Fasten screws & 50 & 36.40 / 16.00 & 69.00 / 48.00 & 5 & 12.00 / 0.00 & 80.00 / 60.00 & 3.14 & 22.38 & 178.63 & 32.7 & 65.9 & 922.5\\
Insert tubes & 50 & 6.00 / 0.00 & 79.47 / 71.33 & 5 & 40.00 / 0.00 & 100.00 / 100.00 & 2.79 & 8.15 & 11.55 & 33.8 & 25.3 & 43.4\\
Plug in charger & 50 & 0.00 / 0.00 & 7.33 / 7.33 & 5 & 40.00 / 40.00 & 100.00 / 100.00 & 3.18 & 5.54 & 7.22 & 33.0 & 27.1 & 39.9\\
Pour balls into vase & 50 & 4.00 / 4.00 & 39.33 / 39.33 & 5 & 80.00 / 80.00 & 100.00 / 100.00 & 3.65 & 8.00 & 18.09 & 36.8 & 45.8 & 93.8\\
Play xylophone & 50 & 8.00 / 8.00 & 30.67 / 30.67 & 5 & 0.00 / 0.00 & 100.00 / 100.00 & 3.69 & 9.66 & 15.57 & 30.3 & 36.9 & 79.2\\
Deposit coin & 50 & 16.00 / 2.00 & 24.27 / 19.33 & 5 & 20.00 / 0.00 & 100.00 / 100.00 & 3.55 & 5.32 & 12.02 & 29.5 & 26.7 & 71.3\\
Insert key & 50 & 14.40 / 0.00 & 11.50 / 0.00 & 5 & 15.00 / 0.00 & 66.00 / 60.00 & 4.03 & 3.75 & 25.04 & 33.5 & 19.5 & 173.1\\
Build tower & 50 & 16.40 / 2.00 & 55.67 / 52.00 & 5 & 22.00 / 0.00 & 82.00 / 80.00 & 3.18 & 10.57 & 34.78 & 30.2 & 36.1 & 160.8\\
\midrule
\rowcolor{sensepurple!15}
\textbf{Long-Horizon} & 400 & 21.45 / 8.25 & 40.36 / 26.75 & 40 & 41.37 / 20.00 & 95.00 / 90.00 & 3.24 & 16.71 & 38.42 & 27.3 & 66.9 & 195.8\\
Put bottles into dustbin & 50 & 35.30 / 10.00 & 94.33 / 91.33 & 5 & 61.00 / 40.00 & 100.00 / 100.00 & 2.83 & 7.85 & 10.13 & 28.1 & 36.3 & 50.8\\
Play tic tac toe & 50 & 10.70 / 0.00 & 58.70 / 36.00 & 5 & 65.00 / 20.00 & 100.00 / 100.00 & 3.73 & 10.79 & 13.54 & 30.1 & 34.4 & 45.7\\
Classify objects & 50 & 69.00 / 56.00 & 31.20 / 20.67 & 5 & 51.00 / 40.00 & 100.00 / 100.00 & 3.14 & 17.84 & 33.42 & 29.8 & 80.3 & 151.0\\
Fill pen holder & 50 & 5.00 / 0.00 & 48.27 / 26.00 & 5 & 28.00 / 0.00 & 100.00 / 100.00 & 2.89 & 21.18 & 55.06 & 26.1 & 99.2 & 263.2\\
Fill egg holder & 50 & 5.50 / 0.00 & 7.20 / 0.67 & 5 & 27.00 / 20.00 & 70.00 / 60.00 & 4.58 & 14.59 & 84.66 & 37.0 & 66.2 & 658.9\\
Organize table & 50 & 39.50 / 0.00 & 49.50 / 9.33 & 5 & 45.00 / 0.00 & 90.00 / 60.00 & 2.92 & 30.53 & 65.50 & 10.2 & 121.2 & 237.4\\
Play stacking toy & 50 & 6.60 / 0.00 & 3.67 / 0.00 & 5 & 14.00 / 0.00 & 100.00 / 100.00 & 2.73 & 20.91 & 34.19 & 28.2 & 65.4 & 122.5\\
Make kong & 50 & 0.00 / 0.00 & 30.00 / 30.00 & 5 & 40.00 / 40.00 & 100.00 / 100.00 & 3.09 & 9.95 & 10.84 & 29.1 & 32.4 & 36.9\\
\midrule
\rowcolor{sensepurple!15}
\textbf{Memory} & 300 & 43.03 / 38.67 & 42.27 / 41.22 & 30 & 73.67 / 70.00 & 96.67 / 96.67 & 1.54 & 10.11 & 10.55 & 6.7 & 27.1 & 29.8\\
Cover blocks & 50 & 49.30 / 46.00 & 100.00 / 100.00 & 5 & 86.00 / 80.00 & 100.00 / 100.00 & 1.20 & 6.66 & 6.78 & 6.1 & 16.6 & 17.2\\
Match and pick from conveyor & 50 & 62.00 / 62.00 & 66.00 / 66.00 & 5 & 100.00 / 100.00 & 100.00 / 100.00 & 1.57 & 4.50 & 4.50 & 5.6 & 11.4 & 11.4\\
Swap T & 50 & 18.00 / 18.00 & 0.00 / 0.00 & 5 & 0.00 / 0.00 & 100.00 / 100.00 & 1.51 & 5.59 & 7.32 & 7.2 & 10.4 & 22.6\\
Press by number & 50 & 70.00 / 70.00 & 67.33 / 67.33 & 5 & 100.00 / 100.00 & 100.00 / 100.00 & 1.41 & 5.92 & 5.92 & 7.6 & 21.5 & 21.5\\
Imitate sorting sequence & 50 & 58.90 / 36.00 & 15.60 / 9.33 & 5 & 56.00 / 40.00 & 80.00 / 80.00 & 1.48 & 27.97 & 28.77 & 7.6 & 80.1 & 83.9\\
Swap blocks & 50 & 0.00 / 0.00 & 4.67 / 4.67 & 5 & 100.00 / 100.00 & 100.00 / 100.00 & 2.07 & 9.99 & 9.99 & 6.3 & 22.3 & 22.3\\
\midrule
\rowcolor{sensepurple!15}
\textbf{Open} & 400 & 34.36 / 31.00 & 24.09 / 22.92 & 40 & 17.63 / 15.00 & 73.00 / 67.50 & 8.13 & 8.64 & 39.91 & 105.3 & 42.1 & 227.2\\
Align blocks & 50 & 50.00 / 50.00 & 58.00 / 58.00 & 5 & 0.00 / 0.00 & 100.00 / 100.00 & 8.13 & 4.48 & 5.83 & 105.3 & 20.0 & 26.9\\
Solve equation & 50 & 40.00 / 40.00 & 50.00 / 50.00 & 5 & 20.00 / 20.00 & 100.00 / 100.00 & 8.13 & 4.07 & 6.21 & 105.3 & 17.7 & 26.5\\
Stack blocks by language & 50 & 48.00 / 40.00 & 14.67 / 11.33 & 5 & 8.00 / 0.00 & 100.00 / 100.00 & 8.13 & 5.20 & 9.58 & 105.3 & 20.2 & 35.5\\
General pickup & 50 & 84.00 / 84.00 & 60.67 / 60.67 & 5 & 100.00 / 100.00 & 100.00 / 100.00 & 8.13 & 3.46 & 3.46 & 105.3 & 16.8 & 16.8\\
Classify objects by language & 50 & 46.00 / 30.00 & 8.73 / 3.33 & 5 & 4.00 / 0.00 & 100.00 / 100.00 & 8.13 & 14.63 & 43.29 & 105.3 & 62.7 & 183.1\\
Pick from conveyor by image & 50 & 4.00 / 4.00 & 0.00 / 0.00 & 5 & 0.00 / 0.00 & 20.00 / 20.00 & 8.13 & 14.44 & 49.26 & 105.3 & 87.1 & 355.1\\
Store tools in toolbox & 50 & 2.50 / 0.00 & 0.67 / 0.00 & 5 & 5.00 / 0.00 & 40.00 / 0.00 & 8.13 & 10.36 & 150.49 & 105.3 & 47.5 & 909.0\\
Pour by language & 50 & 0.40 / 0.00 & 0.00 / 0.00 & 5 & 4.00 / 0.00 & 24.00 / 20.00 & 8.13 & 12.51 & 51.16 & 105.3 & 65.0 & 265.1\\
\midrule
\rowcolor{sensepurple!15}
\textbf{Overall / mean} & \textbf{2100} & \textbf{29.28 / 23.10} & \textbf{36.64 / 30.56} & \textbf{210} & \textbf{37.19 / 27.62} & \textbf{86.40 / 81.90} & 3.05 & 12.47 & 38.63 & 27.7 & 54.9 & 216.6\\
\bottomrule
\end{tabular}}
\vspace{3pt}
\begin{minipage}{\textwidth}\fontsize{8}{9.6}\selectfont
\emph{Note.} $N$ denotes the number of evaluation episodes. Performance entries report mean score (0--100) / success rate (\%). Group means average tasks within each group; the overall mean averages all 42 tasks. ICL costs average the study costs associated with each task's 5 evaluated layouts. Evaluation costs average all episodes, including successes, failures and timeouts, and exclude ICL costs; the $10\times$ costs include both initial execution and continuation. Tokens count input (including cached input) and output. Time excludes simulator action calls and trajectory replay. The overall ICL cost averages 34 task-specific estimates and 1 shared estimate for the Open tasks, while the overall evaluation cost averages 210 episodes.
\end{minipage}
\end{table}

\noindent\textbf{Observation and control settings.}
Our RoboDojo experiments use the dual-arm ARX X5 robot, and GPT-6-Astra receives $640\times480$ RGB images from head, left-wrist and right-wrist cameras, measured joint positions, gripper states and end-effector poses. Privileged information is withheld: object poses, depth images, segmentation masks, contact information, hidden simulator state, task source code and evaluator diagnostics. For each arm, GPT-6-Astra outputs a target end-effector position $(x,y,z)$ in meters in the environment coordinate frame, a target orientation as a unit quaternion $(w,x,y,z)$, and an open or close gripper command. Each target differs from the current measured pose by at most 5\,cm in position and 0.35\,rad in orientation. Inverse kinematics converts these end-effector targets into joint targets, with each control step limiting end-effector translation to 2\,cm, rotation to 0.1\,rad and each joint's rotation to 0.05\,rad. Control runs at 25\,Hz; an action request can repeat a target for 1--10 control steps. After each action request, GPT-6-Astra uses the returned images and measured robot state to choose the next action.

\noindent\textbf{Budgets and continuation.}
We evaluate GPT-6-Astra + ICL with the $1\times$ and $10\times$ step limits. Each task has a standard control-step limit $B$ ranging from 200 to 1,900 steps, with evaluation limits of $B$ and $10B$, respectively.
We first run each episode for up to $B$ control steps, or until earlier termination, and measure the $1\times$ result from this initial execution. If the episode has not terminated, we continue the same episode up to a total of $10B$ steps and record the $10\times$ result. Execution ends when the simulator terminates the episode or GPT-6-Astra stops.

\FloatBarrier
\subsection{RoboLab}
\noindent\textbf{Benchmark and tasks.}
RoboLab~\cite{yang2026robolab} evaluates task-generalist manipulation policies in high-fidelity simulation. Its RoboLab-120 task library comprises 120 language-conditioned tasks organized along three competency axes: visual recognition, relational understanding, and procedural reasoning. Tasks span three difficulty levels and include object selection, spatial rearrangement, stacking, and reorientation. The benchmark supports evaluation of policies trained on real-world data without requiring co-training on simulation data.

\noindent\textbf{Policies and evaluation metric.}
Table~\ref{tab:robolab_results} compares GPT-6-Astra with nine baselines spanning vision-language-action models (VLAs) and world action models (WAMs). We report success rate (SR), expressed as a percentage, together with policy type and model size. Baseline results are taken from the official RoboLab leaderboard for the default language instructions; each listed baseline is evaluated over 1,200 episodes. Model sizes follow the reported parameter counts. These published results provide reference points for the agent evaluation; the table does not establish a comparison at matched inference cost.


\begin{table}[!htbp]
\fontsize{8}{9.6}\selectfont
\setlength{\tabcolsep}{4pt}\renewcommand{\arraystretch}{1.15}

\centering
\caption{Overall performance on RoboLab. Success rate (SR) is reported in percent; higher is better. Policies are ordered by success rate, and the best result is shown in bold.}
\label{tab:robolab_results}

\begin{tabularx}{\linewidth}{@{}Xlrl@{}}
\toprule
Policy & Type & SR (\%) & Size \\
\midrule
\textbf{GPT-6-Astra 10x} & Agent & \textbf{69.2} & -- \\
\textbf{GPT-6-Astra 1x} & Agent & 57.1 & -- \\
\midrule
FLUX 3 Action~\cite{bfl2026flux3action} & WAM & 42.9 & 7B \\
Cosmos3-Nano-Policy~\cite{nvidia2026cosmos3} & WAM & 36.8 & 16B \\
$\pi_{0.5}$~\cite{pi2025pi05} & VLA & 28.0 & 3.3B \\
DreamZero~\cite{ye2026dreamzero} & WAM & 25.7 & 14B \\
Cosmos3-Edge-Policy~\cite{nvidia2026cosmos3} & WAM & 22.9 & 4B \\
$\pi_{0}$-FAST~\cite{pertsch2025fast} & VLA & 15.5 & 3B \\
GR00T N1.6~\cite{bjorck2025gr00t} & VLA & 7.2 & 3B \\
$\pi_{0}$~\cite{black2024pi0} & VLA & 5.0 & 3.3B \\
paligemma-binning~\cite{beyer2024paligemma} & VLA & 3.4 & 3B \\
\bottomrule
\end{tabularx}

\vspace{3pt}
\begin{minipage}{\linewidth}
\fontsize{8}{9.6}\selectfont
\emph{Note.} Baseline scores and model sizes follow the default-instruction results on the \href{https://research.nvidia.com/labs/srl/projects/robolab/leaderboard.html}{official RoboLab leaderboard}~\cite{yang2026robolab} (accessed 15 October 2026). Each listed baseline reports 1,200 evaluation episodes (120 tasks with 10 episodes per task). VLA: vision-language-action model; WAM: world action model. B denotes billions of parameters; -- indicates that model size is not reported here.
\end{minipage}
\end{table}

\noindent\textbf{Results.}
Under the standard ($1\times$) budget, GPT-6-Astra achieves a success rate of 57.1\%, exceeding all baselines listed in Table~\ref{tab:robolab_results}. It exceeds FLUX 3 Action, the strongest listed baseline at 42.9\%, by 14.2 percentage points. Among the listed VLAs, $\pi_{0.5}$ performs best at 28.0\%, trailing GPT-6-Astra by 29.1 points. The aggregate results show an advantage for GPT-6-Astra on this comparison, while task-level outcomes would be needed to identify which competency axes account for the difference.

\FloatBarrier
\subsection{Bench2Dex}
\noindent\textbf{Tasks and demonstrations.}
We evaluate all 26 Bench2Dex \cite{yang2026bench2dex} tasks across 12 arm--hand embodiments under Full randomization (\texttt{inv\_cov}, unseen split). The paired cohort contains three fixed episodes per task (indices 5--7), totaling 78 episodes. Preparation uses all 100 released seen-split replay recordings per task, including rendering variants of shared motion trajectories. Demonstrations expose four-camera RGB, active-joint positions, and demonstrated joint targets. Discrepancies between recorded and current task instructions are flagged, with the current instruction remaining authoritative.

\begin{table}[!htbp]
\fontsize{7}{8.4}\selectfont
\setlength{\tabcolsep}{1.5pt}\renewcommand{\arraystretch}{1.15}

\centering
\caption{Task-level performance and computational cost on Bench2Dex~\cite{yang2026bench2dex} under Full randomization. Tasks are grouped by embodiment following the benchmark paper. GPT-6-Astra uses three episodes per task (78 total) at the standard ($1\times$) and extended ($10\times$) interaction budgets. Token and time costs refer to GPT-6-Astra.}
\label{tab:bench2dex_results_costs}

\maxsizebox{\textwidth}{!}{%
\begin{tabular}{@{}llr rrr rrr rrr@{}}
\toprule
\multirow{2}{*}{Task}
& \multirow{2}{*}{Embodiment}
& \multirow{2}{*}{$N$}
& \multicolumn{3}{c}{Success rate (\%)}
& \multicolumn{3}{c}{Tokens (millions)}
& \multicolumn{3}{c}{Time (minutes)} \\
\cmidrule(lr){4-6}
\cmidrule(lr){7-9}
\cmidrule(lr){10-12}
& & &
$\pi_{0.5}$~\cite{pi2025pi05} & GPT-6-Astra $1\times$ & GPT-6-Astra $10\times$
& ICL & Test $1\times$ & Test $10\times$
& ICL & Test $1\times$ & Test $10\times$ \\
\midrule
Baking Tray Prep (32)
& \multirow{3}{*}{IIWA7 + Sharpa}
& 3 & 20.0 & 0.0 & 100.0 & 4.40 & 8.30 & 54.82 & 27.8 & 30.2 & 186.4 \\
Canned Food Tray Arrangement (26)
& & 3 & 6.0 & 0.0 & 33.3 & 2.61 & 9.01 & 52.91 & 22.0 & 23.5 & 144.0 \\
Jigsaw Puzzle Assembly (73)
& & 3 & 0.0 & 0.0 & 0.0 & 3.82 & 13.73 & 72.62 & 28.6 & 47.2 & 236.7 \\
\midrule
Gaming Desk Setup (80)
& JAKA ZU7 + DexHand021
& 3 & 40.0 & 0.0 & 0.0 & 3.55 & 14.16 & 59.65 & 34.9 & 40.4 & 189.0 \\
\midrule
Medicine Shoebox Pack (61)
& \multirow{2}{*}{Panda + Orca}
& 3 & 24.0 & 66.7 & 100.0 & 3.39 & 13.05 & 18.70 & 24.6 & 32.8 & 57.9 \\
Tool Box Loading (22)
& & 3 & 68.0 & 66.7 & 100.0 & 4.65 & 10.50 & 19.09 & 37.7 & 24.5 & 34.8 \\
\midrule
Sports Ball Cup Sort (64)
& Panda + Allegro
& 3 & 22.0 & 66.7 & 100.0 & 4.41 & 10.54 & 12.36 & 32.8 & 32.7 & 40.2 \\
\midrule
Faucet Cup Water Fill (67)
& \multirow{2}{*}{RM65 + Revo2}
& 3 & 18.0 & 33.3 & 100.0 & 3.65 & 13.49 & 41.29 & 36.4 & 32.7 & 108.5 \\
Stationery Category Sorting (24)
& & 3 & 0.0 & 0.0 & 66.7 & 3.25 & 14.69 & 99.48 & 26.7 & 37.6 & 300.0 \\
\midrule
Bimanual Piano Melody (79)
& \multirow{2}{*}{xArm7 + Ability}
& 3 & 4.0 & 66.7 & 100.0 & 3.84 & 8.60 & 11.28 & 26.5 & 18.4 & 28.4 \\
Wine Glass Plate Balance (03)
& & 3 & 38.0 & 0.0 & 0.0 & 4.32 & 10.63 & 11.17 & 30.2 & 24.0 & 25.1 \\
\midrule
Cleaner Box Loading (09)
& \multirow{3}{*}{xArm7 + LEAP}
& 3 & 48.0 & 0.0 & 100.0 & 2.99 & 8.66 & 25.29 & 23.2 & 26.0 & 55.3 \\
Shoebox Accessory Pack (62)
& & 3 & 48.0 & 0.0 & 0.0 & 2.70 & 9.88 & 23.59 & 26.6 & 24.7 & 72.8 \\
Toilet Lid Cleaner Pour (51)
& & 3 & 46.0 & 33.3 & 66.7 & 3.52 & 6.46 & 6.77 & 24.0 & 18.3 & 19.6 \\
\midrule
Breadbasket Fast-Food Loading (60)
& \multirow{3}{*}{UR5 + RH5DG2}
& 3 & 6.0 & 0.0 & 66.7 & 3.32 & 15.81 & 39.18 & 30.8 & 41.0 & 98.8 \\
Citrus Plate Loading (07)
& & 3 & 64.0 & 0.0 & 33.3 & 3.56 & 8.21 & 11.25 & 28.3 & 24.0 & 29.7 \\
Fridge Wine Interhand Pour (34)
& & 3 & 10.0 & 0.0 & 66.7 & 3.06 & 14.95 & 74.84 & 29.6 & 32.3 & 182.8 \\
\midrule
Fruit Bowl Loading (06)
& \multirow{3}{*}{UR5 + RH56DFX}
& 3 & 38.0 & 0.0 & 33.3 & 2.27 & 9.23 & 18.62 & 21.5 & 25.2 & 56.0 \\
Screwdriver Box \& Hammer (12)
& & 3 & 32.0 & 66.7 & 100.0 & 4.14 & 7.72 & 7.93 & 27.8 & 21.5 & 21.8 \\
Trash Disposal (42)
& & 3 & 36.0 & 66.7 & 100.0 & 5.32 & 5.86 & 12.06 & 36.8 & 16.0 & 22.9 \\
\midrule
Fridge Fruit Shelf Sorting (43)
& \multirow{2}{*}{UR5 + Shadow}
& 3 & 4.0 & 33.3 & 66.7 & 3.06 & 16.60 & 32.93 & 27.3 & 37.2 & 89.3 \\
Soup Serving (76)
& & 3 & 0.0 & 0.0 & 0.0 & 3.88 & 20.24 & 79.85 & 31.5 & 60.7 & 268.3 \\
\midrule
Frypan Stand \& Pour (08)
& \multirow{2}{*}{UR5 + Schunk}
& 3 & 12.0 & 0.0 & 0.0 & 3.02 & 12.20 & 28.57 & 25.0 & 34.7 & 71.4 \\
Microwave Bowl Loading (44)
& & 3 & 62.0 & 33.3 & 100.0 & 4.63 & 12.14 & 27.94 & 29.7 & 26.7 & 63.7 \\
\midrule
Ball Box Loading (27)
& \multirow{2}{*}{UR5 + Wuji}
& 3 & 8.0 & 100.0 & 100.0 & 4.08 & 8.92 & 8.92 & 31.5 & 18.2 & 18.2 \\
Condiment Box Loading (21)
& & 3 & 8.0 & 33.3 & 66.7 & 3.50 & 8.75 & 24.49 & 26.7 & 21.8 & 57.2 \\
\midrule
\rowcolor{sensepurple!15}
\multicolumn{2}{l}{\textbf{Overall / mean}}
& \textbf{78}
& \textbf{25.5} & \textbf{25.6} & \textbf{61.5}
& 3.65 & 11.24 & 33.68
& 28.8 & 29.7 & 95.3 \\
\bottomrule
\end{tabular}%
}

\vspace{3pt}
\begin{minipage}{\textwidth}
\fontsize{8}{9.6}\selectfont
\emph{Note.} The $\pi_{0.5}$ is the current state-of-the-art baseline under Full randomization reported on the \href{https://bench2dex.github.io/}{official Bench2Dex project website}~\cite{yang2026bench2dex} (accessed 6 October 2026), achieving 25.5\% SR (331/1300 successes). It is evaluated on 50 episodes per task under the standard budget, separately from the GPT-6-Astra episodes. ICL costs are incurred once per task; test costs are means over all three episodes, including failures. The $10\times$ costs include the original $1\times$ execution and its extension, excluding preparation and discarded/interrupted attempts. Tokens include cached input; reported test token counts are lower bounds because usage accounting is partial. Times measure tool-inclusive inference wall time, excluding simulator stepping and replay. The final row averages ICL costs over 26 tasks and test costs over 78 episodes.
\end{minipage}
\end{table}

\noindent\textbf{Observation and control settings.}
GPT-6-Astra receives stereo left/right and wrist left/right RGB images at $640\times480$, measured active-joint positions, robot kinematics, and camera calibration. Camera intrinsics and camera-to-robot-base transforms are obtained from the simulator after randomization. We supply kinematics and calibration as setup information readily accessible in a calibrated robotic deployment, consistent with our objective of assessing GPT-6-Astra's capabilities given available test-time information. Object states, depth, segmentation, contacts, and evaluator feedback are excluded. Actions are absolute active-joint targets executed at 20\,Hz, with three 60\,Hz physics steps per target. GPT-6-Astra chooses the action-chunk length, and our wrapper adds no interpolation or clipping.

\noindent\textbf{Budgets and continuation.}
The standard budget follows the benchmark, $B_t=\lceil 1.5E_t/3 \rceil$, where $E_t$ is the expert reference duration in physics steps. $B_t$ ranges from 593 to 1,542 control steps. SR uses official stable-success early stopping, with a 0.5\,s dwell except for Faucet Cup Water Fill, which requires 3\,s.
Only standard-budget timeouts receive additional interaction up to a total of $10B_t$, while successes and other terminal failures retain their outcomes. Because the original $1\times$ runs were already complete, we reconstruct timeout endpoints using the original reset configuration and exact recorded-action replay, then resume the original episode context. Replay must match archived numerical states and boundary stage histories within fixed tolerances. The continuation additionally permits an observation-grounded explicit give-up, counted as failure.

\noindent\textbf{Results and comparison.}
GPT-6-Astra achieves 25.6\% SR at $1\times$, comparable to the published $\pi_{0.5}$ baseline of 25.5\%, and reaches 61.5\% at $10\times$, an increase of 35.9 percentage points. The baseline uses 50 episodes per task, evaluated separately from our three episodes per task. Reported test-token counts in Table~\ref{tab:bench2dex_results_costs} are lower bounds because usage accounting is partial; tokens-per-success values derived from these counts are therefore also lower bounds and do not establish a cross-benchmark token-cost ranking.

\FloatBarrier
\subsection{DexJoCo}
\noindent\textbf{Tasks and demonstrations.}
We evaluate all 11 DexJoCo \cite{wang2026dexjoco} tasks (six single-arm and five bimanual) under \texttt{rand\_full}, using five fixed test seeds per task (100000--100004; 55 episodes). This regime randomizes object placement and table height, external camera pose, lighting, and table texture, while dynamics randomization is disabled. For each task, preparation has access to 100 released \texttt{rand\_full} replay recordings through a restricted reader that exposes robot proprioception, demonstrated actions, timestamps, and selected RGB videos. GPT-6-Astra also receives the pinned robot model. Each task's resulting notes, reference examples, and control utilities are frozen as one package before testing.


\begin{table}[!htbp]
\fontsize{7}{8.4}\selectfont
\setlength{\tabcolsep}{2.5pt}\renewcommand{\arraystretch}{1.15}

\centering
\caption{Task-level performance and computational cost on DexJoCo~\cite{wang2026dexjoco} under rand-full randomization. GPT-6-Astra uses five test episodes per task at the standard ($1\times$) and extended ($10\times$) interaction budgets. Token and time costs refer to GPT-6-Astra.}
\label{tab:dexjoco_rand_full_results_costs}

\maxsizebox{0.9\linewidth}{!}{
\begin{tabular}{@{}l r rrr rrr rrr@{}}
\toprule
Task & $N$
& \multicolumn{3}{c}{Success rate (\%)}
& \multicolumn{3}{c}{Tokens (millions)}
& \multicolumn{3}{c}{Time (minutes)} \\
\cmidrule(lr){3-5}
\cmidrule(lr){6-8}
\cmidrule(lr){9-11}
& & $\pi_{0.5}$~\cite{pi2025pi05} & GPT-6-Astra $1\times$ & GPT-6-Astra $10\times$
& ICL & Test $1\times$ & Test $10\times$
& ICL & Test $1\times$ & Test $10\times$ \\
\midrule
Hammer Nail  & 5 & 17.3 & 100.0 & 100.0 & 4.87 & 4.71 & 4.71 & 35.3 & 13.5 & 13.5 \\
Click Mouse  & 5 & 54.7 & 100.0 & 100.0 & 4.98 & 2.84 & 2.84 & 27.8 & 6.1 & 6.1 \\
Pick Bucket  & 5 & 78.7 & 100.0 & 100.0 & 3.36 & 9.21 & 9.21 & 22.1 & 16.7 & 16.7 \\
Pinch Tongs  & 5 & 18.7 & 80.0 & 80.0 & 3.95 & 3.00 & 3.00 & 29.6 & 6.7 & 6.7 \\
Fold Glasses & 5 & 39.3 & 60.0 & 100.0 & 4.00 & $15.96$ & $19.46$ & 24.4 & 32.6 & 41.5 \\
Water Plant  & 5 & 75.3 & 100.0 & 100.0 & 5.33 & 1.64 & 1.64 & 32.3 & 4.9 & 4.9 \\
\midrule
Unlock iPad & 5 & 32.7 & 20.0 & 20.0 & 4.38 & $3.66$ & $3.66$ & 32.6 & 10.7 & 10.7 \\
Hanoi       & 5 & 33.3 & 100.0 & 100.0 & 4.09 & 5.87 & 5.87 & 28.1 & 12.9 & 12.9 \\
Assembly    & 5 & 0.7 & 60.0 & 80.0 & 6.29 & $16.76$ & $31.95$ & 37.2 & 48.2 & 96.7 \\
Microwave   & 5 & 69.3 & 100.0 & 100.0 & 9.80 & 6.24 & 6.24 & 38.0 & 13.4 & 13.4 \\
Photograph  & 5 & 23.3 & 40.0 & 80.0 & 7.26 & 12.46 & 27.46 & 43.0 & 30.3 & 85.6 \\
\midrule
\rowcolor{sensepurple!15}
\textbf{Overall / mean}
& \textbf{55} & \textbf{40.3} & \textbf{78.2} & \textbf{87.3}
& 5.30 & $7.49$ & $10.55$
& 31.9 & 17.8 & 28.0 \\
\bottomrule
\end{tabular}%
}

\vspace{3pt}
\begin{minipage}{\linewidth}
\fontsize{8}{9.6}\selectfont
\emph{Note.} The $\pi_{0.5}$ success rates are the rand-full results in DexJoCo Table II~\cite{wang2026dexjoco}, which reports a 40.3\% mean over 11 tasks and evaluates three training seeds with 50 test episodes per task per seed. GPT-6-Astra uses five test seeds per task.
ICL preparation is incurred once per task. Test costs average all five scored episodes, including failures and the scored step-zero Pinch Tongs action-envelope abort. $10\times$ test costs include the original $1\times$ session and any continuation, excluding preparation and discarded invalid attempts. Tokens include cached input. 
Time is Codex session wall time minus simulator stepping, including other model and tool work. The final row averages ICL costs over 11 tasks and test costs over 55 episodes.
\end{minipage}
\end{table}

\noindent\textbf{Observation and control settings.}
At test time, GPT-6-Astra receives $640\times640$ RGB from one external camera and one wrist camera for single-arm tasks, or two wrist cameras for bimanual tasks. Public robot state contains end-effector pose and 16 hand-joint values per arm (23 or 46 values). Object state, depth, contacts, rewards, and success flags are withheld. Actions specify absolute world-frame end-effector position and rotation vector plus 16 hand-joint targets per arm (22 or 44 values), executed at 30\,Hz. GPT-6-Astra may submit 1--30 actions per decision. The interface converts rotation vectors to the quaternion format expected by the simulator.

\noindent\textbf{Budgets and continuation.}
DexJoCo's task-specific standard limits range from 1,000 to 1,500 action steps. Success is determined by the simulator's task-specific success flag. Only episodes that exhaust the standard limit without a terminal outcome continue toward $10B$, for which a fresh GPT-6-Astra session resumes the same live simulator instance with the episode's observation and action history. No reset or trajectory replay is used for these continuations. For \texttt{click\_mouse}, we reproduce the benchmark client's 30-step initial alignment and count those steps against the task budget.

\noindent\textbf{Results and comparison.}
GPT-6-Astra achieves 78.2\% SR at $1\times$, exceeding the published $\pi_{0.5}$ baseline of 40.3\% by 37.9 percentage points, and reaches 87.3\% at $10\times$, a further increase of 9.1 percentage points. The baseline reports results over three training seeds with 50 test episodes per task per seed, while GPT-6-Astra uses five test seeds per task.

\FloatBarrier
\subsection{LHPR-VLN}
\noindent\textbf{Benchmarks and tasks.}
LHPR-VLN~\cite{song2025lhvln} evaluates long-horizon vision-language navigation. Given a global natural-language instruction, the policy must reach multiple navigation goals in a strictly specified order. Following these instructions requires visually grounding successive goals, tracking completed stages, and using earlier observations to guide subsequent movements.

\noindent\textbf{Evaluation subset.}
The official unseen test split contains 205 episodes across 27 scenes. Many episodes contain subtask target objects that are not visible along the reference trajectories. We manually checked target visibility along each episode's reference trajectory and retained an episode only when every subtask target object was visible. The resulting subset contains 114 episodes across 25 scenes, with two to four goals per episode. The complete retained episode list appears below.

\noindent\textbf{Protocol and metrics.}
Each episode starts in a fresh session with the global instruction and a frozen ICL workspace containing exploration notes, code, and observations from training scenes that do not overlap with the test scenes. Test episodes do not share interaction history. During navigation, the policy receives three RGB views at a resolution of $366\times366$, facing left, forward, and right at $+60^\circ$, $0^\circ$, and $-60^\circ$. Available actions are 0.25\,m forward motion, $30^\circ$ left or right turns, and \texttt{STOP}. Three consecutive blocked forward actions trigger a random turn and a forward action, both counted toward the action budget. Ground-truth trajectories, target coordinates, and scores are withheld. To submit a goal, the policy issues \texttt{STOP}, which advances the stage regardless of success. A submission succeeds within 1\,m geodesic distance of the original target or a matching target object in the specified room. Environment feedback identifies the current stage without revealing submission success. The $1\times$ and $10\times$ settings allow 500 and 5,000 actions per sequence. Extended evaluation retains successful $1\times$ outcomes and evaluates only initially unsuccessful episodes with the larger budget, preserving earlier scores when they are higher. Metrics are averaged over all 114 episodes. SR measures full-sequence success. ISR, CSR, and CGT describe individual-goal and sequential success, with definitions in the table note.


\begin{table}[!htbp]
\fontsize{8}{9.6}\selectfont
\setlength{\tabcolsep}{4pt}\renewcommand{\arraystretch}{1.15}

\centering
\caption{Overall performance on LHPR-VLN~\cite{song2025lhvln}. SR, ISR, CSR, and CGT are percentages, with higher values indicating better performance. Token consumption is reported in millions. Policies are ordered by SR. Bold values mark the highest navigation scores.}
\label{tab:lhpr_vln_results}

\begin{tabularx}{\linewidth}{@{}Xlrrrrlr@{}}
\toprule
Policy & Type & SR & ISR & CSR & CGT & Size & Tokens \\
\midrule
\textbf{GPT-6-Astra Max + ICL $10\times$} & Agent & \textbf{28.95} & \textbf{62.87} & \textbf{52.97} & \textbf{52.58} & -- & 2.587 \\
\textbf{GPT-6-Astra Max + ICL $1\times$} & Agent & 21.05 & 54.68 & 44.55 & 43.91 & -- & 1.317 \\
\midrule
Claude Code Opus-5 & Agent & 5.26 & 26.87 & 17.82 & 17.84 & -- & -- \\
FantasyVLN~\cite{zuo2026fantasyvln} & VLN & 2.44 & 11.01 & 9.64 & 8.99 & 7B & -- \\
Aux-think~\cite{wang2025auxthink} & VLN & 0.65 & 3.16 & 2.04 & 1.47 & 8B & -- \\
MGDM~\cite{song2025lhvln} & VLN & 0.00 & 2.35 & 1.65 & 2.92 & 7B & -- \\
\bottomrule
\end{tabularx}

\vspace{3pt}
\begin{minipage}{\linewidth}
\fontsize{8}{9.6}\selectfont
\emph{Note.} SR is the success rate for complete sequences. ISR is the independent success rate for individual goals. CSR is the conditional success rate, which gives greater weight to successes that follow a successful subtask. CGT further weights CSR by ground-truth path length. Tokens are mean test-time input plus output tokens per episode, including failures. Cached input and reasoning output are counted once. The $10\times$ cost includes $1\times$ consumption and additional evaluation for initially unsuccessful episodes. Exploration and technical retries are excluded. -- indicates an unreported value.
\end{minipage}
\end{table}

\noindent\textbf{Results.}
GPT-6-Astra Max + ICL achieves 21.05\% SR at $1\times$, compared with 5.26\% for Claude Code Opus-5 and 2.44\% for FantasyVLN in Table~\ref{tab:lhpr_vln_results}. Extended evaluation raises SR to 28.95\%, increasing completed episodes from 24 to 33. ISR rises from 54.68\% to 62.87\%, while CSR and CGT improve by 8.42 and 8.67 percentage points. Average test-time token consumption increases from 1.317\,M to 2.587\,M, or $1.96\times$ the initial cost.

\noindent\textbf{Discussion.}
Full-sequence success remains below 30\% under extended evaluation despite individual-goal success above 60\%. Reaching some goals improves ISR, whereas SR requires every goal to be completed in order. A failed submission advances the sequence irreversibly, making goal verification and progress tracking consequential. Aggregate scores alone do not establish how much either contributes to the remaining failures. Baseline comparisons do not control for inference cost, and the reported token consumption should be considered alongside success rates.

{\normalsize\raggedright
\setlength{\parskip}{3pt}
\noindent\textbf{Episode IDs.} Each batch/index pair below identifies an episode as \texttt{batch\_b\_episode\_e}, where $b$ is the batch number and $e$ is the listed episode index.\par\nobreak
\noindent\textbf{Batch 1:} \texttt{2, 10, 19, 20, 55, 56, 60, 61, 62, 65, 81, 82, 84, 94, 103, 119, 128, 130}\par
\noindent\textbf{Batch 2:} \texttt{1, 17, 20, 35, 52, 54, 69, 78, 85, 87, 88, 99, 103}\par
\noindent\textbf{Batch 3:} \texttt{10, 24, 26, 36, 44, 46, 63, 69, 83, 86, 91, 97, 101, 103, 106, 107, 112}\par
\noindent\textbf{Batch 4:} \texttt{31, 32, 36, 58, 66, 68, 110, 111, 112, 114}\par
\noindent\textbf{Batch 5:} \texttt{3, 4, 7, 16, 21, 28, 38, 40, 45, 52, 53, 64, 72, 74, 75, 86, 88, 100}\par
\noindent\textbf{Batch 6:} \texttt{34, 37}\par
\noindent\textbf{Batch 7:} \texttt{13, 47, 56, 66, 74, 82, 83, 85, 99, 120, 123, 146, 160}\par
\noindent\textbf{Batch 8:} \texttt{5, 30, 63, 66, 77, 82, 85, 100, 112, 115, 134, 141, 146, 154, 156, 161, 166, 180, 191, 205, 206, 211, 223}\par
}

\subsection{VLN-CE}
\noindent\textbf{Benchmarks and tasks.}
VLN-CE~\cite{krantz2020vlnce} evaluates instruction-guided navigation in continuous indoor environments. Given a natural-language route instruction, the policy must navigate from a specified starting pose to a single goal and stop. This requires grounding landmarks, tracking progress along the route, and choosing movements from visual observations without a predefined navigation graph.

\noindent\textbf{Evaluation subsets.}
R2R~\cite{anderson2018r2r} uses the 100-episode val-unseen subset from Open-Nav~\cite{qiao2025opennav}, which covers 10 scenes and is also adopted by SmartWay~\cite{shi2025smartway} and AgentCanvas~\cite{agentcanvas2026}. For RxR~\cite{ku2020rxr}, we select 100 English guide episodes from the val-unseen split of 3,669 English episodes across 11 scenes, matching the R2R subset's episode counts in each of the same 10 scenes. With seed 0, we draw 20,000 stratified 100-episode subsets and retain the candidate with the smallest sum of two-sample Kolmogorov--Smirnov distances to the full English split for instruction word count and reference-path length. We then attempt 60,000 random within-scene episode replacements, accepting a change only if it reduces this sum. Instruction length is the whitespace-separated word count, and reference-path length is the sum of Euclidean distances between consecutive reference waypoints. The complete episode lists for both subsets appear below.

\noindent\textbf{Protocol and metrics.}
Each episode starts in a fresh session with a frozen ICL workspace prepared using training scenes that do not overlap with the test scenes. Test episodes do not share interaction history. Observations contain a single forward-facing RGB image at $512\times512$ resolution. Forward actions move 0.25\,m. R2R uses $15^\circ$ left or right turns, while RxR uses $30^\circ$ turns and permits $30^\circ$ upward or downward camera tilts. Ground-truth routes, goal coordinates, and evaluation scores are withheld. \texttt{STOP} terminates the episode and succeeds within 3\,m geodesic distance of the goal. The $1\times$ and $10\times$ limits are 500 and 5,000 actions. Extended evaluation is applied to initially unsuccessful episodes. Reported extended results retain prior successes and the better per-episode value of each metric. Metrics are averaged over each 100-episode subset. NE is the final geodesic distance to the goal. OS is the oracle success rate, measuring whether the trajectory ever enters the 3\,m success region. SR measures successful completion, and SPL~\cite{anderson2018evaluation} weights success by path efficiency. nDTW~\cite{ilharco2019ndtw} measures alignment with the reference trajectory using normalized dynamic time warping.

\begin{table}[!htbp]
\footnotesize
\setlength{\tabcolsep}{2.5pt}\renewcommand{\arraystretch}{1.15}

\centering
\caption{Overall performance on VLN-CE~\cite{krantz2020vlnce} R2R and RxR val-unseen. OS, SR, SPL, and nDTW are percentages, and NE is in meters. Bold values mark the best reported navigation scores.}
\label{tab:vlnce_results}

\begin{tabularx}{\linewidth}{@{}Xlllrrrrrrrr@{}}
\toprule
\multirow{2}{*}{Policy} & \multirow{2}{*}{Size} & \multirow{2}{*}{Type} & \multirow{2}{*}{Observation} & \multicolumn{4}{c}{R2R val-unseen} & \multicolumn{4}{c}{RxR val-unseen} \\
\cmidrule(lr){5-8}\cmidrule(lr){9-12}
& & & & NE$\downarrow$ & OS$\uparrow$ & SR$\uparrow$ & SPL$\uparrow$ & NE$\downarrow$ & SR$\uparrow$ & SPL$\uparrow$ & nDTW$\uparrow$ \\
\midrule
\textbf{GPT-6-Astra Max + ICL $10\times$} & -- & Agent & Mono. RGB & \textbf{2.34} & \textbf{91} & \textbf{88} & \textbf{76.98} & \textbf{1.36} & \textbf{90} & \textbf{73.52} & \textbf{82.98} \\
\textbf{GPT-6-Astra Max + ICL $1\times$} & -- & Agent & Mono. RGB & 3.45 & 84 & 80 & 68.47 & 2.08 & 82 & 65.61 & 77.85 \\
\midrule
GPT-6-Astra Ultra~\cite{dai2026lightsup} & -- & Agent & Mono. RGB & 2.9 & 83.7 & 81.3 & 71.5 & -- & -- & -- & -- \\
Robostral Navigate~\cite{bounhar2026robostral} & 8B + 121M & VLN & Mono. RGB & 3.2 & 81.3 & 77.4 & 74.2 & 3.47 & 75.1 & 68.7 & -- \\
Qwen-RobotNav-8B~\cite{zhang2026qwenrobotnav} & 8B & VLN & Pano. RGB & 3.53 & 78.5 & 72.1 & 66.6 & 3.58 & 76.5 & 65.7 & 72.5 \\
ABot-N1~\cite{gong2026abotn1} & 4B + 2B & VLN & 3-view RGB & 3.32 & 75.2 & 70.9 & 67.5 & 3.13 & 73.9 & 63.9 & -- \\
Image2Nav~\cite{wang2026image2sim} & 4B & VLN & Mono. RGB & 3.71 & 76.1 & 70.3 & 65.6 & 3.74 & 70.7 & 59.1 & 71.8 \\
GroundingVLN~\cite{li2026groundingvln} & 4B & VLN & 3-view RGB-D & 3.66 & 74.8 & 69.9 & 64.1 & 3.54 & 75.1 & 62 & -- \\
OmniNav~\cite{xue2026omninav} & 3B & VLN & 3-view RGB & 3.74 & 74.6 & 69.5 & 66.1 & 3.77 & 73.6 & 62 & -- \\
LightNav-0~\cite{wang2026lightnav0} & 4B & VLN & Mono. RGB & 3.91 & 73.7 & 68.5 & 62.8 & 3.66 & 73.6 & 64.5 & 67.4 \\
AstraNav-World~\cite{chen2025astranavworld} & 3B + 5B & WAM & 3-view RGB & 3.86 & 73.9 & 67.9 & 65.4 & 3.82 & 72.9 & 61.5 & -- \\
SEDualVLN~\cite{huang2026sedualvln} & 7B + GPT-4o & VLN & Mono. RGB & 3.75 & 73.7 & 67.3 & 62.5 & 4.12 & 63.9 & 52.4 & 72.8 \\
AgentVLN~\cite{xin2026agentvln} & 3B & VLN & Mono. RGB-D & 3.88 & 73.5 & 67.2 & 64.7 & 3.92 & 69.5 & 61.3 & -- \\
TAMP-Nav~\cite{feng2026embodiednavigator} & 7B & VLN & 4-view RGB-D & 3.85 & 74.5 & 66.2 & 58.8 & 4.32 & 65.7 & 56.9 & -- \\
AwareVLN~\cite{guo2026awarevln} & 8B & VLN & Mono. RGB & 4.02 & 73.5 & 65.4 & 55.1 & 3.95 & 67.6 & 56.1 & -- \\
DualVLN~\cite{wei2026dualvln} & 7B & VLN & Mono. RGB & 4.05 & 70.7 & 64.3 & 58.5 & 4.58 & 61.4 & 51.8 & 70.0 \\
\bottomrule
\end{tabularx}

\vspace{3pt}
\begin{minipage}{\linewidth}
\fontsize{8}{9.6}\selectfont
\emph{Note.} Higher OS, SR, SPL, and nDTW and lower NE indicate better performance. Mono. and Pano. denote monocular and panoramic observations. RGB-D includes depth. Image2Nav uses a $180^\circ$ horizontal field of view. Sizes denote reported backbone or component sizes. -- indicates an unreported value.
\end{minipage}
\end{table}

\noindent\textbf{Results.}
Extended evaluation raises SR from 80\% to 88\% on R2R and from 82\% to 90\% on RxR, adding eight successes per subset in Table~\ref{tab:vlnce_results}. SPL improves by 8.51 and 7.91 percentage points, respectively. NE decreases from 3.45\,m to 2.34\,m on R2R and from 2.08\,m to 1.36\,m on RxR. R2R OS rises from 84\% to 91\%, while RxR nDTW increases from 77.85\% to 82.98\%.

\noindent\textbf{Discussion.}
On R2R, OS exceeds SR by three percentage points under extended evaluation, identifying episodes that reach the goal region without stopping successfully. The RxR nDTW gain indicates closer alignment with reference routes. SPL remains below SR on both datasets, indicating that successful outcomes can still involve detours. Our results cover 100 episodes per dataset. Differences in episode coverage, observation access, and inference budgets limit direct comparisons with the listed baselines.

{\normalsize\raggedright
\setlength{\parskip}{3pt}
\noindent\textbf{Episode IDs.}\par\nobreak
\noindent\textbf{R2R:} \texttt{7, 11, 13, 40, 42, 52, 70, 94, 116, 140, 150, 156, 166, 171, 176, 181, 187, 190, 191, 207, 218, 226, 232, 244, 247, 259, 265, 275, 308, 312, 321, 330, 338, 348, 362, 371, 377, 387, 403, 411, 423, 432, 439, 447, 454, 461, 469, 479, 513, 516, 526, 531, 546, 550, 559, 568, 576, 586, 602, 609, 620, 643, 655, 670, 677, 698, 705, 715, 721, 739, 743, 748, 755, 765, 781, 787, 804, 810, 821, 824, 842, 1051, 1056, 1061, 1071, 1077, 1084, 1085, 1087, 1092, 1106, 1117, 1133, 1139, 1142, 1148, 1284, 1301, 1307, 1406}\par
\noindent\textbf{RxR:} \texttt{198, 311, 377, 683, 886, 901, 912, 1108, 1109, 1196, 1575, 1719, 1866, 1914, 2175, 2220, 2258, 2264, 2274, 2463, 2485, 2561, 2638, 2694, 2800, 2823, 2917, 3468, 3514, 3550, 3564, 3591, 3644, 3673, 4027, 4061, 4101, 4150, 4252, 4312, 4326, 4449, 4502, 4517, 4635, 4753, 5247, 5348, 5431, 5499, 5503, 5843, 5922, 5950, 5968, 6004, 6134, 6180, 6237, 6289, 6319, 6362, 6597, 6623, 6632, 6703, 6968, 6995, 7008, 7121, 7234, 7284, 7433, 7471, 7488, 7569, 7732, 7733, 7741, 7819, 8127, 8246, 8259, 8340, 8686, 8928, 9142, 9156, 9242, 9457, 9514, 9632, 9803, 9848, 10025, 10188, 10387, 10789, 10917, 10971}\par
}

\subsection{SIMPLE}
\label{app:simple_evaluation}
\noindent\textbf{Tasks and evaluation scope.}
SIMPLE~\cite{wei2026simple} evaluates humanoid loco-manipulation on a Unitree G1 with two Dex3 hands. Our study uses the six-task evaluation set reported in Table~1 of the benchmark paper: XMovePick, BendPick, Handover, Tabletop, XMoveBendPick, and PickBetweenTables. Each task has three difficulty levels: L0 varies table texture and distractor type and placement; L1 adds extreme lighting changes; L2 further perturbs the target object's initial position. With 10 released evaluation instances per task--level, this set contains 180 candidate episodes. Outcome coverage differs across the four GPT-6-Astra settings (zero-shot or ICL, at $1\times$ or $10\times$), as specified below.


\begin{table}[!htbp]
\fontsize{7}{8.4}\selectfont
\setlength{\tabcolsep}{1.25pt}\renewcommand{\arraystretch}{1.15}

\centering
\caption{Per-level success on the six-task evaluation set from Table~1 of SIMPLE~\cite{wei2026simple}: GPT-6-Astra vs. published baselines, under the standard ($1\times$) and extended ($10\times$) interaction budgets. Each cell reports successes / judged episodes at levels L0/L1/L2 under the indicated budget; published baselines report $1\times$ only. The Matched-8 total compares the same eight complete task--level cells (80 episodes) across all rows. Six-task totals are provided for reference only.}
\label{tab:simple_perlevel}

\maxsizebox{\textwidth}{!}{%
\begin{tabular}{@{}l *{18}{c} cc@{}}
\toprule
\multirow{2}{*}{Method}
& \multicolumn{3}{c}{XMovePick}
& \multicolumn{3}{c}{BendPick}
& \multicolumn{3}{c}{Handover}
& \multicolumn{3}{c}{Tabletop}
& \multicolumn{3}{c}{XMoveBendPick}
& \multicolumn{3}{c}{PickBetweenTables}
& \multirow{2}{*}{\shortstack{Six-task total\\ (reference only)}}
& \multirow{2}{*}{\shortstack{Matched-8\\ total}} \\
\cmidrule(lr){2-4}\cmidrule(lr){5-7}\cmidrule(lr){8-10}\cmidrule(lr){11-13}\cmidrule(lr){14-16}\cmidrule(lr){17-19}
& L0 & L1 & L2 & L0 & L1 & L2 & L0 & L1 & L2 & L0 & L1 & L2 & L0 & L1 & L2 & L0 & L1 & L2 & & \\
\midrule
\textbf{GPT-6-Astra, ZS $\cdot$ 1$\times$}
& 9/10 & 9/10 & 10/10 & 7/10 & 7/10 & 5/10 & 7/10 & 4/10 & 4/10 & 9/10 & 9/10 & 5/10 & 9/10 & 9/10 & 7/10
& 0/6\textsuperscript{a} & 0/7\textsuperscript{a} & 0/4\textsuperscript{a} & 110/167 (65.9\%) & 65/80 (81.3\%) \\
\textbf{GPT-6-Astra, ICL $\cdot$ 1$\times$}
& 10/10 & 10/10 & 8/10 & 10/10 & 10/10 & 8/10 & 9/10 & 9/10 & 6/10 & 10/10 & 10/10 & 9/10 & -- & -- & --
& 2/3\textsuperscript{a} & 0/2\textsuperscript{a} & 0/1\textsuperscript{a} & 111/126 (88.1\%)\textsuperscript{b} & \underline{77/80 (96.3\%)} \\
\textbf{GPT-6-Astra, ZS $\cdot$ 10$\times$}
& 10/10 & 10/10 & 10/10 & 10/10 & 10/10 & 8/10 & 7/7\textsuperscript{u} & 4/4\textsuperscript{u} & 5/5\textsuperscript{u} & 10/10 & 10/10 & 9/10 & 10/10 & 10/10 & 10/10
& 6/6\textsuperscript{a} & 7/7\textsuperscript{a} & 4/4\textsuperscript{a} & 150/153 (98.0\%) & \textbf{79/80 (98.8\%)} \\
\textbf{GPT-6-Astra, ICL $\cdot$ 10$\times$}
& 10/10 & 10/10 & 10/10 & 10/10 & 10/10 & 8/8\textsuperscript{u} & 9/9\textsuperscript{u} & 9/9\textsuperscript{u} & 6/6\textsuperscript{u} & 10/10 & 10/10 & 9/10 & -- & -- & --
& 3/3\textsuperscript{a} & 2/2\textsuperscript{a} & 1/1\textsuperscript{a} & 117/118 (99.2\%)\textsuperscript{b} & \textbf{79/80 (98.8\%)} \\
\midrule
Psi-0~\cite{wei2026psi0} $\cdot$ 1$\times$
& 10/10 & 10/10 & 6/10 & 10/10 & 10/10 & 10/10 & 7/10 & 7/10 & 10/10 & 10/10 & 10/10 & 8/10 & 10/10 & 9/10 & 9/10
& 7/10 & 5/10 & 6/10 & 154/180 (85.6\%) & 74/80 (92.5\%) \\
DreamZero~\cite{ye2026dreamzero} $\cdot$ 1$\times$
& 10/10 & 10/10 & 10/10 & 9/10 & 9/10 & 8/10 & 7/10 & 8/10 & 9/10 & 9/10 & 10/10 & 7/10 & 0/10 & 0/10 & 1/10
& 5/10 & 3/10 & 3/10 & 118/180 (65.6\%) & 74/80 (92.5\%) \\
ACT~\cite{zhao2023act} $\cdot$ 1$\times$
& 10/10 & 10/10 & 5/10 & 10/10 & 9/10 & 9/10 & 7/10 & 7/10 & 10/10 & 10/10 & 10/10 & 8/10 & 9/10 & 10/10 & 10/10
& 5/10 & 5/10 & 5/10 & 149/180 (82.8\%) & 72/80 (90.0\%) \\
OpenPi $\pi$0.5~\cite{pi2025pi05} $\cdot$ 1$\times$
& 7/10 & 5/10 & 1/10 & 10/10 & 10/10 & 8/10 & 5/10 & 4/10 & 5/10 & 10/10 & 10/10 & 8/10 & 0/10 & 0/10 & 0/10
& 3/10 & 3/10 & 3/10 & 92/180 (51.1\%) & 61/80 (76.3\%) \\
GR00T N1.6~\cite{bjorck2025gr00t} $\cdot$ 1$\times$
& 10/10 & 10/10 & 7/10 & 7/10 & 7/10 & 6/10 & 1/10 & 3/10 & 3/10 & 9/10 & 9/10 & 7/10 & 4/10 & 4/10 & 1/10
& 0/10 & 0/10 & 0/10 & 88/180 (48.9\%) & 66/80 (82.5\%) \\
Diff.\ Policy~\cite{chi2023diffusion} $\cdot$ 1$\times$
& 3/10 & 3/10 & 2/10 & 10/10 & 8/10 & 6/10 & 3/10 & 2/10 & 4/10 & 8/10 & 9/10 & 8/10 & 0/10 & 0/10 & 0/10
& 4/10 & 0/10 & 0/10 & 70/180 (38.9\%) & 51/80 (63.8\%) \\
EgoVLA~\cite{yang2025egovla} $\cdot$ 1$\times$
& 0/10 & 1/10 & 2/10 & 7/10 & 5/10 & 8/10 & 0/10 & 4/10 & 3/10 & 10/10 & 10/10 & 7/10 & 3/10 & 5/10 & 4/10
& 0/10 & 0/10 & 0/10 & 69/180 (38.3\%) & 42/80 (52.5\%) \\
InternVLA-M1~\cite{chen2025internvlam1} $\cdot$ 1$\times$
& 0/10 & 0/10 & 0/10 & 5/10 & 5/10 & 0/10 & 0/10 & 0/10 & 0/10 & 0/10 & 0/10 & 0/10 & 3/10 & 5/10 & 7/10
& 0/10 & 0/10 & 0/10 & 25/180 (13.9\%) & 10/80 (12.5\%) \\
H-RDT~\cite{bi2026hrdt} $\cdot$ 1$\times$
& 0/10 & 0/10 & 2/10 & 0/10 & 0/10 & 1/10 & 0/10 & 1/10 & 0/10 & 0/10 & 0/10 & 0/10 & 0/10 & 0/10 & 0/10
& 0/10 & 0/10 & 0/10 & 4/180 (2.2\%) & 2/80 (2.5\%) \\
\bottomrule
\end{tabular}%
}

\vspace{3pt}
\begin{minipage}{\textwidth}
\fontsize{8}{9.6}\selectfont
\emph{Note.} For the Matched-8 total, \textbf{best} is in bold and \underline{second best} underlined across all rows. Six-task totals are provided for reference only.
\textsuperscript{a}~GPT-6-Astra PickBetweenTables results use an outcome-selected subset and a revised release metric; these episodes are included only in the reference Six-task total.
\textsuperscript{u}~For $10\times$, denominators below 10 indicate incomplete outcome coverage; unknown episodes are excluded.
The Matched-8 total uses the same eight complete task--level cells for every row (80 episodes): XMovePick L0--L2, BendPick L0--L1, and Tabletop L0--L2. Cells are selected by completeness, not success rate. This is a comparison on the shared complete subset, not an estimate for all SIMPLE tasks.
\textsuperscript{b}~GPT-6-Astra ICL Six-task total covers 5 tasks (XMoveBendPick has no ICL). Six-task totals use differing denominators across methods (baselines /180; GPT-6-Astra includes XMoveBendPick and the PickBetweenTables subset), so this column is a descriptive reference, not a matched comparison.
\end{minipage}
\end{table}


\noindent\textbf{Coverage and comparison subsets.}
At the standard ($1\times$) budget, zero-shot (ZS) evaluation has complete outcomes for five tasks (XMovePick, BendPick, Handover, Tabletop, and XMoveBendPick), totaling 150 instances. Four of these tasks also have complete ICL outcomes, forming Common-4: XMovePick, BendPick, Handover, and Tabletop, with 120 instances. Cross-budget comparisons use Matched-8, the eight task--level combinations with complete outcomes in all four settings: XMovePick L0--L2, BendPick L0--L1, and Tabletop L0--L2, totaling 80 instances across three tasks. PickBetweenTables has only partial, outcome-selected coverage and is included solely in the reference aggregates, outside both comparison subsets.

\noindent\textbf{Demonstration preparation.}
For task-level in-context learning (ICL), GPT-6-Astra explores the released training demonstrations through an information-bounded reader and compiles task notes, reference numerical data, image indices, and runnable helpers into a single hash-frozen package; the same package is shared across L0/L1/L2, every test episode starts from a fresh session, and no model weights are updated. Preparation accesses 99--100 released demonstrations per task; XMoveBendPick has no ICL package and is reported zero-shot only.

\noindent\textbf{Observation and control settings.}
At test time GPT-6-Astra receives a head-mounted left RGB image (typically $640\times360$), a 32-column robot-interface state, the task text, and the current episode's history. It is additionally given nominal robot metadata (static URDF, joint order, and limits), robot-only forward/inverse kinematics, and NumPy/SciPy code tools; object pose, contacts, and online scoring feedback are withheld. Actions are continuous physical chunks of shape $(T,36)$---finger and arm joint angles, torso roll/pitch/yaw, base height, base translation $(v_x,v_y)$, a turn flag, and a target yaw---with the chunk length $T$ chosen by the model. Each action row is executed at 50\,Hz by the task-specific official System~0 controller: the decoupled whole-body controller for the Teleop tasks, and the $\Psi_0$/AMO motion-planning path for BendPick and Tabletop~\cite{wei2026psi0,li2025amo}. The simulator is paused during inference, so reported action steps exclude thinking time.

\noindent\textbf{Budgets and scoring.}
The standard ($1\times$) per-episode action budget is 800 steps for XMovePick, Handover, Tabletop, and XMoveBendPick, 400 for BendPick, and 1{,}200 for PickBetweenTables; the extended budget is $10\times$. The Matched-8 comparison includes only complete task--level cells; partial results elsewhere use judged denominators, excluding unknown outcomes. Tabletop $1\times$ outcomes are thresholded from the $10\times$ trajectories, and one operator-stopped ICL episode (L2/E01) is counted as a failure by an explicit reporting rule. PickBetweenTables is an outcome-selected successful subset scored with a revised release metric. It is included only in the Six-task totals, which are provided for reference, and excluded from both the Matched-8 comparison in Table~\ref{tab:simple_perlevel} and the Common-4 comparison in Table~\ref{tab:simple_efficiency}. Compute, sensing, tool access, inference latency, and rendering hardware are not matched across systems, so the comparison is descriptive rather than cost-controlled.

\noindent\textbf{Results on the shared complete subset.}
Table~\ref{tab:simple_perlevel} reports per-level success and two aggregate columns. The Matched-8 total compares all four GPT-6-Astra settings and the published baselines on the same 80 instances defined above. The subset is selected by completeness, not by success rate. On this subset, zero-shot success rises from 65/80 (81.3\%) at $1\times$ to 79/80 (98.8\%) at $10\times$, while ICL rises from 77/80 (96.3\%) to 79/80 (98.8\%); both $\Psi_0$~\cite{wei2026psi0} and DreamZero achieve 74/80 (92.5\%) at $1\times$. Thus, ICL improves standard-budget performance by 15.0 percentage points on this subset, while both GPT-6-Astra settings reach the same extended-budget success rate. These results describe the shared complete subset and do not estimate performance on excluded task--level cells. The Six-task total retains all judged outcomes for reference only: task coverage and denominators differ across rows, and the ICL aggregate covers five tasks because XMoveBendPick has no ICL package.

\begin{table}[!htbp]
\fontsize{8}{9.6}\selectfont
\setlength{\tabcolsep}{4pt}\renewcommand{\arraystretch}{1.15}

\centering
\caption{GPT-6-Astra standard-budget ($1\times$) success, efficiency and cost on SIMPLE: zero-shot vs. ICL. Common-4 covers all 120 episodes from XMovePick, BendPick, Handover and Tabletop; it differs from the 80-episode Matched-8 subset in Table~\ref{tab:simple_perlevel}. Efficiency is measured on the 81 episodes solved in both settings. Six-task success is provided for reference only. Psi-0 is the published baseline.}
\label{tab:simple_efficiency}

\begin{tabular}{@{}lccc@{}}
\toprule
Metric & GPT-6-Astra Zero-shot & GPT-6-Astra ICL & Psi-0~\cite{wei2026psi0} (ref.) \\
\midrule
\rowcolor{sensepurple!15}
\multicolumn{4}{@{}l}{\textit{Success rate}} \\
Common-4 & 85/120 (70.8\%) & \textbf{109/120 (90.8\%)} & \underline{108/120 (90.0\%)} \\
Six-task (reference only)\textsuperscript{c} & 110/167 (65.9\%) & 111/126 (88.1\%) & 154/180 (85.6\%) \\
\midrule
\rowcolor{sensepurple!15}
\multicolumn{4}{@{}l}{\textit{Per-episode efficiency on matched successes}\textsuperscript{a}} \\
Mean action steps & 376.9 & 210.4 & -- \\
Mean test tokens (millions) & 2.64 & 6.40 & -- \\
Mean test wall-time (min) & 47.0 & 31.4 & -- \\
\midrule
\rowcolor{sensepurple!15}
\multicolumn{4}{@{}l}{\textit{ICL preparation (one-time, per task)}} \\
Tokens (millions) & -- & 8.34 & -- \\
Wall-time (minutes) & -- & 90.4 & -- \\
\bottomrule
\end{tabular}

\vspace{3pt}
\begin{minipage}{\textwidth}
\fontsize{8}{9.6}\selectfont
\emph{Note.} For Common-4 success at $1\times$, the \textbf{best} is in bold and the \underline{second best} underlined across methods. \textsuperscript{a}~Per-episode means over the 81 cases solved in both settings.
\textsuperscript{c}~Six-task success rates use differing denominators (GPT-6-Astra ZS /167, incl.\ XMoveBendPick and the PickBetweenTables subset; GPT-6-Astra ICL = 5 tasks, /126) and are a descriptive reference.
ICL preparation is a one-time per-task cost; XMoveBendPick (no ICL) and PickBetweenTables (no comparable cost summary) are excluded from these efficiency figures.
\end{minipage}
\end{table}

\noindent\textbf{Standard-budget ICL and efficiency.}
Table~\ref{tab:simple_efficiency} separately compares zero-shot and ICL at $1\times$ on the four common tasks (XMovePick, BendPick, Handover, and Tabletop), where all 120 episodes are judged in both settings. This is a broader subset than Matched-8 and is used only for the standard-budget comparison. GPT-6-Astra reaches 85/120 (70.8\%) zero-shot and 109/120 (90.8\%) with task-level ICL, a 20.0-percentage-point gain; the latter is comparable to the published $\Psi_0$ result of 108/120 (90.0\%), and the one-case difference does not establish statistical superiority. Spatial generalization (L2) remains the main weakness at $1\times$: ICL is 39/40 at both L0 and L1 but 31/40 at L2. On the 81 episodes solved in both settings, ICL reduces mean action steps from 376.9 to 210.4 (44.2\%), while mean test token consumption rises from 2.64M to 6.40M. Mean test wall-clock time falls from 47.0 to 31.4 minutes; preparation additionally costs 8.34M tokens and 90.4 minutes per task, paid once. These cost figures use the matched-success set and should not be interpreted as measurements on the Matched-8 subset.

\FloatBarrier
\subsection{BEHAVIOR}
\noindent\textbf{Benchmarks and tasks.} BEHAVIOR-1K~\cite{li2024behavior1k} is a household embodied-AI benchmark covering 1,000 everyday activities grounded in human needs. Built on the OmniGibson simulator, it supports realistic interactions with rigid objects, deformable objects, and liquids. Its long-horizon activities require navigation and manipulation to achieve task-specific goal conditions.

We evaluate on all 100 tasks from the BEHAVIOR-1K 2026 Challenge, using public-test episode 0 with rollout seed 0 for each task and performing a single evaluation rollout per task instance. We retain the benchmark's default R1Pro dual-arm mobile manipulator, with the arm-controller modification described below. Each rollout starts from the official task-instance initialization and proceeds through closed-loop interaction conditioned on head-mounted and bilateral wrist RGB-D observations, robot proprioception, and the task instruction. The simulator operates at 30 native steps per second, and each model-issued action can be repeated for 1–100 simulation steps. A rollout terminates upon environment termination, exhaustion of the official task-specific step budget, or an explicit agent finish call. Following the official evaluator, the step budget is set to 1.5 times the task's mean human-demonstration length.

\noindent\textbf{Protocol and Metrics.} We use GPT-6 Astra with maximum reasoning effort for both demonstration-based in-context learning and evaluation. During the ICL stage, all official training data are provided including RGB videos, actions, proprioceptive states, and annotations. We adopt a 21-dimensional action interface, consisting of three base-control values, four torso joint targets, two six-dimensional absolute end-effector pose targets, and two gripper commands. Each end-effector target specifies a position and axis-angle orientation in the robot's current base frame, which the official inverse-kinematics controller translates into arm motion. For consistency with this interface, demonstration arm commands are converted from joint-space targets to end-effector poses using the official forward-kinematics implementation.

We assess performance using the official evaluator, reporting binary task success, final predicate-based Q-score, simulator step count and simulated time, and base and end-effector travel distances together with their official human-normalized counterparts. Model token usage and wall-clock durations for individual stages are recorded separately to characterize computational cost. 

\noindent\textbf{Results.} Across the 100 evaluation rollouts on the BEHAVIOR-1K 2026 Challenge, our method successfully completes 37 tasks, achieving an overall task success rate of 37\%. 

\noindent\textbf{Other Discussion.}The aggregate task success rate increases from 1\% with the 23-dimensional joint-space interface to 37\% with the 21-dimensional end-effector-pose interface. This substantial improvement is consistent with the hypothesis that task-space action representations better align language-model planning with robotic execution. By specifying the desired position and orientation of each end effector, the model can express manipulation intent directly while delegating joint-level kinematic computation to the official controller.

Qualitative inspection of the trajectories suggests that the model can extract procedural knowledge from demonstrations, identify task stages, and adapt its plans based on visual feedback. However, failures due to imprecise grasp alignment, arm collisions, navigation obstructions, and ineffective recovery strategies highlight persistent challenges in contact-rich manipulation. Taken together, these findings suggest that action representation plays a critical role in embodied task performance, but do not establish that high-level reasoning alone is sufficient for reliable robotic manipulation. 
\FloatBarrier
\endgroup

\section{Harness Ablation Study}
\label{app:harness_ablation}

\begin{table*}[!htbp]
\fontsize{7}{8.4}\selectfont
\setlength{\tabcolsep}{2.5pt}\renewcommand{\arraystretch}{1.15}

\centering
\caption{\textbf{Harness Design Ablations.}
We evaluate the impact of harness design on GPT-6 Astra across
42 RoboDojo tasks, covering (A) agent components, (B) action representation and chunking, and (C) VLA collaboration and ICL.
VLA denotes RoboDojo-fine-tuned $\pi_{0.5}$.
$L$ denotes actions per GPT decision in (B) and VLA-proposed actions reviewed per GPT call in (C).
Performance entries report score / success rate (\%), with ``Average'' computed as the unweighted mean across the five categories.
Token consumption includes both input and output tokens and is reported in millions (M). ICL uses knowledge packages from the corresponding main tasks and a general-purpose knowledge package for Open tasks, with ICL-only results taken from Table~\ref{tab:robodojo_results}.}
\label{tab:ablation_harness}

\maxsizebox{\textwidth}{!}{%
\begin{tabular}{l*{7}{r}}
\toprule
Configuration
& \textbf{Average}
& Generalization
& Precision
& Long-Horizon
& Memory
& Open
& Tokens (M) $\downarrow$ \\
\midrule

\textbf{Baseline} 
& 22.78 / 15.56 & 27.08 / 22.22 & 9.79 / 0.00 & 23.13 / 4.17 & 22.22 / 22.22 & 31.67 / 29.17 & 2.56 \\
\midrule

\rowcolor{sensepurple!15}
\multicolumn{8}{l}{\textbf{A. Agent Components}} \\

w/ planning subagent
& 26.03 / 19.44 & 37.08 / 33.33 & 11.88 / 4.17
& 21.46 / 4.17 & 40.56 / 38.89 & 19.17 / 16.67 & 2.96 \\

w/ execution subagent
& 21.56 / 15.28 & 36.39 / 33.33 & 14.79 / 4.17
& 10.63 / 0.00 & 25.56 / 22.22 & 20.42 / 16.67 & 2.55 \\

w/ feedback subagent
& \textbf{30.00} / \textbf{24.72} & 30.83 / 27.78 & 16.88 / 12.50
& 24.38 / 8.33 & 33.33 / 33.33 & 44.58 / 41.67 & 2.74 \\

w/ history compression
& 28.03 / 21.11 & 35.42 / 33.33 & 7.29 / 0.00
& 24.38 / 4.17 & 38.89 / 38.89 & 34.17 / 29.17 & 0.86 \\
\midrule

\rowcolor{sensepurple!15}
\multicolumn{8}{l}{\textbf{B. Action Representation and Chunking}} \\

w/ directional motion commands
& 26.61 / 18.06 & 24.58 / 22.22 & 16.88 / 8.33
& 29.38 / 4.17 & 23.89 / 22.22 & 38.33 / 33.33 & 1.91 \\

w/ joint-angle
& 16.08 / 9.72 & 9.86 / 5.56 & 5.21 / 0.00
& 19.79 / 0.00 & 22.22 / 22.22 & 23.33 / 20.83 & 1.12 \\

w/ fixed action chunk ($L=4$)
& 17.97 / 11.39 & 27.50 / 25.00 & 8.96 / 0.00
& 18.13 / 0.00 & 11.11 / 11.11 & 24.17 / 20.83 & 0.43 \\

w/ adaptive action chunk
& \textbf{26.36} / \textbf{20.00} & 30.83 / 27.78 & 18.13 / 8.33
& 30.21 / 12.50 & 22.22 / 22.22 & 30.42 / 29.17 & 1.15 \\
\midrule

\rowcolor{sensepurple!15}
\multicolumn{8}{l}{\textbf{C. System 1--System 2 Collaboration}} \\

VLA ($\pi_{0.5}$)~\cite{pi2025pi05}
& 11.41 / 6.91 & 13.37 / 8.17 & 12.40 / 5.50
& 23.54 / 14.67 & 5.78 / 4.56 & 1.98 / 1.67 & -- \\

w/ VLA open-loop ($\pi_{0.5}$)~\cite{pi2025pi05}
& 11.21 / 5.83 & 21.67 / 16.67 & 1.88 / 0.00
& 32.50 / 12.50 & 0.00 / 0.00 & 0.00 / 0.00 & 0.39 \\

w/ VLA closed-loop
& 31.15 / 23.61 & 35.97 / 30.56 & 18.54 / 8.33
& 36.25 / 16.67 & 35.00 / 33.33 & 30.00 / 29.17 & 7.66 \\

w/ VLA closed-loop ($L=4$)
& 34.64 / 25.83 & 38.61 / 33.33 & 20.63 / 8.33
& 44.79 / 25.00 & 33.33 / 33.33 & 35.83 / 29.17 & 2.80 \\

w/ ICL
& 39.24 / 29.67 & 34.92 / 28.33 & 28.63 / 15.00
& 41.38 / 20.00 & 73.67 / 70.00 & 17.63 / 15.00 & 9.57 \\

w/ ICL + VLA closed-loop
& \textbf{60.08} / \textbf{54.17} & 54.58 / 50.00 & 49.38 / 45.83
& 60.42 / 41.67 & 66.67 / 66.67 & 69.38 / 66.67 & 19.33 \\

\bottomrule
\end{tabular}%
}
\vspace{3pt}
\end{table*}

All experiments are evaluated on 42 tasks of RoboDojo~\cite{chen2026robodojo} under a consistent evaluation protocol, with three evaluation runs per task using different seeds, except for ICL, which follows Table~\ref{tab:robodojo_results}. We report both task scores, which capture partial task completion, and success rates (SR), which measure full task completion. Performance is further evaluated across five dimensions: Generalization, Precision, Long-Horizon, Memory, and Open. We additionally report the average token consumption per episode, including both input and output tokens from GPT-6-Astra and auxiliary agents, in millions (M), as a proxy for inference cost. Unless otherwise specified, each ablation is compared against baseline followed by our RoboRSI~\cite{opensensenova2026roborsi} 
to isolate the impact of individual design choices.

\label{sec:harness_settings}

\subsection{Agent Components}
\label{sec:agent_components}

We evaluate four components individually: (i) a planning subagent that decomposes tasks into subgoals and completion criteria, with its outputs incorporated into the main agent's context; (ii) an execution subagent that generates EEF actions from the main agent's subgoals; (iii) a feedback subagent that evaluates execution outcomes and suggests continuation, adjustment, or retry; and (iv) history compression that retains the latest four turns and consolidates earlier interactions into a rolling summary capped at 2,000 tokens.

This study investigates how different agent components affect the embodied performance of GPT-6-Astra and whether increasing agent complexity consistently improves task execution. As shown in Table~\ref{tab:ablation_harness}, introducing a feedback subagent yields the largest improvement, increasing the average score by 7.22 points and the success rate by 9.16 percentage points, with only a 7.03\% increase in token consumption. History compression improves the score by 5.25 points and the success rate by 5.55 percentage points while reducing token consumption by 66.41\%, demonstrating benefits in both performance and efficiency. The planning subagent provides a smaller overall score improvement of 3.25 points at a 15.63\% increase in token consumption, although its substantial gain in the Memory category indicates that its benefits vary across task categories. In contrast, the execution subagent reduces the score by 1.22 points at approximately the same token cost as the baseline. These results suggest that effective agent design depends on how additional reasoning supports task execution, rather than on the number of components alone.

Feedback and history compression offer complementary benefits: explicit outcome evaluation improves task performance with a modest increase in token consumption, while compressing earlier interactions improves performance and substantially reduces context overhead. These findings motivate further investigation of structured execution memory and targeted verification mechanisms for tracking task progress and failures. Future harness design should assess additional components by their contributions to both task performance and total token consumption.

\FloatBarrier
\section{Full Real Robot Results}
\label{app:real_world}


\newenvironment{rwcard}{\par\smallskip\noindent\fontsize{8}{9.6}\selectfont\setlength{\tabcolsep}{4pt}\renewcommand{\arraystretch}{1.20}%
  \tabularx{\linewidth}{@{}>{\columncolor{sensepurple!6}\bfseries\raggedright\arraybackslash}p{3.5cm} >{\raggedright\arraybackslash}X@{}}}%
  {\endtabularx\par\smallskip}
\newenvironment{rwtasks}{\par\vspace{3pt}\noindent\fontsize{8}{9.6}\selectfont\setlength{\tabcolsep}{4pt}\renewcommand{\arraystretch}{1.20}%
  \tabularx{\linewidth}{@{}>{\raggedright\arraybackslash}X >{\centering\arraybackslash}p{1.2cm} >{\centering\arraybackslash}p{1.5cm} >{\centering\arraybackslash}p{1.9cm}@{}}%
  \toprule \textbf{Task (instruction given to GPT-6-Astra)} & \textbf{Trials} & \textbf{Success} & \textbf{Wall-clock time} \\ \midrule}%
  {\bottomrule\endtabularx\par\smallskip}
\newcommand{\rwhead}[1]{\par\smallskip\noindent{\color{sensepurple}\textbf{#1}}\par\nobreak}

This appendix documents the real-world experiments of Sec.~\ref{sec:real_world}. App.~\ref{app:rw_platforms} summarizes all platforms, and App.~\ref{app:rw_franka3}--\ref{app:rw_g1} document each platform, from the simplest to the most complex embodiment, in the same five parts: setting, context provided, tasks and results, findings, and visualization.

\subsection{Platforms and Results}
\label{app:rw_platforms}

Table~\ref{tab:rw_platforms} summarizes the cameras, action interfaces, supplied context, and task outcomes across all eight platforms. Figure~\ref{fig:rw_gallery_full} provides task visualizations grouped by outcome, while the following subsections detail each platform's protocols and results.

\clearpage
\par\medskip\noindent\begin{minipage}{\linewidth}
\fontsize{8}{9.6}\selectfont
\setlength{\tabcolsep}{4pt}\renewcommand{\arraystretch}{1.2}

\centering
\captionof{table}{\textbf{Real-world platforms, the context supplied to GPT-6-Astra, and results} (successes / trials), ordered from simple to complex embodiments. No demonstrations are used on any platform. Cameras list only the streams given to GPT-6-Astra.}
\label{tab:rw_platforms}

\begin{tabularx}{\textwidth}{@{}>{\raggedright\arraybackslash}p{2.2cm} >{\raggedright\arraybackslash}p{2.1cm} >{\raggedright\arraybackslash}p{2.6cm} >{\raggedright\arraybackslash}X >{\raggedright\arraybackslash}p{3.7cm}@{}}
\toprule
\textbf{Platform} & \textbf{Cameras} & \textbf{Action interface} & \textbf{Context beyond the task instruction} & \textbf{Tasks and results} \\
\midrule
\rowcolor{sensepurple!15}\multicolumn{5}{@{}l}{\textit{Single arm with gripper}} \\
Franka 3, Robotiq 2F-85 & Wrist ZED Mini + third-person ZED 2 & End-effector pose + grasp label & Robot settings & Block stacking 6/6 \newline Cloth folding 2/4 \newline Plug insertion 0/3 \\
Franka Panda, Franka Hand & Calibrated wrist webcam & Absolute end-effector pose + gripper open/close & Tool schemas, joint limits, safety constraints & Tape-roll stacking 3/3 \newline Towel folding 2/2; cloth folding 1/2 \newline Plug insertion 0/5 \\
AgileX PiPER, Pika gripper & Wrist + side view & End-effector pose and gripper commands; IK by local tools & Step-level task prompt, admissible grasp points & Container nesting 3/3 \newline Cloth folding 3/3 \newline Plug insertion 0/3 \\
\rowcolor{sensepurple!15}\multicolumn{5}{@{}l}{\textit{Dual arm with grippers}} \\
AgileX Piper, two arms & Front + two wrist views & Controls both arms directly without predefined protocols & Robot specs and official manuals & 12 tasks, 14/14:\newline { batteries to plates, batteries to box, blocks to basket, items to grids, cup stacking, basket lift, bimanual milk-carton hold, pens to holder, towel-roll pyramid, towel folding, T-shirt folding, flower insertion} \\
\rowcolor{sensepurple!15}\multicolumn{5}{@{}l}{\textit{Dual arm with dexterous hands}} \\
Tianji M6S CCS-809, Wuji Gen-1 hands & Head Azure Kinect & End-effector pose + grasp label & Tool schemas, joint limits, hand presets; human-written grasp scripts & Toy grasp 3/3 \newline Bottle grasp 1/3$^{\dagger}$ \newline Model-written grasp code 0/3 \\
\rowcolor{sensepurple!15}\multicolumn{5}{@{}l}{\textit{Wheeled mobile manipulator}} \\
Galaxea R1 Pro & Head stereo, five chassis, two wrist D405 & GPT-6-Astra controls exposed robot parts directly & One-time exploration on the robot with GPT-6-Astra & Find washer and press button 2/2$^{\ddagger}$ \newline Flowers to basket 0/2$^{\S}$ \newline 6 tabletop manipulation tasks, 9/9:\newline { basket lift, towel rolls to tray, sponges to tank, clothes to basket, spill wiping and tidying, bottles and towels to tray} \\
AgileX Tracer + xArm7 & Arm-mounted RealSense D455 & Three navigation waypoints & None (no robot model, SLAM or collision detection) & Navigation and search 9/23 (five tasks) \\
\rowcolor{sensepurple!15}\multicolumn{5}{@{}l}{\textit{Humanoid}} \\
Unitree G1 (29 DoF), two Wuji Gen-1 hands & Head Azure Kinect & 14 arm-joint targets, binary hand open/close; IMU turn, timed walk at 0.15\,m/s, stop & Tool schemas, joint limits, binary hand presets, locomotion and safety constraints & Toy pick-and-drop 2/3 \newline Turn, approach and stop 3/3 \\
\bottomrule
\end{tabularx}

\vspace{3pt}
{\fontsize{8}{9.6}\selectfont\raggedright $^{\dagger}$Succeeded after human feedback.\quad $^{\ddagger}$Judged on the button press.\quad $^{\S}$Both runs ended by a torso over-temperature fault.
\par}
\end{minipage}\par\medskip

\par\medskip\noindent\begin{minipage}{\linewidth}
\centering
\captionof{figure}{\textbf{Full gallery of zero-shot real-world tasks, grouped by outcome.} Three frames per task in temporal order; chips give successes over all counted trials. (A) Tasks in which every counted trial succeeded; for the dual-arm Piper and the Galaxea R1 Pro tabletop tasks, counts are those of the reported rollouts (App.~\ref{app:rw_piper} and~\ref{app:rw_galaxea}). A1 and A2, and A12 and A13, show two rollouts of the same task. (B) Tasks with moderate completion, ordered by success rate: some trials fail, or, as in B1, every counted trial succeeds but the result is imperfect (the folded cloth remains untidy). (C) Tasks in which no trial succeeded.}
\label{fig:rw_gallery_full}
\vspace{4pt}
\includegraphics[width=\linewidth,height=0.80\textheight,keepaspectratio]{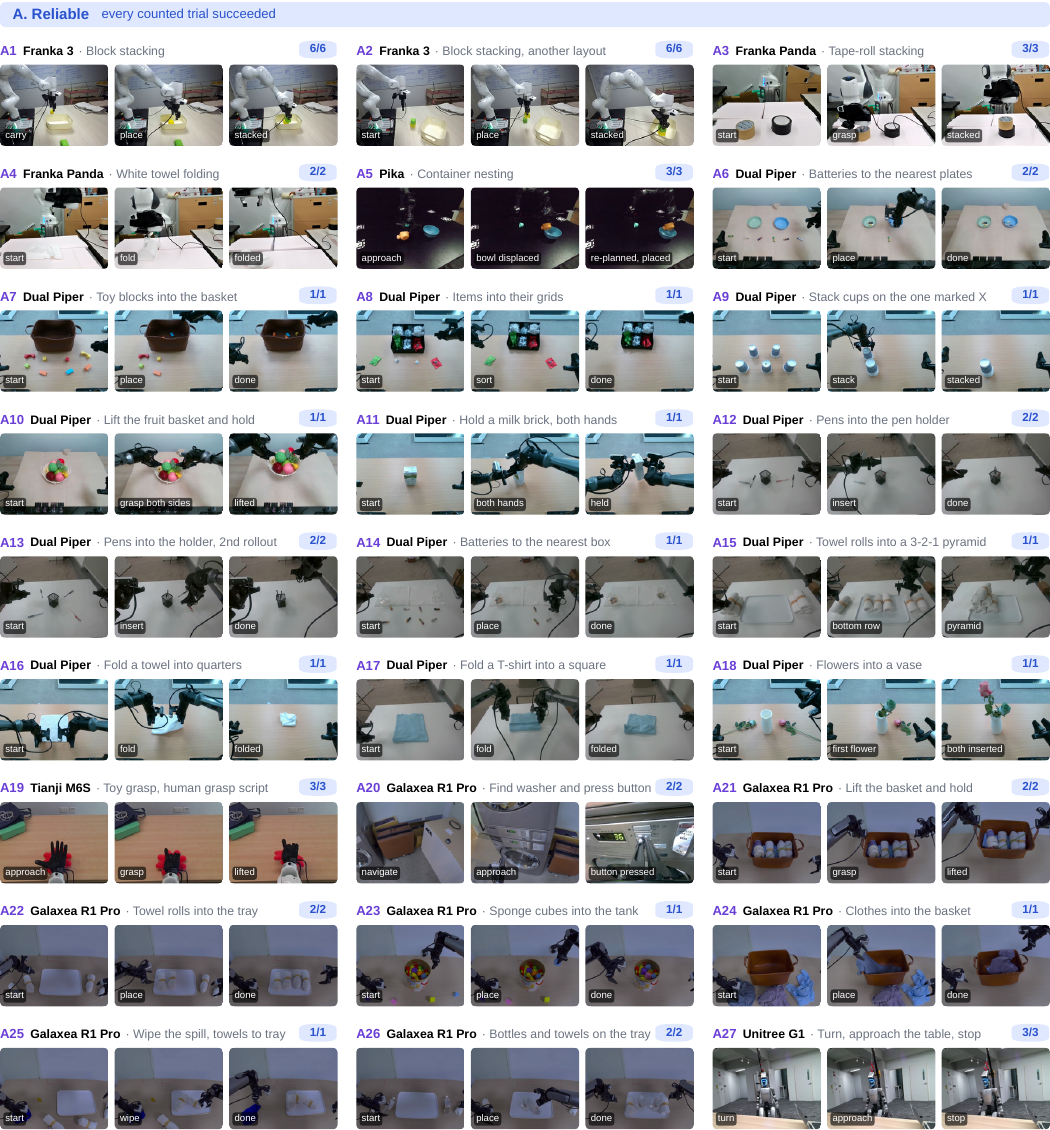}
\end{minipage}\par\medskip
\noindent\begin{minipage}{\linewidth}
\centering
\includegraphics[width=\linewidth,height=0.80\textheight,keepaspectratio]{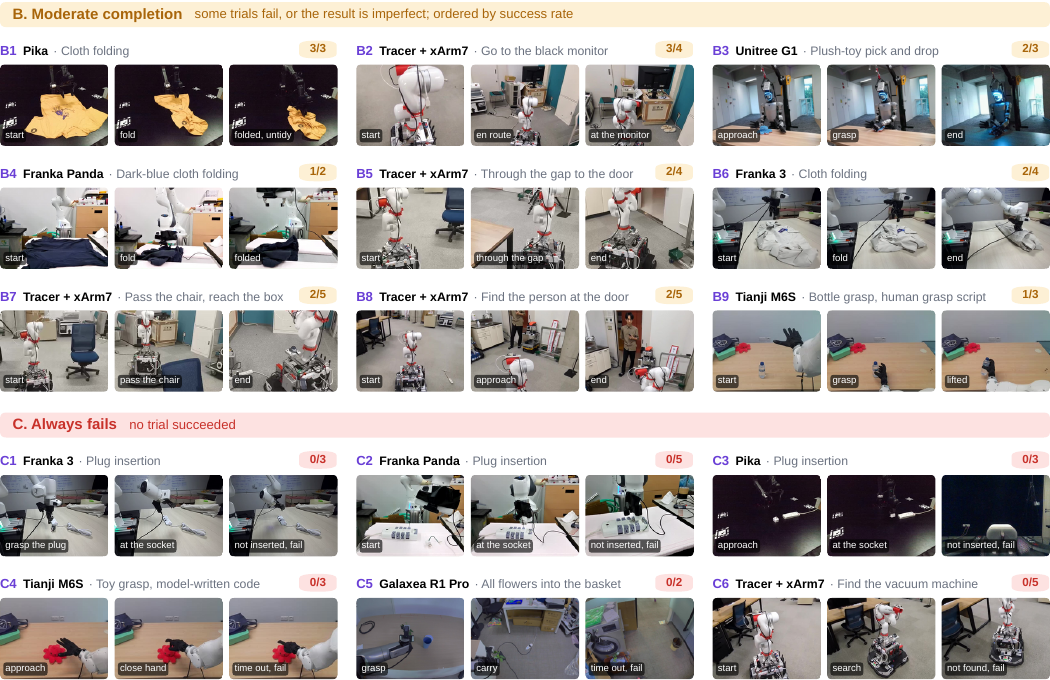}
\end{minipage}\par\medskip

\subsection{Single Arm with Gripper --- Franka 3: Block Stacking, Cloth Folding and Plug Insertion}
\label{app:rw_franka3}

\rwhead{Setting}
\begin{rwcard}
Robot & Franka Research 3 (7-DoF) with a Robotiq 2F-85 parallel gripper \\
Cameras given to GPT-6-Astra & Third-person Stereolabs ZED 2 and wrist ZED Mini; RGB only \\
Action interface & End-effector pose and a grasp label, issued as short trajectory segments at 15\,Hz with gripper open/close commands \\
Harness & DROID-style setup on ROS~2: each segment is executed by the joint-trajectory controller, and the measured joint, end-effector and gripper states are returned; exterior and wrist videos are recorded at 15\,fps; Model: GPT-6-Astra, Reasoning Effort: max \\
Budget and success rule & One hour per trial; an operator marks each run as success or failure when it ends. The recorded runs last 7--16\,min \\
\end{rwcard}

\rwhead{Context provided}
\begin{itemize}[leftmargin=1.4em,itemsep=0pt,topsep=1pt]
\item Task instruction, robot settings, two RGB views and robot-state feedback.
\item No demonstrations and no semantic manipulation primitives.
\end{itemize}

\rwhead{Tasks and results}
\begin{rwtasks}
Place the yellow block into the box, then put the green block into the box as well, stacking it on top of the yellow block & 6 & 6 & 7--11\,min$^{*}$ \\
First, lay the collar area flat; then, fold the left sleeve inward, followed by the right sleeve. Finally, fold the bottom hem of the garment up toward the collar & 4 & 2 & 10--15\,min$^{*}$ \\
Grasp the plug and insert it into the socket, aligning it with the slots & 3 & 0 & 9--16\,min$^{*}$ \\
\end{rwtasks}
{\fontsize{8}{9.6}\selectfont $^{*}$Trial counts and times are those of the recorded runs. Two folding trials ended in execution errors and are counted as failures; the three plug-insertion trials were marked failed by the operator.\par}

\rwhead{Findings}
\begin{itemize}[leftmargin=1.4em,itemsep=1pt,topsep=1pt]
\item \textbf{Rigid pick-and-place transfers from simulation.} Block stacking succeeds in all 6 trials across different initial configurations, matching simulation.
\item \textbf{Folding is judged without a shared criterion.} The operator marks a folding run as a success from the final state of the cloth; how flat or well aligned the fold must be is not specified, and the judgment was not unified with the other platforms (App.~\ref{app:rw_pika}), so folding results are not directly comparable across them.
\item \textbf{Contact-rich tasks end in protective stops.} With thin cloth the gripper easily touches the table and the robot halts; a first failed insertion typically collides with the socket and halts the robot. A simulator would let execution continue and leave room for further attempts.
\end{itemize}

\par\noindent\begin{minipage}{\linewidth}
\rwhead{Visualization}
\begin{center}
\includegraphics[width=\linewidth]{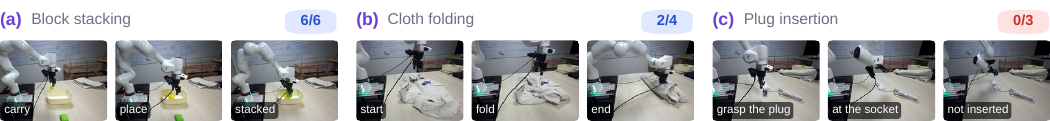}
\captionof{figure}{Franka 3: (a) block stacking, (b) cloth folding and (c) plug insertion, exterior view; in (c) the plug reaches the socket but is not inserted.}
\end{center}
\end{minipage}

\subsection{Single Arm with Gripper --- Franka Panda: Tape-Roll Stacking, Towel Folding and Plug Insertion}
\label{app:rw_panda}

\rwhead{Setting}
\begin{rwcard}
Robot & Franka Panda with the Franka Hand parallel gripper \\
Cameras given to GPT-6-Astra & One calibrated wrist webcam (intrinsics and hand--eye), 640$\times$480 RGB at 25\,fps \\
Action interface & Tool calls: camera image, robot and gripper state, end-effector pose, absolute end-effector moves, gripper open and force-controlled close \\
Harness & ROS~2 backend; collision-scene setup and gripper timeouts are handled by the backend or operator \\
Budget and success rule & No common time limit: runs last 7--37\,min, and the successful dark-blue-cloth run took 196\,min. A person scores each trial afterwards: it is a success only when the instructed physical goal and its spatial constraints are achieved and the model reports completion; all other trials count as failures. Each physical motion request is approved by the operator before it is executed \\
\end{rwcard}

\rwhead{Context provided}
\begin{itemize}[leftmargin=1.4em,itemsep=0pt,topsep=1pt]
\item Task instruction, tool schemas, joint limits and safety constraints.
\item No demonstrations; no pick, place, stack, fold or insert tools, no custom IK, joint trajectories or shell access.
\end{itemize}

\rwhead{Tasks and results}
\begin{rwtasks}
Pick one tape roll and place it on another & 3 & 3 & 10\,min$^{*}$ \\
Fold the white towel & 2 & 2 & 20\,min$^{*}$ \\
Fold the dark-blue cloth & 2 & 1 & 9\,min; 196\,min \\
Insert the white plug into the socket & 5 & 0 & 22, 37, 7, 19, 27\,min \\
\end{rwtasks}
{\fontsize{8}{9.6}\selectfont $^{*}$Only one trial of this task has a model log; the time is that trial's.\par}

\rwhead{Findings}
\begin{itemize}[leftmargin=1.4em,itemsep=1pt,topsep=1pt]
\item \textbf{Sub-tasks are solved in isolation.} When inserting the plug, GPT-6-Astra grasps it without checking whether the resulting pose admits insertion.
\item \textbf{Cautious termination under uncertainty.} In trial 1 (117 tool calls, 10.6\,M tokens), GPT-6-Astra stops pushing when small moves produce diminishing motion, withdraws 2.2\,mm and reports the insertion as unverified.
\item \textbf{No recovery from hardware faults.} When the driver reports a collision, GPT-6-Astra declares the task failed.
\end{itemize}

\par\noindent\begin{minipage}{\linewidth}
\rwhead{Visualization}
\begin{center}
\includegraphics[width=\linewidth]{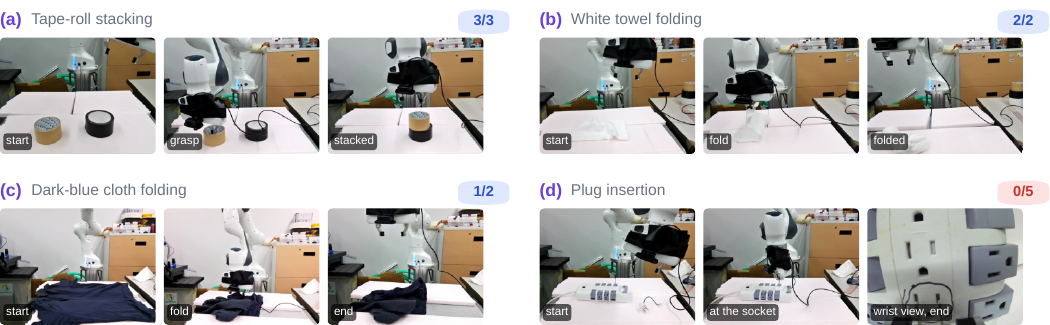}
\captionof{figure}{Franka Panda: (a) tape-roll stacking, (b) white-towel folding, (c) dark-blue-cloth folding (the successful run) and (d) plug insertion, ending with the final wrist view. Frames (a)--(c) come from time-compressed recordings and are labelled by phase.}
\end{center}
\end{minipage}

\subsection{Single Arm with Gripper --- PiPER with Pika Gripper: Container Nesting, Cloth Folding and Plug Insertion}
\label{app:rw_pika}
\rwhead{Setting}
\begin{rwcard}
Robot & One AgileX PiPER arm (6 joints) with a Pika parallel gripper \\
Cameras given to GPT-6-Astra & Wrist RealSense D405 (640$\times$480 RGB) and a third-person Azure Kinect DK from the front right (1280$\times$720 RGB). Depth is not given as an image, but a tool returns the depth and 3D point of a queried wrist-image pixel \\
Action interface & GPT-6-Astra outputs target end-effector poses, motion paths and gripper commands; local tools solve IK, check and execute the trajectory, and return the measured state with new images of both views \\
Harness & \texttt{gpt-6-astra}, with reasoning effort \texttt{high} for container nesting and plug insertion, and \texttt{max} for folding; every trial starts from a fresh context, without the trajectories or interaction records of other trials \\
Budget and success rule & 60 minutes from the first model request (plus 60\,s grace); a person judges success from the final state: containers must be stably nested in the intended order, the plug must be correctly seated in the socket, and folding is assessed qualitatively from the overall folded result \\
\end{rwcard}

\rwhead{Context provided}
\begin{itemize}[leftmargin=1.4em,itemsep=0pt,topsep=1pt]
\item \textbf{Task prompt:} The task goal, step guidance and manipulation advice (e.g.\ cup wall or rim grasps).
\item \textbf{System prompt:} Tool and coordinate conventions, joint-coordination principles, and instructions to verify action outcomes and replan when needed.
\item \textbf{Observations and feedback:} Current images, measured robot state, calibration, joint limits and within-trial tool feedback. The model chooses grasp points, poses and routes; no demonstrations or reference trajectories are provided.
\end{itemize}

\rwhead{Tasks and results}
\begin{rwtasks}
Nest the containers into the bowl & 3 & 3 & 15--24\,min \\
Fold the clothes & 3 & 3 & 32--60\,min \\
Insert the plug into the socket & 3 & 0 & 7--14\,min$^{*}$ \\
\end{rwtasks}
{\fontsize{8}{9.6}\selectfont Two earlier folding runs with an incomplete task prompt are not counted. $^{*}$Two of the three insertion runs ended in a J6 collision fault of the joint driver.\par}

\rwhead{Findings}
\begin{itemize}[leftmargin=1.4em,itemsep=1pt,topsep=1pt]
\item \textbf{Re-planning from feedback.} When the receiving bowl is displaced by contact, GPT-6-Astra detects the change, raises the container and realigns before placing it (run 3, 07:06--10:40); after an empty grasp it re-grasps the cloth (folding run 2, 12:46--17:21).
\item \textbf{The prompt shapes the strategy.} Unless alternative grasp points are named, GPT-6-Astra keeps trying to grasp a cup by its handle even when the pose is awkward or unreachable.
\item \textbf{Insertion is limited by observation and reachability.} The pins and the socket are poorly visible, and poses that would reveal them are limited by joint reachability. Reorienting the initially upward-facing plug for insertion can also bring the arm close to its joint limits.
\item \textbf{Folding is judged without a shared criterion.} An operator judges success from the final cloth state: a smaller, layered cloth packet remaining on the table after release, without requiring perfectly flat or aligned folds. All three reported trials met this practical criterion, although wrinkles and uneven edges remain (Fig.~\ref{fig:rw_gallery_full}, B1). One run reached the 60-minute limit and was subsequently accepted based on its final state. Since the success criterion was not standardized across platforms (App.~\ref{app:rw_franka3}), the reported folding success rates are not directly comparable.
\end{itemize}

\par\noindent\begin{minipage}{\linewidth}
\rwhead{Visualization}
\begin{center}
\includegraphics[width=\linewidth]{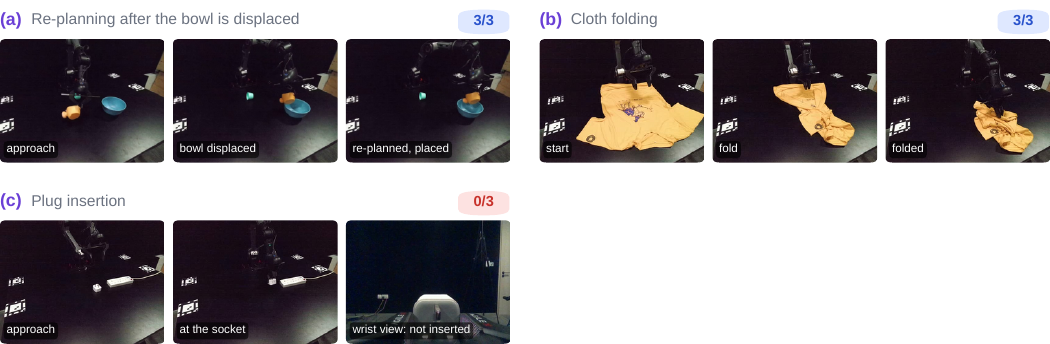}
\captionof{figure}{PiPER with Pika gripper: (a) re-planning after the receiving bowl is displaced, (b) cloth folding and (c) plug insertion, whose last frame is the wrist view with the plug covering the socket.}
\end{center}
\end{minipage}

\subsection{Dual Arm with Grippers --- AgileX Piper Series: Zero-Shot Bimanual Tasks}
\label{app:rw_piper}

\rwhead{Setting}
\begin{rwcard}
Robot & Two AgileX Piper/PiperX arms with parallel grippers \\
Cameras given to GPT-6-Astra & One front view and one wrist view per arm \\
Action interface & GPT-6-Astra controls both arms directly, without predefined control protocols \\
Harness & GPT-6-Astra is called through the Codex SDK within the Agentic Orchestration Policy framework\footnote{\url{https://github.com/ZexinHe/Agentic-Orchestration-Policy}} \\
Budget and success rule & No predefined time limit per rollout. GPT-6-Astra decides when a task is complete; a person then judges success from the physical outcome \\
\end{rwcard}

\rwhead{Context provided}
\begin{itemize}[leftmargin=1.4em,itemsep=0pt,topsep=1pt]
\item The task instruction, verbatim as in the table below.
\item A one-time exploration of each robot before the rollouts, which exposes detailed robot specifications such as the URDF model, camera specifications and links to the official documentation.
\item No demonstrations; every task is run zero-shot.
\end{itemize}

\rwhead{Tasks and results}
\noindent These are demonstration rollouts that illustrate what GPT-6-Astra can do; each reported rollout succeeded. Times are elapsed wall-clock times.
\begin{rwtasks}
Put the batteries into their nearest plates. & 2 & 2 & 16--20\,min \\
Put the toy blocks on the table into the basket. & 1 & 1 & 18\,min \\
Put the items in respective grids. & 1 & 1 & 11\,min \\
Stack the paper cups to the one marked with X. & 1 & 1 & 18\,min \\
Lift the fruit basket in the air, with a clear margin above the table, and hold still. & 1 & 1 & 6\,min \\
Hold the milk brick in the air with both hands, showing a clear margin above the table. & 1 & 1 & 20\,min \\
Put the pens into the pen holder. & 2 & 2 & 11--14\,min \\
Put the batteries into their nearest box. & 1 & 1 & 21\,min \\
Stack the towel rolls into a pyramid on the tray, three at the bottom, two in the middle, and one on top. & 1 & 1 & 15\,min \\
Fold the towel into a quarter square, from bottom to top, then from left to right. & 1 & 1 & 14\,min \\
Fold the T-shirt into a square, from bottom to top. & 1 & 1 & 8\,min \\
Insert the plastic flowers into the vase. & 1 & 1 & 21\,min \\
\end{rwtasks}

\rwhead{Findings}
\begin{itemize}[leftmargin=1.4em,itemsep=1pt,topsep=1pt]
\item \textbf{Instructions with relations and order.} GPT-6-Astra follows spatial relations (``nearest plate'', ``the cup marked X'') and prescribed orders (fold bottom-to-top, then left-to-right; a 3-2-1 pyramid).
\item \textbf{Bimanual coordination.} Both arms are used together to hold a carton or lift a basket clear of the table, showing that GPT-6-Astra can coordinate two arms without a bimanual controller.
\item \textbf{Time varies across rollouts.} For tasks with more than one rollout, the elapsed time varies, partly because of network latency and Codex retries in some runs.
\end{itemize}

\par\noindent\begin{minipage}{\linewidth}
\rwhead{Visualization}
\begin{center}
\includegraphics[width=\linewidth]{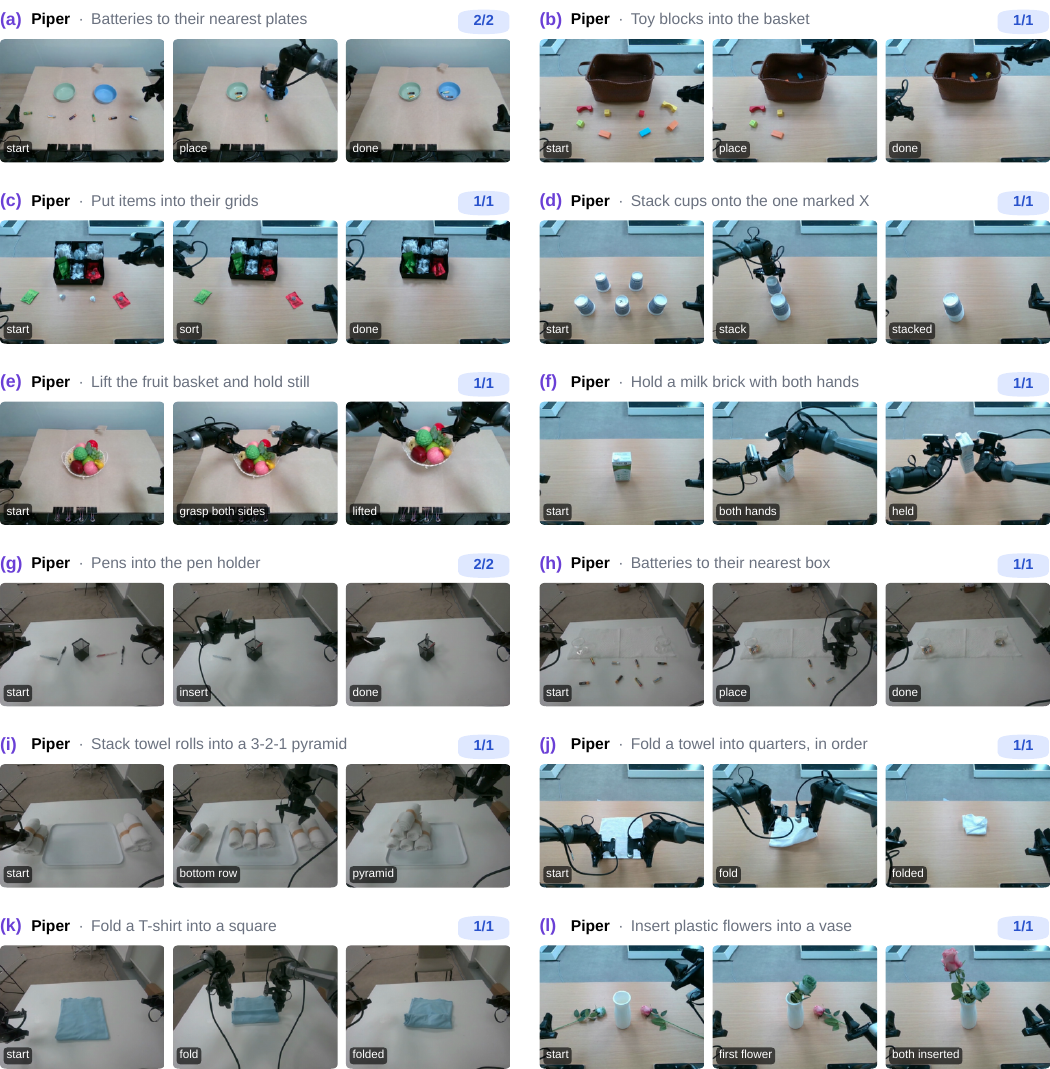}
\captionof{figure}{AgileX Piper/PiperX: all tabletop manipulation tasks, from simple placement to ordered folding and insertion.}
\end{center}
\end{minipage}

\subsection{Dual Arm with Dexterous Hands --- Tianji M6S: Toy and Bottle Grasping}
\label{app:rw_tianji}

\rwhead{Setting}
\begin{rwcard}
Robot & Tianji M6S CCS-809 with two Wuji Gen-1 dexterous hands \\
Cameras given to GPT-6-Astra & Head-mounted Azure Kinect; RGB only \\
Action interface & End-effector pose and a grasp label \\
Harness & Tool calls through the OpenAI Responses API; every model call, tool call and camera frame is archived. Collision-scene setup and gripper timeouts are handled by operator \\
Budget and success rule & 60 minutes per attempt; a person approves the final state \\
\end{rwcard}

\rwhead{Context provided}
\begin{itemize}[leftmargin=1.4em,itemsep=0pt,topsep=1pt]
\item Task instruction, tool schemas, joint limits, predefined hand open/close profiles, RGB observations and robot-state feedback.
\item In the human-coding condition, grasp scripts written by a person; no demonstrations in either condition.
\end{itemize}

\rwhead{Tasks and results}
\begin{rwtasks}
Grasp the toy, grasp code written by GPT-6-Astra & 3 & 0 & 60\,min \\
Grasp the toy, human-written grasp script & 3 & 3 & 9, 22, 26\,min \\
Grasp the bottle, human-written grasp script & 3 & 1$^{\dagger}$ & 29, 56, 38\,min \\
\end{rwtasks}
{\fontsize{8}{9.6}\selectfont $^{\dagger}$After a timed-out attempt, success followed human feedback in the final 10 minutes. A fourth attempt in the model-written condition, with a human hint, also failed.\par}

\rwhead{Findings}
\begin{itemize}[leftmargin=1.4em,itemsep=1pt,topsep=1pt]
\item \textbf{Direct dexterous control is hard.} Code written entirely by GPT-6-Astra failed, including after it summarized its failures and after a human hint.
\item \textbf{Human priors make the task feasible.} Following human-written grasp scripts, the toy grasp succeeds in every trial.
\item \textbf{Real-world perturbations matter.} Small disturbances and object deformation affect the grasp; together with overheating and safety, these are the main differences from simulation.
\end{itemize}

\par\noindent\begin{minipage}{\linewidth}
\rwhead{Visualization}
\begin{center}
\includegraphics[width=\linewidth]{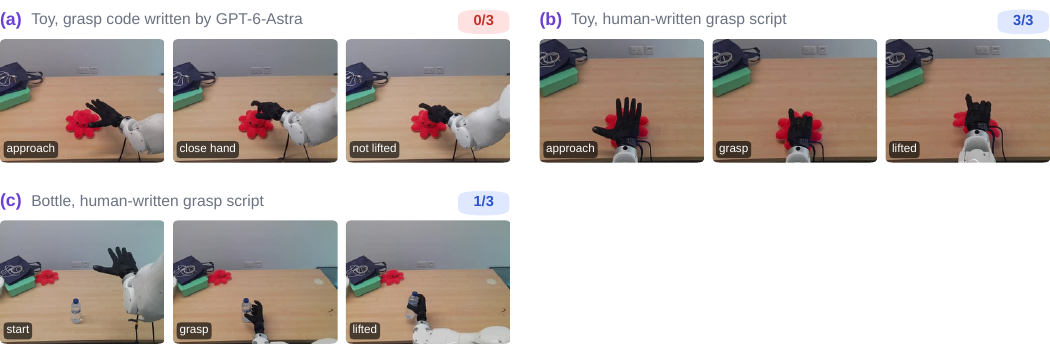}
\captionof{figure}{Tianji M6S with Wuji hands: (a) toy grasp with grasp code written by GPT-6-Astra (round 1; the toy is never lifted), (b) toy grasp and (c) bottle grasp with human-written grasp scripts. Frames from the head-camera recordings.}
\end{center}
\end{minipage}

\subsection{Wheeled Mobile Manipulator --- Galaxea R1 Pro: Appliance Use, Flower Collection and Tabletop Tasks}
\label{app:rw_galaxea}

\rwhead{Setting}
\begin{rwcard}
Robot & Galaxea R1 Pro (2025): wheeled base, 4-axis torso, two 7-DoF arms with parallel grippers \\
Cameras given to GPT-6-Astra & Head stereo pair, five chassis cameras and two wrist RealSense D405 (RGBD); LiDAR \\
Action interface & GPT-6-Astra controls exposed robot parts directly, without predefined control protocols \\
Harness & Tabletop manipulation tasks use the Agentic Orchestration Policy framework (App.~\ref{app:rw_piper}); navigation tasks run in the Codex app directly \\
Budget and success rule & One hour per task; a task is failed unless the physical outcome is verified in fresh images before the deadline \\
\end{rwcard}

\rwhead{Context provided}
\begin{itemize}[leftmargin=1.4em,itemsep=0pt,topsep=1pt]
\item For tabletop manipulation, only the two arms and grippers are exposed for control, and only the head and wrist cameras for observation.
\item For navigation, the torso and chassis are additionally exposed for control, and the five chassis cameras and the LiDAR for observation.
\item A one-time exploration of the robot before the rollouts, which exposes its specifications.
\item No demonstrations; every task is run zero-shot.
\end{itemize}

\rwhead{Tasks and results}
\begin{rwtasks}
Find washer and press button & 2 & 2$^{\ddagger}$ & 41--52\,min \\
Put all flowers into the basket near the washing machine & 2 & 0$^{\S}$ & 60\,min \\
Lift the basket in the air, and hold still. & 2 & 2 & 6--9\,min \\
Put the towel rolls into the tray, side by side. & 2 & 2 & 13--18\,min \\
Put the sponge cubes into the tank. & 1 & 1 & 13\,min \\
Put the clothes into the basket. & 1 & 1 & 25\,min \\
Clean up the dirty water with the blue towel. Put the clean white towel rolls into the tray. & 1 & 1 & 15\,min \\
Put the two bottles of water into the tray, upright in the center, then put the rolled towels onto the tray, one on each side. & 2 & 2 & 32--43\,min \\
\end{rwtasks}
{\fontsize{8}{9.6}\selectfont $^{\ddagger}$Judged on the Warm-button press. The wash cycle did not start in either run, which we attribute to the washer: it requires the Start button to be pressed shortly after Warm, and it showed an unexpected fault code in one run. \quad $^{\S}$0 of 4 and 1 of 4 flowers placed; both runs ended by a torso over-temperature fault. \quad The last six tasks are tabletop manipulation runs.\par}

\rwhead{Findings}
\begin{itemize}[leftmargin=1.4em,itemsep=1pt,topsep=1pt]
\item \textbf{Self-built context.} GPT-6-Astra discovers the robot's interfaces itself and records discrepancies with the documentation, such as a chassis message type that differs from the vendor guide.
\item \textbf{Contact without force sensing.} GPT-6-Astra uses joint-effort change as a contact signal and approaches in small steps, and when reflective button surfaces corrupt the wrist depth, it cross-checks several depth patches against RGB.
\item \textbf{Long runs meet hardware limits.} As a task proceeds, some runs hit hardware over-temperature faults that significantly affect completion; we regard this as a limitation of hardware reliability rather than of GPT-6-Astra.
\end{itemize}

\par\noindent\begin{minipage}{\linewidth}
\rwhead{Visualization}
\begin{center}
\includegraphics[width=\linewidth]{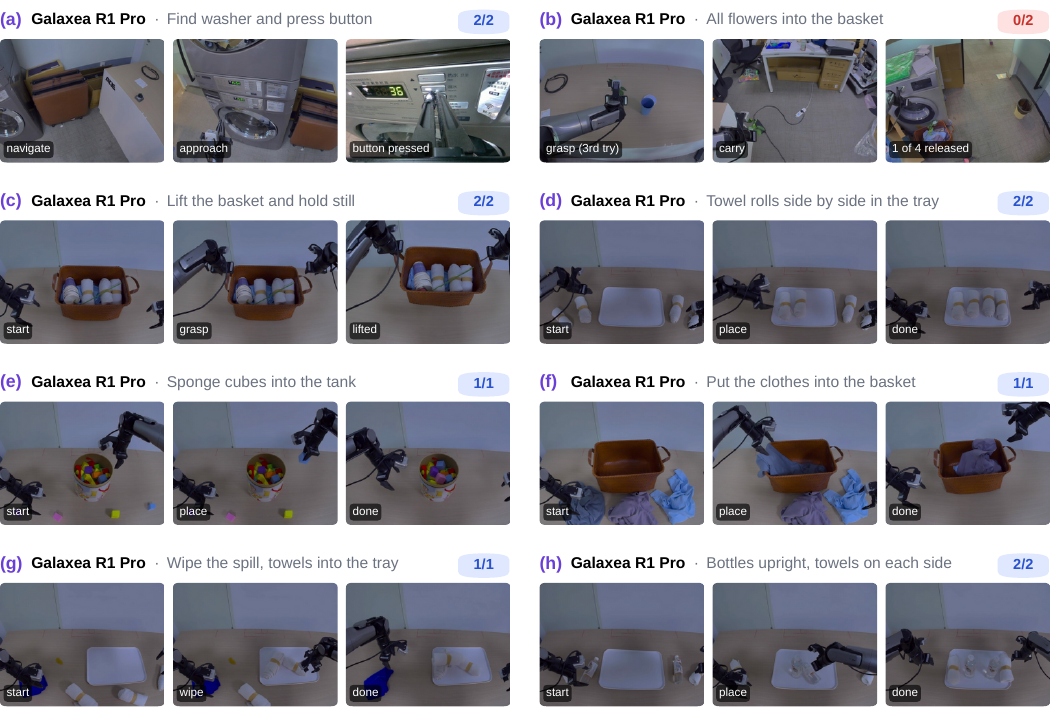}
\captionof{figure}{Galaxea R1 Pro: (a) washer task and (b) flower collection, from the task records; (c--h) tabletop manipulation tasks.}
\end{center}
\end{minipage}

\subsection{Wheeled Mobile Manipulator --- AgileX Tracer + xArm7: Navigation and Search}
\label{app:rw_agilex}

\rwhead{Setting}
\begin{rwcard}
Robot & AgileX Tracer 1.0 mobile base with a UFactory xArm7 \\
Cameras given to GPT-6-Astra & Forward-facing RealSense D455 on the arm, uncalibrated, 1280$\times$720 RGB \\
Action interface & Three navigation waypoints per decision, followed by a waypoint controller (cruise 0.7\,m/s); the pose is estimated by integrating the commanded velocity, with no odometry, SLAM or collision detection \\
Harness & \texttt{gpt-6-astra} through the OpenAI API; each decision receives the latest RGB frame (downscaled to 640\,px width) and returns the next waypoints \\
Budget and success rule & No fixed time limit; runs last 0.2--8.3\,min and end when the model reports completion or being blocked, on a timeout, or by operator interruption. A person scores each trial afterwards: it is a success only when the instructed physical goal and its spatial constraints are achieved and the model reports completion; all other trials count as failures \\
\end{rwcard}

\rwhead{Context provided}
\begin{itemize}[leftmargin=1.4em,itemsep=0pt,topsep=1pt]
\item Task instruction and RGB input only: no robot model, SLAM or collision detection, and no demonstrations.
\end{itemize}

\rwhead{Tasks and results}
\begin{rwtasks}
Pass the blue chair on the left, then approach the wooden box & 5 & 2 & 0.5--3.6\,min \\
Pass through the tripod--table gap, then approach the door & 4 & 2 & 2.5--8.3\,min \\
Approach the black monitor on the left & 4 & 3 & 0.7--2.8\,min \\
Find the person at the door on the right & 5 & 2 & 1.2--3.0\,min \\
Find the vacuum machine & 5 & 0 & 0.2--1.1\,min \\
\end{rwtasks}

\rwhead{Findings}
\begin{itemize}[leftmargin=1.4em,itemsep=1pt,topsep=1pt]
\item \textbf{Losing track of progress.} The robot loses its heading and, in some trials, forgets that it has already reached the goal area.
\item \textbf{Over-checking constraints.} When told to pass a chair on the left, GPT-6-Astra repeatedly confirms the side instead of progressing toward the goal, and loses its way in some trials.
\item \textbf{No recovery after collisions.} After a collision GPT-6-Astra almost never plans a recovery trajectory.
\end{itemize}

\par\noindent\begin{minipage}{\linewidth}
\rwhead{Visualization}
\begin{center}
\includegraphics[width=\linewidth]{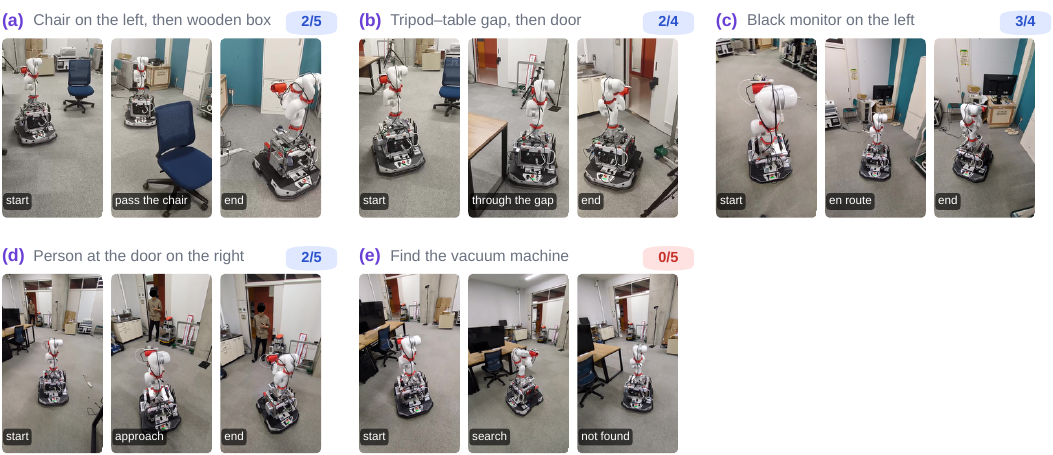}
\captionof{figure}{AgileX Tracer + xArm7: one recorded run per navigation task, third-person phone video. Chips give successes over all trials of the task.}
\end{center}
\end{minipage}

\subsection{Humanoid --- Unitree G1: Tabletop Manipulation and Locomotion}
\label{app:rw_g1}

\rwhead{Setting}
\begin{rwcard}
Robot & Unitree G1 humanoid (29 body DoF) with two Wuji Gen-1 dexterous hands; the right thumb is mechanically frozen, and waist, legs and body height are locked during manipulation \\
Cameras given to GPT-6-Astra & Head-mounted Azure Kinect, RGB only; depth is recorded but neither given to the model nor used as a collision guard. Measured robot state is additionally available through tools \\
Action interface & Manipulation: \texttt{move\_arms} (14 absolute arm-joint targets in radians, clamped to joint limits and interpolated by the offboard arm controller) and \texttt{set\_hands} (binary open/close profile per hand, no individual-finger, grip-width or force control; hands driven by the onboard Jetson). Locomotion: \mbox{\texttt{turn\_to\_heading}} (IMU yaw feedback), \mbox{\texttt{walk\_forward}} (fixed 0.15\,m/s for a commanded duration, at most 1.5\,s per step) and \texttt{stop}, executed by the vendor high-level controller; GPT-6-Astra never commands leg joints, and no translational odometry is available to the policy \\
Harness & Tool calls through the OpenAI Responses API (\texttt{gpt-6-astra}, reasoning effort \texttt{max}, mode \texttt{pro}, \texttt{store=false}); model calls, tool calls and camera frames are archived \\
Budget and success rule & Six retained runs, three per task; manipulation and locomotion use different tool sets and are evaluated separately. No common 60-minute cutoff applies: both successful manipulation runs exceeded one hour. A person judges the binary outcome from third-person video after the run, separately from the model's completion statement; the requested $\sim$10\,cm stopping distance is approximate and the final gap was not independently measured. Operator emergency stop throughout; the locomotion runs additionally use a slack safety-gantry strap \\
\end{rwcard}

\rwhead{Context provided}
\begin{itemize}[leftmargin=1.4em,itemsep=0pt,topsep=1pt]
\item Task instruction, embodiment description, tool schemas, joint limits, camera placement, binary hand open/close profiles, locomotion and safety constraints, RGB observations, robot-state feedback, and the current run's interaction history.
\item No demonstrations, no demonstration-derived ICL package, and no semantic pick, place or navigation primitives; every motion is chosen by the model, one tool call at a time.
\end{itemize}

\rwhead{Tasks and results}
\begin{rwtasks}
Pick up the blue plush toy and then drop it back onto the table & 3 & 2 & 74:31; $\sim$87:47 \\
Turn, walk to directly in front of the table and stop about 10\,cm from it without touching it$^{\dagger}$ & 3 & 3 & 8--13\,min \\
\end{rwtasks}
{\fontsize{8}{9.6}\selectfont $^{\dagger}$~The locomotion instruction was given in Chinese; the wording above is a translation. Wall-clock times are listed for the two successful manipulation runs; per-run details below.\par}

\begin{center}
\fontsize{8}{9.6}\selectfont
\setlength{\tabcolsep}{4pt}\renewcommand{\arraystretch}{1.15}

\begin{tabular}{lcccl}
\toprule
\textbf{Run} & \textbf{Outcome} & \textbf{Tool calls} & \textbf{Image calls} & \textbf{Elapsed time} \\
\midrule
pickRun1 & Failure & 29 & 14 & ---$^{*}$ \\
pickRun2 & Success & 74 & 36 & $\sim$87\,min\,47\,s \\
pickRun3 & Success & 65 & 32 & 74\,min\,31\,s \\
walkRun1 & Success & 18 & 9 & 12\,min\,49\,s \\
walkRun2 & Success & 12 & 6 & 8\,min\,13\,s \\
walkRun3 & Success & 13 & 7 & 10\,min\,12\,s \\
\bottomrule
\end{tabular}
\end{center}
{\fontsize{8}{9.6}\selectfont $^{*}$~Elapsed time unavailable from the archived records.\par}

Manipulation succeeds in 2/3 trials and locomotion in 3/3, giving 5/6 overall. The failed manipulation run drops the toy off the table, and recovery is not completed within the fixed-body workspace. With three trials per task, these results establish feasibility rather than reliability.

\rwhead{Findings}
\begin{itemize}[leftmargin=1.4em,itemsep=1pt,topsep=1pt]
\item \textbf{Observation-conditioned approach under a safety constraint.} In walk run 1, GPT-6-Astra shortens successive forward commands from 0.8\,s to 0.5, 0.3, 0.2 and 0.15\,s after fresh images, and all three locomotion runs stop at a self-estimated 0.15--0.20\,m rather than 0.10\,m, citing the no-contact requirement. This shows conservative progress assessment, not metric stopping accuracy.
\item \textbf{Valid joint targets do not guarantee faithful execution.} In pick run 3, all 26 arm motions return \texttt{stalled\_short\_of\_target} (18) or \texttt{timeout} (8), with median final tracking error 0.07065\,rad and maximum 0.1904\,rad, yet the human-labelled outcome is a success: useful motion occurs without meeting the joint-target criterion. GPT-6-Astra notices the discrepancies and adjusts, but reliable error compensation is not established.
\item \textbf{Slow supervision remains costly on hardware.} Across 108 archived API calls, median request latency is 24.84\,s, and API request wall time accounts for 88.8--91.7\% of elapsed run time; image capture takes 4.5--9.0\%. The remainder includes tool execution and harness overhead and is not a direct measure of physical motion time. Both successful manipulation runs last over an hour.
\end{itemize}

\par\noindent\begin{minipage}{\linewidth}
\rwhead{Visualization}
\begin{center}
\includegraphics[width=\linewidth]{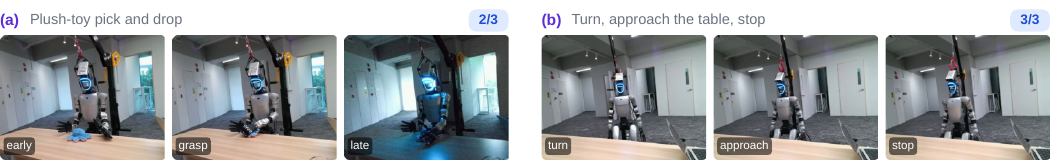}
\captionof{figure}{Unitree G1 with Wuji hands: fixed-body plush-toy pick-and-drop (2/3 successes) and table approach through the vendor velocity interface (3/3), evaluated separately. Frames are from third-person recordings; GPT-6-Astra received the head-camera RGB view and tool-returned robot state. Binary hand profiles were used.}
\end{center}
\end{minipage}

\end{document}